%% file: arxiv.tex
\documentclass[10pt,twocolumn,letterpaper]{article}

\usepackage[pagenumbers]{wacv}

\input{preamble}
\usepackage{float}
\usepackage{listings}

\AtBeginDocument{\crefname{appendix}{Sec.}{Secs.}}

\definecolor{wacvblue}{rgb}{0.21,0.49,0.74}
\usepackage[pagebackref,breaklinks,colorlinks,allcolors=wacvblue]{hyperref}
\usepackage[accsupp]{axessibility}  %
\def\wacvPaperID{1893} %
\def\confName{WACV}
\def\confYear{2027}

\title{Goal Images Are Control Targets:\\
Spatially-Grounded Goal-Image Synthesis for Robotic Manipulation}

\author{
Riddhi Chatterjee \quad
Ritish Shrirao \quad
Kushagra Mehrotra \quad
Viswanath Gopalakrishnan
\\
International Institute of Information Technology Bangalore, India
\\
{\footnotesize\tt \{Riddhi.Chatterjee, Ritish.Shrirao, Kushagra.Mehrotra, viswanath.g\}@iiitb.ac.in}
}

\begin{document}
\maketitle
\raggedbottom %
\input{sec/0_abstract}
\input{sec/1_intro}
\input{sec/2_related}
\input{sec/3_method}
\input{sec/4_experiments}
\input{sec/5_discussion}
{
    \small
    \bibliographystyle{ieeenat_fullname}
    \bibliography{main}
}

\clearpage
\appendix
\setcounter{figure}{0}\setcounter{table}{0}\setcounter{equation}{0}
\renewcommand{\thefigure}{S\arabic{figure}}\renewcommand{\theHfigure}{S\arabic{figure}}
\renewcommand{\thetable}{S\arabic{table}}\renewcommand{\theHtable}{S\arabic{table}}
\renewcommand{\theequation}{S\arabic{equation}}\renewcommand{\theHequation}{S\arabic{equation}}
\twocolumn[{\centering\Large\bfseries Supplementary Material\par\vspace{1.5em}}]
\raggedbottom
\thispagestyle{empty}

\noindent This supplementary material provides: downstream results with a third, CALVIN-fine-tuned goal-conditioned policy, Octo-ft, including its training recipe, chaining results and mask ablation (\cref{sec:supp:octo}); the full per-dimension dissociation analysis under all three CALVIN policies (\cref{sec:supp:downstream}); module-sensitivity and failure-case analysis (\cref{sec:supp:failure}); implementation and training details (\cref{sec:supp:impl}); full I2I-Bench results (\cref{sec:appendix:existing-eval}); editing-metric ablations under both editing evaluators (\cref{sec:appendix:edit-ablation}); the data-extraction pipeline and evaluation frameworks (\cref{sec:appendix:methods}); and all verbatim VLM evaluation prompts (\cref{sec:appendix:vlm-prompts}). The downstream success tables, the training-data quality study, the chained-prediction visualization, and runtimes appear in the main paper.

\section{Downstream Results with a CALVIN-Fine-Tuned Octo Policy}
\label{sec:supp:octo}

GR-MG and GHIL-Glue, the two CALVIN policies of the main paper, are both publicly released checkpoints. To test whether the generator ranking depends on these particular policies, we fine-tune a third, architecturally different goal-conditioned policy ourselves, Octo-ft, and repeat the CALVIN evaluation with it. This section describes the policy and reports its downstream results. This section also repeats the main paper's mask ablation under Octo-ft, and Octo-ft enters the per-dimension dissociation analysis and its significance test (\cref{sec:supp:downstream}).

\noindent\textbf{Policy and training.}
We fine-tune all parameters of Octo-Base~\cite{Octo2024} on $1{,}214$ language-annotated windows (${\sim}64$k frames, all $34$ tasks) from the CALVIN training split, which is disjoint from the validation episodes our test rows are drawn from; \cref{tab:supp-octo-hparams} lists the full recipe. The policy predicts CALVIN's relative actions from a two-frame observation window, and goal images are relabeled uniformly from the future of each window. Two design choices matter for using it to rank goal images. First, each training sample is conditioned on either the goal image or the language instruction, with equal probability, so the policy learns to act from the goal image alone. Second, the goal image receives its own random augmentation, independent of the observation's; with a shared augmentation the policy could match goal and observation through identical crop and colour artifacts, a shortcut that is absent at test time, where the goal comes from a separate generator. The policy is trained only on real frames and never sees any generator's output. We use the final checkpoint (step $30$k), which we have released\refckpt. An intermediate checkpoint at step $15$k, midway through the cosine learning-rate decay, solved none of the first $40$ test rows when given the ground-truth goal image and instruction, against $30$ of $40$ for the final checkpoint, so we select checkpoints by closed-loop success rather than by validation loss. At test time the policy is frozen and conditioned on the goal image alone, on the same $1{,}087$ test rows, with the same seed, $360$-step horizon and chaining protocol as the main paper.

\begin{table}[t]
  \caption{\textbf{Octo-ft fine-tuning and evaluation recipe.} The policy is trained only on real CALVIN frames and never sees any generator's output.}
  \label{tab:supp-octo-hparams}
  \centering\footnotesize
  \setlength{\tabcolsep}{3pt}
  \begin{tabular}{@{}ll@{}}
    \toprule
    \multicolumn{2}{@{}l}{\emph{Training}} \\
    Initialization & Octo-Base (released checkpoint) \\
    Fine-tuned parameters & all \\
    Data & CALVIN D training split, $1{,}214$ \\
     & annotated windows (${\sim}64$k frames), \\
     & all $34$ tasks ($35$--$37$ each) \\
    Cameras & static $256^2$, gripper $128^2$ \\
    Obs.\ window / action chunk & $2$ / $4$ steps \\
    Actions & CALVIN relative actions ($7$-D) \\
    Goal relabeling & uniform over future frames \\
    Task conditioning & goal image or instruction \\
     & ($p{=}0.5$ each) \\
    Augmentation & resized crop, colour jitter; \\
     & goal augmented independently \\
    Optimizer & AdamW (wd $0.01$, clip $1.0$) \\
    Learning rate & cosine, peak $3{\times}10^{-4}$ \\
     & ($1$k warmup) \\
    Batch size / steps & $32$ / $30$k (${\sim}14$ epochs) \\
    Hardware & $1{\times}$ RTX 6000 Ada, ${\approx}4$\,h \\
    \midrule
    \multicolumn{2}{@{}l}{\emph{Evaluation}} \\
    Checkpoint & final (step $30$k) \\
    Conditioning & goal image only \\
    Protocol & as for GR-MG and GHIL-Glue: \\
     & $1{,}087$ rows, seed $0$, $360$ steps \\
    \bottomrule
  \end{tabular}
\end{table}

\begin{figure*}[t]
  \centering
  \includegraphics[width=\linewidth]{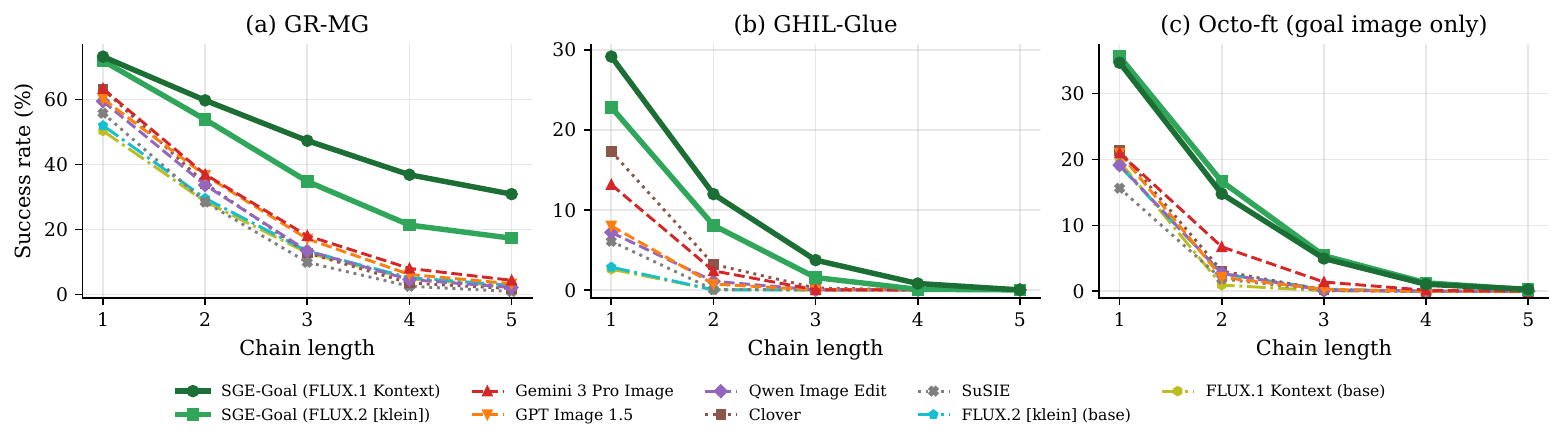}
  \caption{\textbf{Downstream success vs.\ chain length under all three CALVIN policies}: the two pre-trained policies of the main paper, (a)~GR-MG and (b)~GHIL-Glue, and (c)~our CALVIN-fine-tuned Octo (Octo-ft), conditioned on the goal image only. Same test rows, chain skeletons and aggregation in all three panels. Under every policy the SGE-Goal goal images (green) give the highest success at every chain length at which any generator succeeds (all reach $0\%$ under GHIL-Glue at CL${=}5$), while the other generators fall towards zero.}
  \label{fig:supp-chaining-octo}
\end{figure*}

\noindent\textbf{Generator ranking and chaining.}
\Cref{tab:supp-octo} reports the results. Octo-ft ranks the nine generators consistently with GR-MG (Spearman $\rho{=}0.95$) and GHIL-Glue ($\rho{=}0.88$), as closely as those two agree with each other ($\rho{=}0.90$; mean success over CL${=}1$--$5$). Both SGE-Goal variants come within $1.6$ points of the ground-truth oracle at CL${=}1$ and beat every other generator by $13.3$ points (95\% bootstrap interval over test rows $[9.8, 15.0]$). Apart from SuSIE ($15.6\%$), the remaining generators lie within $2.3$ points of each other at CL${=}1$ ($19.0$--$21.3\%$), about two binomial standard errors, so their order should not be over-read. \Cref{fig:supp-chaining-octo} shows how this lead evolves under chaining, alongside the two policies of the main paper. Octo-ft has a lower ceiling than GR-MG (its ground-truth oracle falls from $36.2\%$ at CL${=}1$ to $6.2\%$ at CL${=}3$), but within that ceiling the pattern is the same: at CL${=}2$ the SGE-Goal variants retain $14.8\%$ and $16.7\%$, against $6.7\%$ for the best other generator (Gemini~3 Pro Image), and from CL${=}4$ every other generator is at or below $0.2\%$. The adherence-over-photorealism dissociation also holds under this policy, and remains statistically significant (\cref{sec:supp:downstream}).

\begin{table}[t]
  \caption{\textbf{A third CALVIN policy: Octo fine-tuned on CALVIN (Octo-ft)}, success rate (SR\%, $\uparrow$) at chain lengths $1$--$5$, conditioned on the goal image only. Same $1{,}087$ test rows, seed and rollout protocol as the GR-MG and GHIL-Glue results in the main paper. The two SGE-Goal variants lead at every chain length and are the only generators close to the ground-truth oracle. FK${=}$FLUX.1~Kontext, F2K${=}$FLUX.2~klein, base${=}$untuned editor. \best{Best} (excl.\ oracle) per column in bold; SGE-Goal rows shaded.}
  \label{tab:supp-octo}
  \centering\footnotesize
  \setlength{\tabcolsep}{4pt}
  \begin{tabular}{lccccc}
    \toprule
    Method & 1 & 2 & 3 & 4 & 5 \\
    \midrule
    \textit{GT goal (oracle)} & \textit{36.2} & \textit{17.1} & \textit{6.2} & \textit{1.4} & \textit{0.3} \\
    \oursrow SGE-Goal (FK)    & 34.7 & 14.8 & 5.0 & 1.1 & \best{0.3} \\
    \oursrow SGE-Goal (F2K)   & \best{35.6} & \best{16.7} & \best{5.4} & \best{1.3} & \best{0.3} \\
    Gemini 3 Pro Image        & 21.0 & 6.7 & 1.4 & 0.2 & 0.0 \\
    Clover                    & 21.3 & 3.1 & 0.2 & 0.0 & 0.0 \\
    GPT Image 1.5             & 21.0 & 2.1 & 0.3 & 0.0 & 0.0 \\
    Qwen Image Edit           & 19.1 & 2.7 & 0.1 & 0.0 & 0.0 \\
    SuSIE                     & 15.6 & 1.9 & 0.1 & 0.0 & 0.0 \\
    FLUX.2 klein (base)       & 19.0 & 2.7 & 0.3 & 0.0 & 0.0 \\
    FLUX.1 Kontext (base)     & 19.7 & 0.9 & 0.1 & 0.0 & 0.0 \\
    \bottomrule
  \end{tabular}
\end{table}

\noindent\textbf{Why goal-image-only conditioning.}
We also evaluated Octo-ft with both the goal image and the instruction, as GR-MG is evaluated. Its oracle success at CL${=}1$ rises to $50.1\%$, but the nine generators then span only $4.0$ points ($45.3$--$49.3\%$), compared with $20.0$ points goal-image-only (\cref{tab:supp-octo-lang}): having learned to solve tasks from the instruction alone, the policy largely ignores the goal image when a correct instruction is present. A policy can only measure goal-image quality if it depends on the goal image, so we report the goal-image-only results.

\begin{table}[t]
  \caption{\textbf{Language masks the goal image.} Octo-ft success at CL${=}1$ (SR\%, $\uparrow$) when conditioned on the goal image together with the language instruction, and on the goal image alone. With the instruction, the policy solves most tasks from language and the goal generator barely matters (spread $4.0$ points); without it, generators separate by $20.0$ points. We therefore report goal-image-only results. \best{Best} (excl.\ oracle) per column in bold; SGE-Goal rows shaded.}
  \label{tab:supp-octo-lang}
  \centering\footnotesize
  \setlength{\tabcolsep}{4pt}
  \begin{tabular}{lcc}
    \toprule
    Method & Goal image + instruction & Goal image only \\
    \midrule
    \textit{GT goal (oracle)} & \textit{50.1} & \textit{36.2} \\
    \oursrow SGE-Goal (FK)    & \best{49.3} & 34.7 \\
    \oursrow SGE-Goal (F2K)   & 48.6 & \best{35.6} \\
    Gemini 3 Pro Image        & 45.6 & 21.0 \\
    Clover                    & 48.1 & 21.3 \\
    GPT Image 1.5             & 45.9 & 21.0 \\
    Qwen Image Edit           & 45.7 & 19.1 \\
    SuSIE                     & 45.3 & 15.6 \\
    FLUX.2 klein (base)       & 45.5 & 19.0 \\
    FLUX.1 Kontext (base)     & 46.5 & 19.7 \\
    \midrule
    Spread across generators  & 4.0 & 20.0 \\
    \bottomrule
  \end{tabular}
\end{table}

\noindent\textbf{Mask ablation.}
We repeated the mask ablation of the main paper with Octo-ft, conditioned on the goal image only, on the same test rows and with the same protocol (\cref{tab:supp-octo-ablation}). As under GR-MG, the final mask contributes the most: removing it lowers success at CL${=}1$ by $14.9$ points for FLUX.1~Kontext and $11.0$ points for FLUX.2~klein (paired 95\% bootstrap intervals over test rows $[11.9, 18.1]$ and $[7.7, 14.3]$), and the full model leads both ablations at every chain length. Mask-free fine-tuning also falls below the untuned base for FLUX.1~Kontext ($15.1\%$ vs.\ $19.7\%$ at CL${=}1$; interval for the difference $[{-}7.1, {-}2.1]$). One result differs from GR-MG: for FLUX.2~klein, the variant with only the initial mask does not beat the mask-free one under Octo-ft ($24.6\%$ vs.\ $25.7\%$; interval $[{-}4.0, 1.6]$), so for this editor the gain comes from the final mask alone.

\begin{table}[t]
  \caption{\textbf{Mask ablation under Octo-ft} (SR\%, $\uparrow$), goal image only, at chain lengths $1$--$5$; same test rows and protocol as the GR-MG and GHIL-Glue ablation in the main paper. ``full'' uses both masks and the static-scene loss; base${=}$untuned editor. \best{Best} per block in bold; SGE-Goal rows shaded.}
  \label{tab:supp-octo-ablation}
  \centering\footnotesize
  \setlength{\tabcolsep}{4pt}
  \begin{tabular}{lccccc}
    \toprule
    & 1 & 2 & 3 & 4 & 5 \\
    \midrule
    \multicolumn{6}{@{}l}{\textbf{FLUX.1 Kontext}} \\
    \oursrow SGE-Goal (full)  & \best{34.7} & \best{14.8} & \best{5.0} & \best{1.1} & \best{0.3} \\
    \quad w/o Final Mask      & 19.8 & 6.5 & 1.2 & 0.1 & 0.0 \\
    \quad w/o Masks           & 15.1 & 2.4 & 0.2 & 0.0 & 0.0 \\
    \quad base                & 19.7 & 0.9 & 0.1 & 0.0 & 0.0 \\
    \midrule
    \multicolumn{6}{@{}l}{\textbf{FLUX.2 klein}} \\
    \oursrow SGE-Goal (full)  & \best{35.6} & \best{16.7} & \best{5.4} & \best{1.3} & \best{0.3} \\
    \quad w/o Final Mask      & 24.6 & 3.4 & 0.5 & 0.0 & 0.0 \\
    \quad w/o Masks           & 25.7 & 6.6 & 0.9 & 0.1 & 0.0 \\
    \quad base                & 19.0 & 2.7 & 0.3 & 0.0 & 0.0 \\
    \bottomrule
  \end{tabular}
\end{table}

\section{Full Dissociation Analysis}
\label{sec:supp:downstream}

The main paper reports downstream success and the headline dissociation; here we provide the complete per-dimension correlation analysis, for all three CALVIN policies (GR-MG and GHIL-Glue from the main paper, and Octo-ft, \cref{sec:supp:octo}) and for real-world-aligned SIMPLER-Bridge.

\begin{figure}[t]
  \centering
  \includegraphics[width=\linewidth]{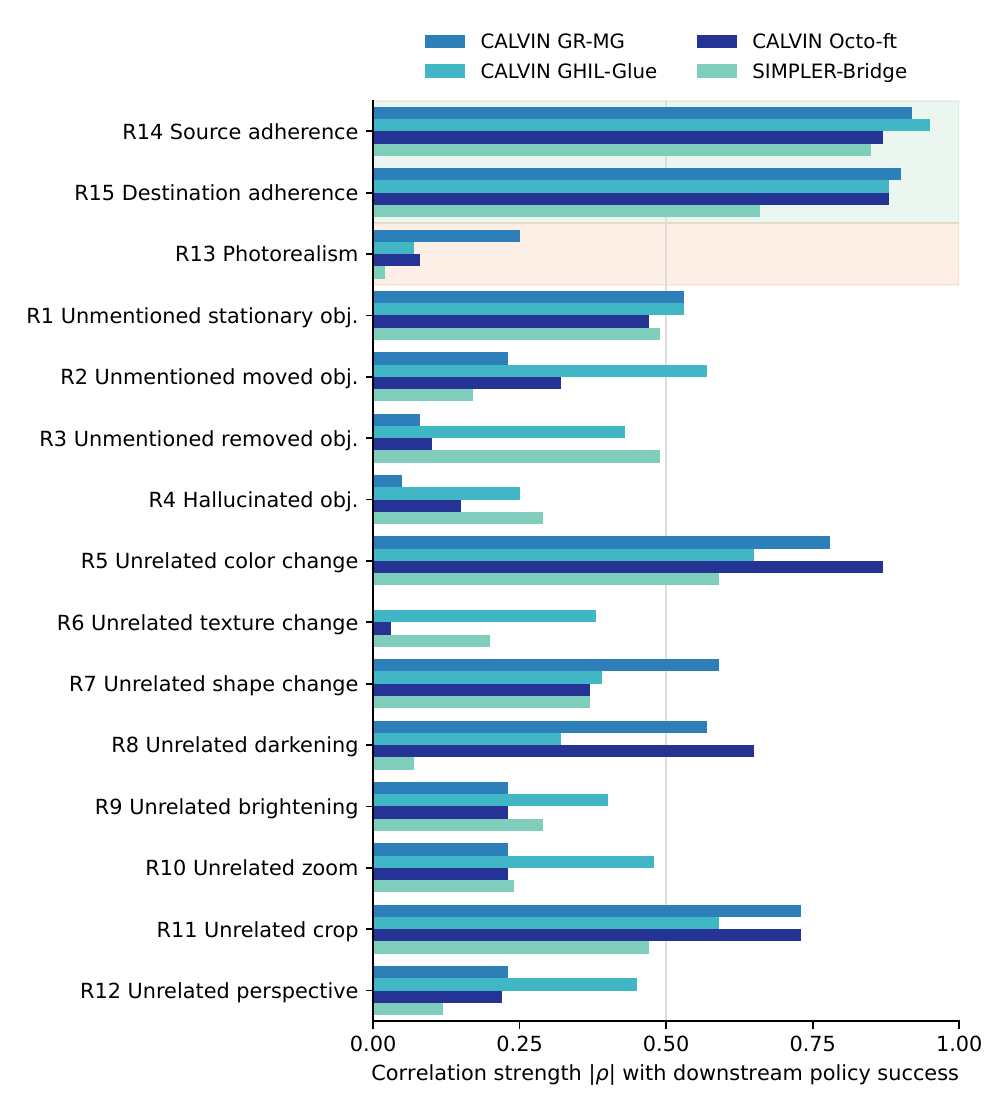}
  \caption{\textbf{Full per-dimension dissociation across all downstream success-rate sources.} Correlation strength (absolute Spearman $|\rho|$; higher means more predictive) of each of the 15 evaluation dimensions with downstream policy success, for the three CALVIN policies (GR-MG, GHIL-Glue, and Octo-ft conditioned on the goal image only) and for real-world-aligned SIMPLER-Bridge. Adherence (R14, R15; green band) is the only group that is strongly and consistently predictive across all four sources; photorealism (R13; orange band) is weak everywhere, and the remaining consistency and scene dimensions are mixed and do not agree across benchmarks. This is the detailed, fully-labelled version of \cref{tab:supp-dissoc-full}.}
  \label{fig:dissociation-supp}
\end{figure}

\noindent\textbf{Full per-dimension dissociation.}
\Cref{tab:supp-dissoc-full} and \cref{fig:dissociation-supp} give the Spearman correlation of \emph{every} evaluation dimension with downstream success, for the three CALVIN policies (GR-MG, GHIL-Glue, and the CALVIN-fine-tuned Octo-ft of \cref{sec:supp:octo}) and for SIMPLER-Bridge. Only \emph{adherence} (R14, R15) is strongly and consistently predictive across all four; the consistency dimensions are mixed and do not agree across benchmarks, and photorealism (R13) is weak everywhere. This is why no aggregate weighting of editing-quality dimensions reliably reproduces the downstream ranking.

\begin{table}[!t]
  \caption{Correlation strength (absolute Spearman $|\rho|$; higher means more predictive) of each dimension with downstream success, for the three CALVIN policies (avg.\ over CL${=}1$--$5$) and for SIMPLER-Bridge. Adherence (R14/R15) is the only consistently strong predictor on all four; consistency dimensions are mixed and inconsistent across benchmarks, and photorealism (R13) is weak everywhere. GHIL = GHIL-Glue; Octo-ft = Octo fine-tuned on CALVIN, conditioned on the goal image only.}
  \label{tab:supp-dissoc-full}
  \centering\scriptsize
  \setlength{\tabcolsep}{2pt}
  \begin{tabular}{@{}llccccc@{}}
    \toprule
    & & & \multicolumn{4}{c}{$|\rho|$ with SR} \\
    \cmidrule(lr){4-7}
    & & & \multicolumn{3}{c}{CALVIN} & \multirow{2}{*}{\shortstack{SIMPLER-\\Bridge}} \\
    \cmidrule(lr){4-6}
    & Dimension & Group & GR-MG & GHIL & Octo-ft & \\
    \midrule
    R14 & Source adherence & Adher. & \best{0.92} & \best{0.95} & \best{0.87} & \best{0.85} \\
    R15 & Destination adherence & Adher. & \best{0.90} & \best{0.88} & \best{0.88} & \best{0.66} \\
    R1 & Unmentioned stationary obj. & Consist. & 0.53 & 0.53 & 0.47 & 0.49 \\
    R2 & Unmentioned moved obj. & Consist. & 0.23 & 0.57 & 0.32 & 0.17 \\
    R3 & Unmentioned removed obj. & Consist. & 0.08 & 0.43 & 0.10 & 0.49 \\
    R4 & Hallucinated obj. & Consist. & 0.05 & 0.25 & 0.15 & 0.29 \\
    R5 & Unrelated color change & Consist. & 0.78 & 0.65 & 0.87 & 0.59 \\
    R6 & Unrelated texture change & Consist. & 0.00 & 0.38 & 0.03 & 0.20 \\
    R7 & Unrelated shape change & Consist. & 0.59 & 0.39 & 0.37 & 0.37 \\
    R8 & Unrelated darkening & Consist. & 0.57 & 0.32 & 0.65 & 0.07 \\
    R9 & Unrelated brightening & Consist. & 0.23 & 0.40 & 0.23 & 0.29 \\
    R10 & Unrelated zoom & Consist. & 0.23 & 0.48 & 0.23 & 0.24 \\
    R11 & Unrelated crop & Consist. & 0.73 & 0.59 & 0.73 & 0.47 \\
    R12 & Unrelated perspective & Consist. & 0.23 & 0.45 & 0.22 & 0.12 \\
    R13 & Photorealism & Photoreal. & 0.25 & 0.07 & 0.08 & 0.02 \\
    \bottomrule
  \end{tabular}
\end{table}

\begin{table}[!t]
  \caption{\textbf{Raw rates (CALVIN, our evaluator) vs.\ CALVIN success.} Per-method R14 (source adherence), R15 (destination adherence), R13 (photorealism), and mean policy success over CL${=}1$--$5$ for the three CALVIN policies, GR-MG, GHIL-Glue and Octo-ft (our CALVIN-fine-tuned Octo policy, conditioned on the goal image only). SGE-Goal leads on adherence but trails on photorealism, and its two variants are the top two generators under all three policies; the photorealism leader (Qwen) sits near the bottom downstream---only adherence tracks success. \best{Best} per column in bold; SGE-Goal rows shaded.}
  \label{tab:supp-rawscores-calvin}
  \centering\scriptsize
  \setlength{\tabcolsep}{3.2pt}
  \begin{tabular}{lcccccc}
    \toprule
     & & & & \multicolumn{3}{c}{CALVIN SR (\%)} \\
    \cmidrule(l){5-7}
    Method & R14 & R15 & R13 & GR-MG & GHIL & Octo-ft \\
    \midrule
    \oursrow SGE-Goal (FLUX.1 Kontext) & \best{0.361} & 0.461 & 0.433 & \best{49.6} & \best{9.1} & 11.2 \\
    \oursrow SGE-Goal (FLUX.2 klein)   & 0.348 & \best{0.469} & 0.224 & 39.8 & 6.5 & \best{11.8} \\
    Gemini 3 Pro Image                 & 0.350 & 0.459 & 0.688 & 26.2 & 3.1 & 5.9 \\
    GPT Image 1.5                      & 0.318 & 0.291 & 0.662 & 24.7 & 1.7 & 4.7 \\
    Clover                             & 0.319 & 0.362 & 0.014 & 23.1 & 4.1 & 4.9 \\
    Qwen Image Edit                    & 0.259 & 0.404 & \best{0.709} & 22.7 & 1.7 & 4.4 \\
    FLUX.2 klein (base)                & 0.184 & 0.203 & 0.485 & 20.6 & 0.6 & 4.4 \\
    FLUX.1 Kontext (base)              & 0.132 & 0.149 & 0.399 & 19.8 & 0.5 & 4.1 \\
    SuSIE                              & 0.231 & 0.180 & 0.001 & 19.5 & 1.2 & 3.5 \\
    \bottomrule
  \end{tabular}
\end{table}

\noindent\textbf{Raw rates track success through adherence, not photorealism.}
\Cref{tab:supp-rawscores-calvin,tab:supp-rawscores-bridge} pair the raw R14/R15/R13 rates per method with downstream success---on CALVIN under all three policies, and on BridgeDataV2 with SIMPLER-Bridge. The dissociation is visible in the rates themselves: the highest-photorealism editors (Qwen, GPT, and the untuned FLUX bases) sit near the bottom downstream on \emph{both} benchmarks, so photorealism does not buy control success, whereas adherence does. On CALVIN, SGE-Goal leads on adherence while trailing on photorealism, and its two variants are the top two generators under \emph{all three} policies, GR-MG, GHIL-Glue and Octo-ft (\cref{tab:supp-rawscores-calvin}). On the real-world-aligned SIMPLER-Bridge benchmark the mechanism is identical, but the adherence leader is Gemini~3 Pro Image, which is accordingly the one generator that outscores SGE-Goal downstream (\cref{tab:supp-rawscores-bridge}); SGE-Goal still leads \emph{source} adherence (R14) and, despite a modest photorealism rate, ranks a close second overall---above every base editor and every open baseline.

\noindent\textbf{What these tables establish.}
SGE-Goal is a fine-tuning \emph{strategy} for turning a base editor into a stronger goal generator for control; the relevant test is therefore whether it (i)~improves substantially over the untuned base models and (ii)~outperforms the mask-free and initial-mask-only variants, both of which are established by the downstream method comparison and mask ablations in the main paper. Against that test, leading every generator on CALVIN under three independent policies and trailing only Gemini---by a small margin---on real-world SIMPLER-Bridge shows the strategy is highly effective, turning small open editors into goal generators competitive with, and usually stronger than, far larger proprietary systems. Surpassing Gemini on every benchmark is not part of this claim, and the one benchmark where Gemini leads is itself consistent with the dissociation: Gemini wins there by leading on adherence, not on the photorealism that its scale most improves.

\begin{table}[t]
  \caption{\textbf{Raw rates (BridgeDataV2, our evaluator) vs.\ SIMPLER-Bridge success.} Per-method R14 (source adherence), R15 (destination adherence), R13 (photorealism), and mean SIMPLER-Bridge policy success. The dissociation holds again: high-photorealism editors collapse downstream (Qwen, GPT, and the FLUX bases all score ${\le}3.1$), so photorealism does not track success---adherence does. Gemini leads adherence (R15) and is the only generator to outscore SGE-Goal; SGE-Goal leads source adherence (R14) and, despite a modest photorealism rate, ranks a close second overall, above every base editor and open baseline. \best{Best} per column in bold; SGE-Goal rows shaded; Clover is CALVIN-only and omitted.}
  \label{tab:supp-rawscores-bridge}
  \centering\footnotesize
  \setlength{\tabcolsep}{2.8pt}
  \begin{tabular}{lcccc}
    \toprule
    Method & R14 & R15 & R13 & SIMPLER SR (\%) \\
    \midrule
    Gemini 3 Pro Image                 & 0.431 & \best{0.558} & \best{0.915} & \best{12.5} \\
    \oursrow SGE-Goal (FLUX.2 klein)   & \best{0.517} & 0.451 & 0.342 & 10.4 \\
    \oursrow SGE-Goal (FLUX.1 Kontext) & 0.514 & 0.417 & 0.591 & 8.3 \\
    SuSIE                              & 0.374 & 0.373 & 0.230 & 7.3 \\
    Qwen Image Edit                    & 0.337 & 0.453 & 0.898 & 3.1 \\
    GPT Image 1.5                      & 0.336 & 0.457 & 0.821 & 2.1 \\
    FLUX.2 klein (base)                & 0.257 & 0.376 & 0.542 & 2.1 \\
    FLUX.1 Kontext (base)              & 0.342 & 0.325 & 0.810 & 2.1 \\
    \bottomrule
  \end{tabular}
\end{table}

\noindent\textbf{Statistical significance.}
With nine generators (the unit of downstream measurement), an exact permutation test over all $9!$ label assignments yields $p{<}10^{-3}$ for source adherence ($\rho{=}0.98$) and $p{=}0.001$ for destination adherence ($\rho{=}0.92$), whereas photorealism is indistinguishable from chance ($\rho{=}0.03$, $p{=}0.95$). Here the $p$-value is the fraction of generator-label permutations that yield a correlation at least as strong as observed, so the dissociation is not a small-sample artifact. These values use GR-MG success at CL${=}1$. The same test against the success of our CALVIN-fine-tuned Octo policy at CL${=}1$ (\cref{sec:supp:octo}) gives $\rho{=}0.78$ ($p{=}0.017$) for source adherence and $\rho{=}0.76$ ($p{=}0.021$) for destination adherence, while photorealism again shows no relation ($\rho{=}{-}0.09$, $p{=}0.82$): the dissociation holds under an independently trained policy.

\section{Module-Sensitivity and Failure-Case Analysis}
\label{sec:supp:failure}

\noindent\textbf{Module sensitivity.}
SGE-Goal comprises three modules: a training-free grounding stage (Step~1, initial mask $M_i$), a final-mask predictor (Step~2, $M_f$), and a goal-image synthesizer (Step~3) conditioned on both masks. Our downstream mask ablation in the main paper isolates each module's contribution and shows that the pipeline is most sensitive to its spatial-grounding components, increasingly so under chaining. Conditioning Step~3 on the predicted final mask contributes the most: removing it lowers GR-MG success at chain length~$1$ from $73.1\%$ to $60.1\%$ for FLUX.1~Kontext and from $71.8\%$ to $63.5\%$ for FLUX.2~klein, with far larger drops at long horizons (e.g., FLUX.1~Kontext falls from $30.9\%$ to $3.5\%$ at chain length~$5$). Removing both masks---i.e., vanilla fine-tuning with no spatial grounding---reduces success further and, under chaining, drops \emph{below} the untuned base editor (e.g., FLUX.1~Kontext on GR-MG at chain length~$2$: $21.3\%$ versus the base's $28.6\%$), indicating that mask-free fine-tuning induces a form of catastrophic forgetting that the explicit masks prevent. The initial and final masks are thus complementary, and their benefit compounds with horizon length. The same ablation under the CALVIN-fine-tuned Octo policy is reported in \cref{sec:supp:octo} (\cref{tab:supp-octo-ablation}).

\begin{figure*}[t]
  \centering
  \includegraphics[width=\linewidth]{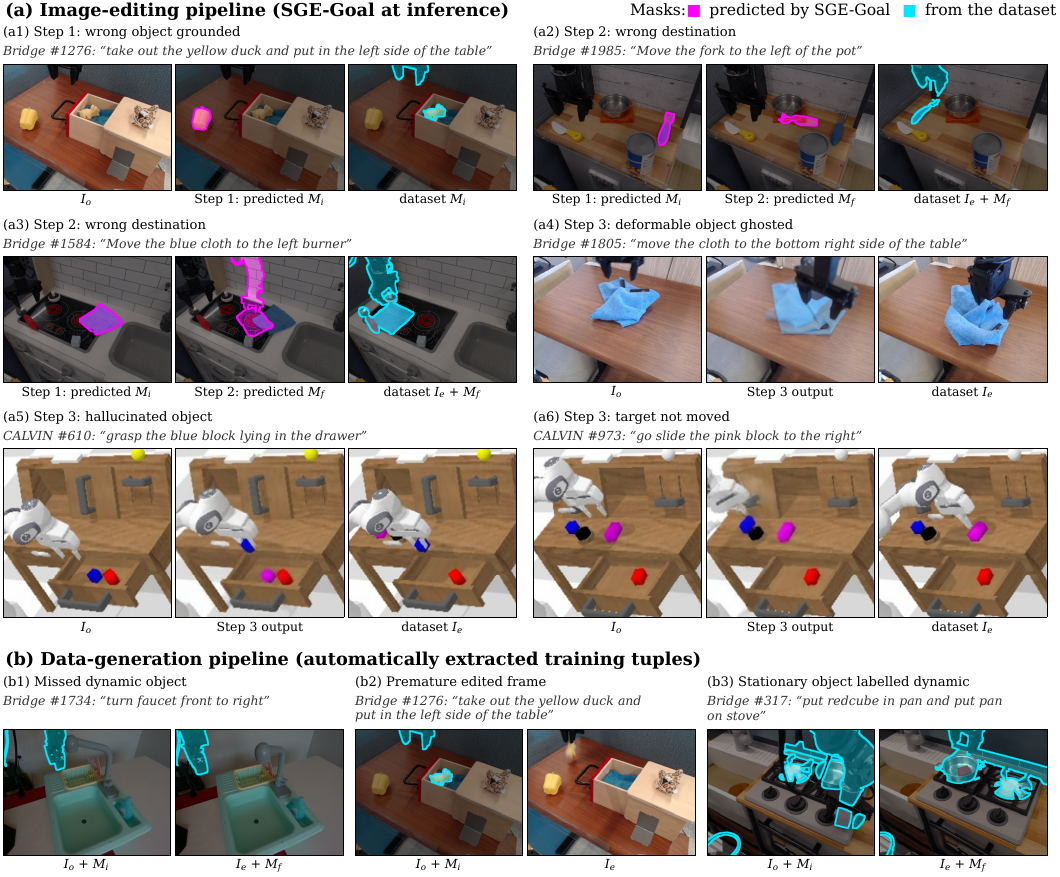}
  \caption{\textbf{Failure cases}, all real test-split examples. \textbf{(a)~Image-editing pipeline.} (a1)~Step~1 grounds a same-coloured distractor (yellow pepper) instead of the occluded target (yellow duck in the box). (a2, a3)~Step~2 places the object in the wrong spot: the fork on the pot instead of to its left (a2), and the cloth on the right burner instead of the left one (a3). (a4--a6)~Step~3, given the dataset masks: a deformable cloth rendered as a translucent double exposure (a4); a hallucinated magenta block where the grasped blue block was (a5); the target block left in place while the arm is altered (a6). Masks are outlined, with the rest of the image dimmed: magenta = predicted by SGE-Goal, cyan = from the dataset (dataset masks also include the robot arm). Steps~1--2 use the FLUX.2~klein pipeline on BridgeDataV2; Step~3 uses FLUX.1~Kontext. \textbf{(b)~Data-generation pipeline}, showing extracted training tuples. (b1)~The faucet's rotation is missed by motion clustering, so both masks contain only the arm. (b2)~The edited frame is taken while the duck is still held by the gripper. (b3)~A stationary burner grate is labelled dynamic and enters both masks.}
  \label{fig:supp-failures}
\end{figure*}

SGE-Goal involves two separate chains of models, which fail in different ways: the \emph{image-editing pipeline} that produces a goal image at inference (Steps~1--3), and the offline \emph{data-generation pipeline} that extracts the training tuples $(I_o, I_e, M_i, M_f, T)$ from robot videos (\cref{sec:appendix:data-pipeline}). \Cref{fig:supp-failures} shows real failures of each, all taken from the test splits.

\noindent\textbf{Failure cases of the image-editing pipeline.}
Errors propagate along the module chain: an incorrect initial mask $M_i$ misleads the final-mask predictor, and a poorly localized $M_f$ yields an edit in the wrong place. We observe the following failure modes (\cref{fig:supp-failures}a; Steps~1--2 are shown on BridgeDataV2 test cases run through the full pipeline from Step~1, and Step~3 is shown given the dataset masks, so that its errors are not inherited from earlier steps).
\begin{itemize}[leftmargin=*,itemsep=1pt,topsep=2pt]
  \item \emph{Step~1, wrong object.} When the target is occluded, or a distractor shares the attribute named in the instruction, grounding selects the distractor (\cref{fig:supp-failures}, a1: the yellow duck is hidden in a box, and the yellow pepper is masked instead); multiple visually near-identical objects cause the same ambiguity.
  \item \emph{Step~2, wrong destination.} Relational destinations are not always resolved, so $M_f$ puts the object in the wrong place (\cref{fig:supp-failures}, a2: the fork is placed on the pot instead of to its left; a3: the cloth stays on the right burner instead of moving to the left one).
  \item \emph{Step~3, synthesis errors given correct masks.} Deformable objects can be rendered as a translucent double exposure of their initial and final shapes (\cref{fig:supp-failures}, a4); an object can be hallucinated at the location the moved object vacated (\cref{fig:supp-failures}, a5: a magenta block appears in the drawer after the blue block is grasped); and the target can be left in place while only the arm changes (\cref{fig:supp-failures}, a6). Articulated objects such as drawers and doors move in 3D, which the 2D masks describe only approximately.
\end{itemize}
Because Step~1 is training-free and relies on the same kind of pre-trained detectors and segmenters as the data-generation pipeline (Florence-2, Grounding DINO, BLIP-2, SAM), it shares their sensitivity to out-of-distribution scenes, discussed below. As the main paper shows, SGE-Goal preserves scene structure across chained steps, so in aggregate these errors compound slowly and downstream success holds up well under chaining.

\noindent\textbf{Failure cases of the data-generation pipeline.}
The extracted tuples are reliable in aggregate (the human study in the main paper), but individual tuples can be wrong in the following ways (\cref{fig:supp-failures}b).
\begin{itemize}[leftmargin=*,itemsep=1pt,topsep=2pt]
  \item \emph{Missed dynamic object.} Dynamic objects are identified by clustering per-object displacement (LoFTR, then $K{=}2$ KMeans). Parts that move little relative to their size, typically articulated parts that rotate in place, can fall into the stationary cluster (\cref{fig:supp-failures}, b1: the faucet is turned, but $M_i$ and $M_f$ contain only the arm).
  \item \emph{Premature edited frame.} $I_e$ is the last frame after which the arm and the dynamic-object masks stop overlapping. If the object's mask is lost or the arm is under-segmented while the object is still held, this criterion fires too early (\cref{fig:supp-failures}, b2: the duck still hangs from the gripper, motion-blurred, instead of resting on the left of the table).
  \item \emph{Stationary object labelled dynamic.} $K{=}2$ clustering always produces a dynamic cluster, and XMem masks can drift while the arm occludes an object, so a stationary object can enter the edit masks (\cref{fig:supp-failures}, b3: the burner grate is included in $M_i$ and $M_f$).
\end{itemize}
Heavily occluded target objects and multiple visually near-identical objects, the dominant residual failure modes of the module chain, can mislead both pipelines.

\noindent\textbf{Limitations of the data-generation pipeline.}
The pipeline chains six large pre-trained models (EVF-SAM, BLIP-2, SAM ViT-H, XMem, LoFTR and, where instructions are missing, VideoChat-Flash), which has two drawbacks. First, it is \emph{compute-intensive}: excluding instruction generation, it takes about $22$\,s per video on one A6000 GPU (Table~5 of the main paper), i.e.\ several hundred GPU-hours for our ${\sim}75$k processed tuples, before counting videos that are rejected. This is a one-time, offline, parallelizable cost, but it grows with every new dataset. Second, and more importantly, these models were themselves trained with human supervision on web and everyday imagery, so they can be \emph{brittle out of distribution}: novel objects, unusual viewpoints, textures or lighting, and robot embodiments unlike those they were trained on can all lower detection, segmentation and tracking quality, and the resulting errors reach the training data unchecked. Our three datasets span real and simulated scenes and single- and multi-camera setups, and the human study shows that extraction is reliable on them in aggregate, but we have not measured how quality degrades further from these domains. Lighter or self-supervised extraction, confidence-based filtering of tuples, and a quantitative brittleness study are natural next steps.

\section{Implementation Details}
\label{sec:supp:impl}

Step~1, which produces the initial edit-region mask $M_i$, is training-free, combining Florence-2 and Grounding DINO detection, BLIP-2 verification, and SAM mask extraction. Steps~2 and~3 of SGE-Goal fine-tune their respective base editors (FLUX.1~Kontext~[Dev], FLUX.2~[klein]) with LoRA adapters under the rectified-flow objective~\cite{RectifiedFlow2023} to predict the final edit-region mask $M_f$ and synthesize the goal image $I_e$, respectively, augmented with our auxiliary channel-consistency loss (Step~2) and static-scene-consistency loss (Step~3); we use $\lambda_{\text{chan}}{=}25$, $\lambda_{\text{static}}{=}7$, and timestep gate $\tau_{\text{rec}}{=}0.35$. All conditioning inputs (original image, masks) are VAE-encoded as clean latents; only the target latent is noised. After processing, the datasets yield ${\sim}41$k/$22$k/$6.5$k training, ${\sim}1$k/$1$k/$358$ validation, and $2$k/$1.1$k/$376$ test/inference tuples for BridgeDataV2-teleoperated / CALVIN-ABCD / LIBERO-10-90-Combined; splits are assigned by MD5 hashing (seed~42) for reproducibility. Downstream evaluation uses the publicly released GR-MG and GHIL-Glue checkpoints without modification, plus the Octo-ft policy we fine-tune on CALVIN (\cref{sec:supp:octo}). All code\refcode{} (for SGE-Goal, the data-generation pipeline, and the diagnostic evaluator), the processed datasets\refdata, and the model checkpoints\refckpt{} have been publicly released; reproduction requires no proprietary models.

\begin{figure*}[t]
  \centering
  \includegraphics[width=\linewidth]{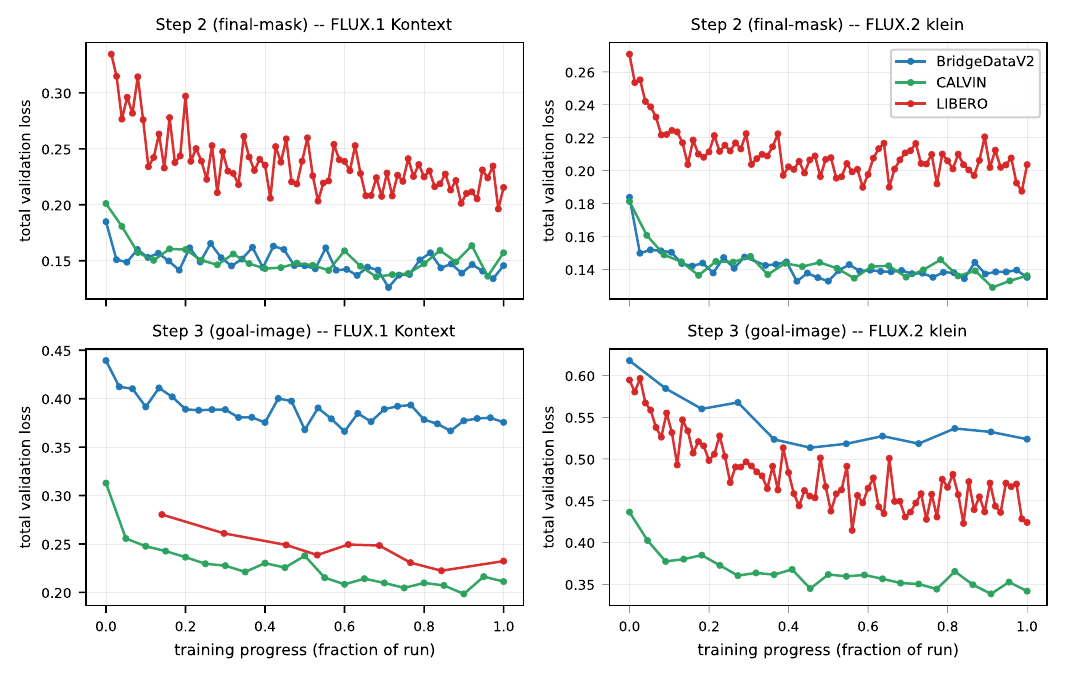}
  \caption{\textbf{Validation loss for SGE-Goal fine-tuning.} Total validation loss---the rectified-flow loss plus the weighted auxiliary consistency loss (channel-consistency, $\lambda{=}25$, for Step~2; static-scene-consistency, $\lambda{=}7$, for Step~3)---on the held-out validation split as fine-tuning progresses, for both fine-tuned stages (Step~2 final-mask predictor, Step~3 goal-image synthesizer) and \emph{both} base editors (FLUX.1~Kontext, FLUX.2~klein), across all three processed datasets. The total validation loss decreases for every run, indicating that the models keep improving on held-out data without overfitting. The $x$-axis is normalized by each run's length; validation uses the val split of each processed dataset.}
  \label{fig:supp-curves}
\end{figure*}

\Cref{tab:supp-hyperparams} lists the full training and inference hyperparameters; this recipe is shared across all fine-tuning runs, apart from the per-editor Step~2 guidance scale and the two locally trained FLUX.1~Kontext LIBERO runs noted in its caption. \Cref{fig:supp-curves} plots the total validation loss (rectified-flow plus weighted auxiliary consistency loss) for both fine-tuned stages and both base editors: it decreases on the held-out validation split of every dataset, confirming stable convergence without overfitting.

\begin{table}[t]
  \caption{\textbf{Full training and inference hyperparameters} for SGE-Goal's LoRA fine-tuning (Steps~2--3). The recipe is shared across all runs; only the listed per-stage, per-editor and per-dataset settings differ, except for the two FLUX.1~Kontext LIBERO runs ($^\dagger$): Step~2 with guidance $3.5$, batch size $43$, no warmup and seed $42$; Step~3 with batch size $72$ ($2{\times}36$) and no warmup. Each run trains for up to $75$ epochs and saves a checkpoint every epoch; the epoch of the checkpoint used is listed at the bottom. Br${=}$BridgeDataV2, Ca${=}$CALVIN, Li${=}$LIBERO; FK${=}$FLUX.1~Kontext, F2K${=}$FLUX.2~klein. Here $\tau_{\text{rec}}$ gates the auxiliary consistency loss to low-noise samples ($\sigma\!<\!\tau_{\text{rec}}$), whose one-step clean estimate decodes reliably; \emph{fraction} is the share of those gated samples per batch on which the (decode-based) auxiliary loss is actually computed ($1.0{=}$all).}
  \label{tab:supp-hyperparams}
  \centering\footnotesize
  \setlength{\tabcolsep}{4pt}
  \begin{tabular}{@{}ll@{}}
    \toprule
    \multicolumn{2}{@{}l}{\emph{Training}} \\
    Base editors & FLUX.1~Kontext~[Dev], FLUX.2~klein \\
    LoRA rank / $\alpha$ / dropout & $16$ / $16$ / $0$ \\
    Optimizer & 8-bit AdamW \\
    Learning rate & $1{\times}10^{-4}$, constant ($200$ warmup) \\
    Batch size & $160$ \\
    Gradient accumulation & $1$ \\
    Epochs & up to $75$ (checkpoint every epoch) \\
    Precision & bf16 (TF32) \\
    Max.\ text tokens & $512$ \\
    Resolution & $256$ (Br, Ca) / $128$ (Li) \\
    Seed & $0$ \\
    Gate $\tau_{\text{rec}}$ / fraction & $0.35$ / $1.0$ \\
    \midrule
    \multicolumn{2}{@{}l}{\emph{Per stage}} \\
    Guidance scale, Step~2 & $4.0$ (F2K) / $1.0$ (FK)$^\dagger$ \\
    Guidance scale, Step~3 & $1.0$ \\
    Step~2 auxiliary loss & channel-consistency ($\lambda{=}25$) \\
    Step~3 auxiliary loss & static-scene-consistency ($\lambda{=}7$) \\
    \midrule
    \multicolumn{2}{@{}l}{\emph{Inference}} \\
    Denoising steps / guidance scale & $20$ / $1.0$ \\
    \midrule
    \multicolumn{2}{@{}l}{\emph{Checkpoint epoch used (Br / Ca / Li)}} \\
    Step~2, FK / F2K & $38$ / $22$ / $56$ \quad $17$ / $22$ / $59$ \\
    Step~3, FK / F2K & $18$ / $18$ / $64$ \quad $9$ / $15$ / $71$ \\
    \bottomrule
  \end{tabular}
\end{table}

\input{sec/supp_ported}

\end{document}

%% file: preamble.tex
\usepackage{graphicx}
\usepackage{amsmath}
\usepackage{amssymb}
\usepackage{booktabs}
\usepackage{multirow}
\usepackage{makecell}
\usepackage{subcaption}
\usepackage{xcolor}
\usepackage{colortbl}
\usepackage{pifont}%
\usepackage{etoolbox}

\definecolor{ourshl}{rgb}{0.90,0.96,0.90}
\newcommand{\oursrow}{\rowcolor{ourshl}}
\newcommand{\best}[1]{\textbf{#1}}

\AtEndEnvironment{figure}{\vspace{-5pt}}             %

\patchcmd{\abstract}{\vspace*{12pt}}{\vspace*{4pt}}{}{}
\patchcmd{\endabstract}{\vspace*{12pt}}{\vspace*{6pt}}{}{}

\patchcmd{\wacvsection}{{7pt}}{{5pt}}{}{}
\patchcmd{\wacvsubsection}{{5pt}}{{4pt}}{}{}

\newcommand{\linkcode}{\href{https://github.com/Riddhi-Chatterjee/SGE-Goal}{\nolinkurl{github.com/Riddhi-Chatterjee/SGE-Goal}}}
\newcommand{\linkdata}{\nolinkurl{huggingface.co/datasets/RiddhiCh/}\{\href{https://huggingface.co/datasets/RiddhiCh/LGRM_BridgeDataV2_Teleoperated}{\nolinkurl{LGRM_BridgeDataV2_Teleoperated}}, \href{https://huggingface.co/datasets/RiddhiCh/Calvin-Dataset-Final}{\nolinkurl{Calvin-Dataset-Final}}, \href{https://huggingface.co/datasets/RiddhiCh/LGRM_Libero_10_90_Combined}{\nolinkurl{LGRM_Libero_10_90_Combined}}\}}
\newcommand{\linkckpt}{\href{https://huggingface.co/RiddhiCh/SGE-Goal}{\nolinkurl{huggingface.co/RiddhiCh/SGE-Goal}}}
\newcommand{\fncode}{\footnote{\linkcode\label{fn:code}}}
\newcommand{\fndata}{\footnote{\raggedright\linkdata\label{fn:data}}}
\newcommand{\fnckpt}{\footnote{\linkckpt\label{fn:ckpt}}}
\newcommand{\refcode}{\textsuperscript{\ref{fn:code}}}
\newcommand{\refdata}{\textsuperscript{\ref{fn:data}}}
\newcommand{\refckpt}{\textsuperscript{\ref{fn:ckpt}}}

%% file: sec/0_abstract.tex
\begin{abstract}
Goal-conditioned manipulation policies are steered by a \emph{goal image}, a picture of the desired post-manipulation scene synthesized from the current observation and a language instruction. The prevailing assumption is that better image-editing quality yields better goals, so goal generators are built and compared using general image-editing metrics. We show this assumption is false. The reality is much more nuanced. Across two independent goal-conditioned policies on CALVIN, the image editors that top aggregate editing-quality leaderboards (Gemini~3 Pro Image and GPT Image 1.5) produce \emph{worse} control targets than our proposed far smaller finetuned open models, and the gap widens sharply for the multi-step (long-horizon) tasks robots actually face. A per-dimension analysis explains why. Downstream policy success is governed almost entirely by manipulation \emph{adherence} (selecting the right object and the right final state; Spearman $\rho\!\approx\!0.9$ with success), while photorealism and aesthetic fidelity, precisely where large proprietary editors excel, are uncorrelated with success ($\rho\!\approx\!0$). This dissociation holds on both CALVIN and the real-world-aligned SIMPLER-Bridge benchmark. We then present \textbf{SGE-Goal}, a spatially-grounded decomposed editing framework that predicts initial and final 2D edit-region masks before goal-image synthesis. SGE-Goal turns open base models (FLUX.1~Kontext, FLUX.2~klein) into the strongest goal generators we evaluate on CALVIN, and second only to Gemini~3 Pro Image on SIMPLER-Bridge. Base models fine-tuned with SGE-Goal also hold up under chaining far better than the best proprietary editor, sustaining roughly seven times the success rate at the longest (five-step) chains. We have released all code\fncode, the processed datasets\fndata, model checkpoints\fnckpt, our automated data-generation pipeline, and a 15-dimension human-aligned diagnostic evaluator. Our results argue that goal generators must be selected by downstream utility, not by editing-quality leaderboards.
\end{abstract}

%% file: sec/1_intro.tex
\section{Introduction}
\label{sec:intro}

\begin{figure}[!t]
  \centering
  \includegraphics[width=1.0\linewidth]{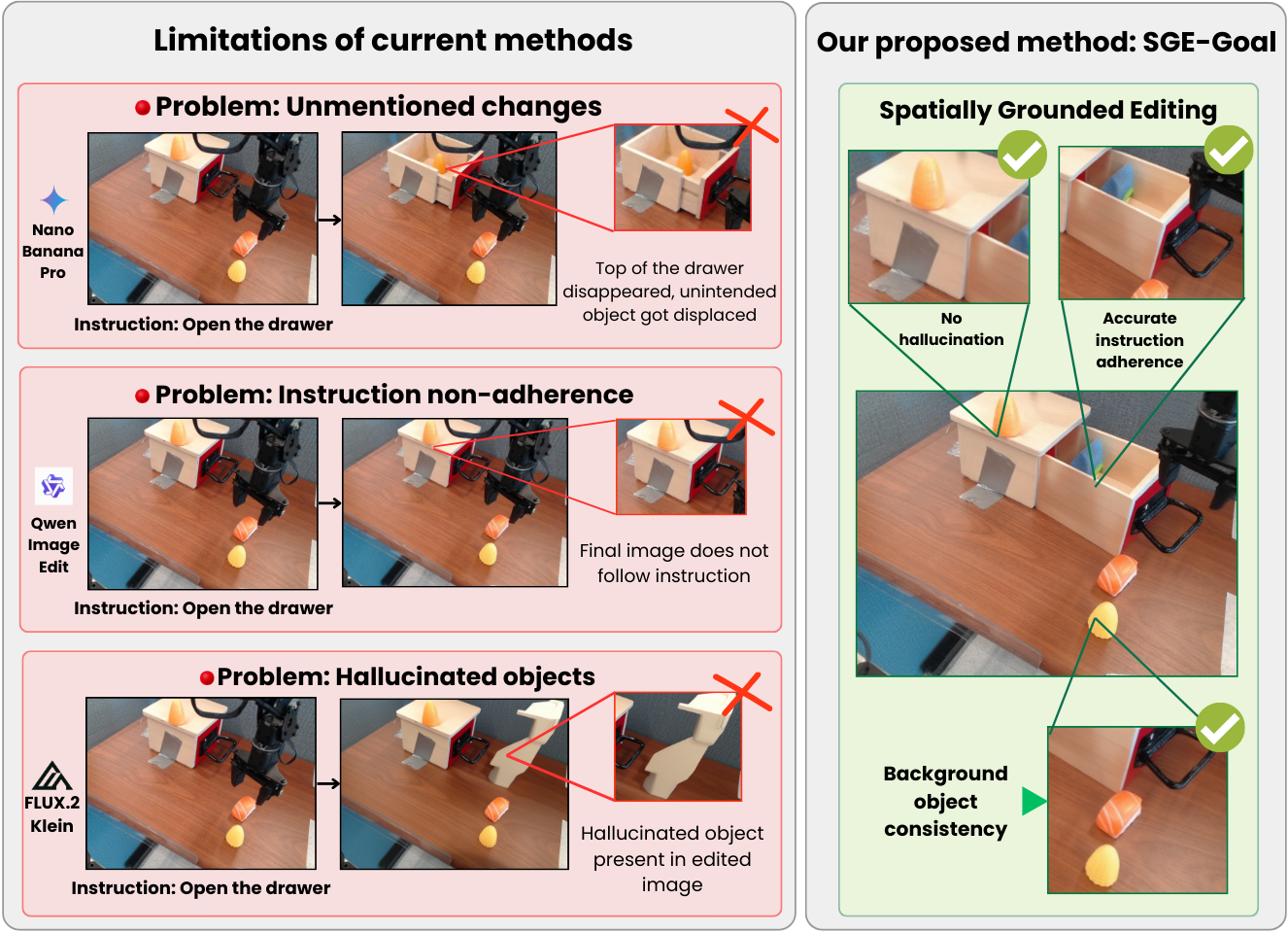}
  \caption{\textbf{Goal images are control targets, not artworks.} General-purpose editors try optimizing for editing quality in general, but for robotic manipulation a goal image must primarily (i)~manipulate the \emph{correct} object to the \emph{correct} final state and (ii)~avoid spurious changes that mislead the policy. SGE-Goal achieves this through explicit spatial grounding, while strong general editors introduce unmentioned changes, miss the instruction, or hallucinate objects.}
  \label{fig:teaser}
\end{figure}

Goal-conditioned policy learning is a powerful paradigm for robotic manipulation~\cite{SuSIE2023,UniPi2023,Clover2024,Seer2024,LfVoid2023}. Instead of acting from language alone, the robot is given a \emph{goal image}: a picture of the desired post-manipulation scene that turns an ambiguous instruction into an explicit, geometric target. Conditioning a policy on a visual goal together with language has been shown to yield substantially higher manipulation success than language alone~\cite{MUTEX2023}. Because the policy cannot observe the future, this goal image must be \emph{synthesized} from the current observation and the instruction, an image-editing task. Goal-image quality is therefore decisive: a faithful goal gives the policy a clear target, while an unfaithful one misleads it from the first step.

A natural and now-standard methodology has emerged: generate goal images with a strong language-guided image editor, and compare candidate generators using image-editing quality metrics and benchmarks~\cite{I2IBench2025,DICE2025,PICABench2024}. The implicit assumption is that \emph{a better quality edit is a better goal}. Modern proprietary editors such as Gemini~3 Pro Image and GPT Image~1.5 produce strikingly photorealistic edits and dominate such benchmarks, suggesting that the goal-image synthesis problem is essentially solved by scale.

\begin{figure}[t]
  \centering
  \includegraphics[width=\linewidth]{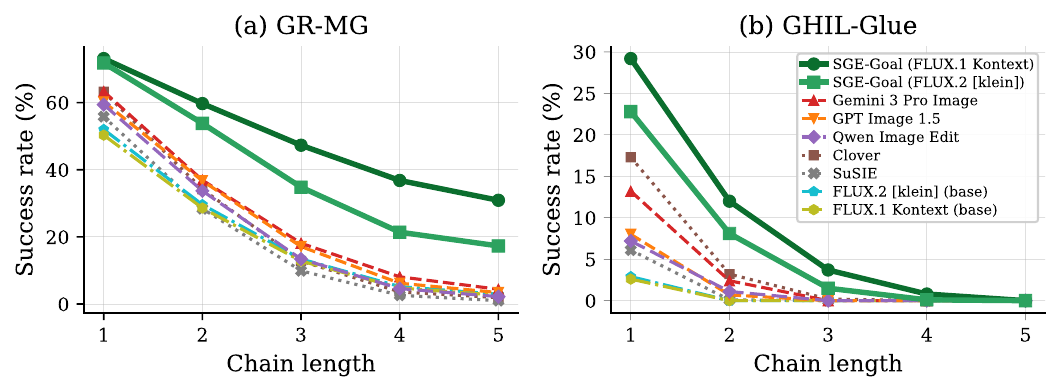}
  \caption{\textbf{Downstream policy success vs.\ chain length} for two pre-trained policies on CALVIN. SGE-Goal goal images (green) yield the best success rates at every chain length and retain performance as chain length increases, while the strongest general editors collapse: at chain length~5, SGE-Goal (FLUX.1~Kontext~[Dev]) reaches $30.9\%$ vs.\ Gemini~3 Pro's $4.4\%$ under GR-MG. Long-horizon manipulation is where goal-image utility matters most. Values are previewed here and tabulated in full in \cref{tab:downstream}; base${=}$untuned editor.}
  \label{fig:chaining}
\end{figure}

\noindent\textbf{This assumption is false. The reality is more nuanced, and that is our starting point.}
We evaluate goal images not by how they look but by what they are \emph{for}: the success of a downstream manipulation policy conditioned on them. Across two independent, pre-trained goal-conditioned policies (GR-MG~\cite{GRMG2025} and GHIL-Glue~\cite{GHILGlue2025}) on CALVIN~\cite{CALVIN2022}, we find that the editors topping aggregate image-editing quality (Gemini~3 Pro Image and GPT Image 1.5) yield \emph{worse} downstream policy success rates than our proposed far smaller finetuned open models, and the gap \emph{widens} for the multi-step tasks robots actually perform (\cref{fig:chaining}). The two policies agree on the method ordering almost perfectly (Spearman $\rho\!=\!0.98$), so this is not a quirk of one policy.

\noindent\textbf{Why does editing quality mislead?}
A fine-grained, 15-dimension analysis of every goal image reveals a clean \emph{dissociation}, which we examine in detail in \cref{sec:exp:dissociation}. Manipulation \textbf{adherence}, selecting the correct object and reaching the correct final state, is the dominant predictor of policy success ($\rho\!\approx\!0.9$), whereas \textbf{photorealism} and aesthetic fidelity, the dimensions that most impress human viewers and on which proprietary editors win, are essentially uncorrelated with it ($\rho\!\approx\!0$). A goal image is a control signal, not an artwork; a policy rewards a target that points at the right object in the right place, free of spurious changes, not visual polish.

Guided by this finding, we present \textbf{SGE-Goal}, a spatially-grounded, decomposed editing framework designed to maximize adherence and suppress the spurious changes that mislead policies. Rather than mapping instruction-plus-image directly to an edited image, SGE-Goal first predicts where the manipulated objects \emph{are} and where they \emph{will be}---an initial and a final 2D edit-region mask---and then synthesizes the goal conditioned on both (\cref{fig:teaser}). This decomposition disentangles spatial reasoning from appearance synthesis and prevents the ``shortcuts'' (background drift, hallucination) that entangled end-to-end editors take. Crucially, SGE-Goal is a lightweight fine-tuning recipe: it turns ordinary open base models (FLUX.1~Kontext~\cite{FluxKontext2024}, FLUX.2~klein~\cite{FLUX2klein2025}) into the strongest goal generators we evaluate on CALVIN, while running fully on-premise.

To train SGE-Goal without manual labels, we introduce a \textbf{fully automated data-generation pipeline} that extracts spatially-grounded training tuples from raw robot videos using a chain of vision foundation models. To diagnose \emph{why} methods differ, we build a \textbf{15-dimension, human-aligned evaluation framework} ($91.7\%$ agreement with human judgment) whose granularity is what enables the dissociation analysis of \cref{sec:exp:dissociation}. We stress that we do not propose this, or any, image-editing metric as a substitute for downstream evaluation; it is a diagnostic that explains downstream behavior.

\vspace{0.15cm}
\noindent\textbf{Contributions.}
\begin{itemize}\itemsep0.15em
  \item We propose \textbf{SGE-Goal}, a spatially-grounded decomposed editing framework that turns small open base models into the strongest goal generators we evaluate on CALVIN and second only to Gemini~3 Pro Image on the real-world-aligned SIMPLER-Bridge benchmark. Base models fine-tuned with SGE-Goal also hold up under chaining far better than the best proprietary editor.
  \item We have released code\refcode, the processed datasets\refdata, model checkpoints\refckpt, an automated data-generation pipeline, and a 15-dimension human-aligned diagnostic evaluator.
  \item Using our own closed-loop setup, we confirm that visual-goal conditioning beats text-only conditioning at \emph{every} chain length ($1$--$5$), extending the single-step finding of MUTEX~\cite{MUTEX2023} to the long-horizon regime and motivating the pursuit of high-quality visual goals.
  \item We show empirically that image-editing quality is a \emph{misaligned proxy} for goal-image utility in robotic manipulation. On both CALVIN and SIMPLER-Bridge, the editing-quality ranking of generators fails to match their downstream success-rate ranking, so goal generators must be selected by downstream success rather than by editing-quality leaderboards.
  \item We identify and validate the mechanism, an adherence/photorealism \emph{dissociation} in which adherence predicts policy success and photorealism does not, and we show it holds on both CALVIN and SIMPLER-Bridge.
\end{itemize}

%% file: sec/2_related.tex
\section{Related work}
\label{sec:related}

\noindent\textbf{Goal-image synthesis for manipulation.}
Visual goals provide an information-rich interface between language and low-level control~\cite{RT1,RT2,Octo2024,OpenVLA2024}. SuSIE~\cite{SuSIE2023} pioneered subgoal synthesis by fine-tuning InstructPix2Pix~\cite{InstructPix2Pix2023} on robot video, but predicts only near-future subgoals whose quality degrades rapidly when chained toward an end goal. UniPi~\cite{UniPi2023} and Clover~\cite{Clover2024} generate goal/subgoal frames via text-conditioned video generation and closed-loop re-planning, while Seer~\cite{Seer2024} uses predictive inverse-dynamics models; all are prone to hallucinated objects, confusion between similar objects, and incoherent or drifting predictions. LfVoid~\cite{LfVoid2023} applies pre-trained text-to-image models without task-specific fine-tuning and frequently departs from the original scene. A limitation common to these methods is the lack of explicit \emph{spatial} control over where objects move and explicit \emph{consistency} control over what should not change. SGE-Goal addresses both via mask-based spatial grounding.

\noindent\textbf{Goal-conditioned policies.}
Our evaluation conditions pre-trained policies on synthesized goals. GR-MG~\cite{GRMG2025} is a multi-modal goal-conditioned policy leveraging partially-annotated data; GHIL-Glue~\cite{GHILGlue2025} hierarchically filters subgoal images to suppress off-task drift; MUTEX~\cite{MUTEX2023} learns unified policies from multimodal task specifications and reports that visual goals improve success over text-only conditioning. We use GR-MG and GHIL-Glue as fixed downstream evaluators of goal-image utility and corroborate the visual-vs-text trend reported by MUTEX.

\noindent\textbf{Language-guided image editing.}
Instruction-based editing has advanced rapidly with diffusion models~\cite{StableDiffusion2022,DDPM2020}. InstructPix2Pix~\cite{InstructPix2Pix2023} and HIVE~\cite{HIVE2023} follow natural-language edits; inversion methods such as Prompt-to-Prompt~\cite{PromptToPrompt2023} manipulate cross-attention; localized methods such as LIME~\cite{LIME2025} restrict edits to a region of interest. These methods model the object's \emph{initial} location but not where it ends up after manipulation, making them ill-suited to displacement. Recent open editors (Qwen Image Edit~\cite{QwenImageEdit2024}, FLUX.1~Kontext~\cite{FluxKontext2024}, FLUX.2~klein~\cite{FLUX2klein2025}, BAGEL~\cite{BAGEL2024}) and proprietary systems (Gemini~3 Pro Image, GPT Image~1.5) offer strong general-purpose editing; we use them both as base models for fine-tuning and as strong baselines, and show that their general strength does not translate into manipulation utility.

\noindent\textbf{Evaluating edits and the missing link to control.}
Classical metrics (PSNR, LPIPS~\cite{LPIPS2018}, CLIP score~\cite{CLIP2021}) give limited semantic insight, while recent benchmarks add fine-grained editing dimensions (I2I-Bench~\cite{I2IBench2025}), operation detection (DICE~\cite{DICE2025}), physical realism (PICABench~\cite{PICABench2024}), or LLM judges (RISEBench~\cite{RISEBench2024}). Two gaps matter for robotics: none is \emph{curated} for manipulation (they weight photorealism heavily---dimensions we show are uncorrelated with policy success), and none is \emph{validated against downstream control}. We provide a manipulation-specific, human-aligned diagnostic and, more importantly, measure how each editing dimension relates to actual policy success, exposing the proxy gap.

%% file: sec/3_method.tex
\section{Method}
\label{sec:method}

We describe three components: a fully automated pipeline that extracts spatially-grounded training data from robot videos (\cref{sec:method:data}); SGE-Goal, a three-step decomposed editing framework trained on this data (\cref{sec:method:edit}); and a 15-dimension diagnostic evaluator (\cref{sec:method:eval}).

\subsection{Automated data extraction}
\label{sec:method:data}

\begin{figure}[t]
  \centering
  \includegraphics[width=1.0\linewidth]{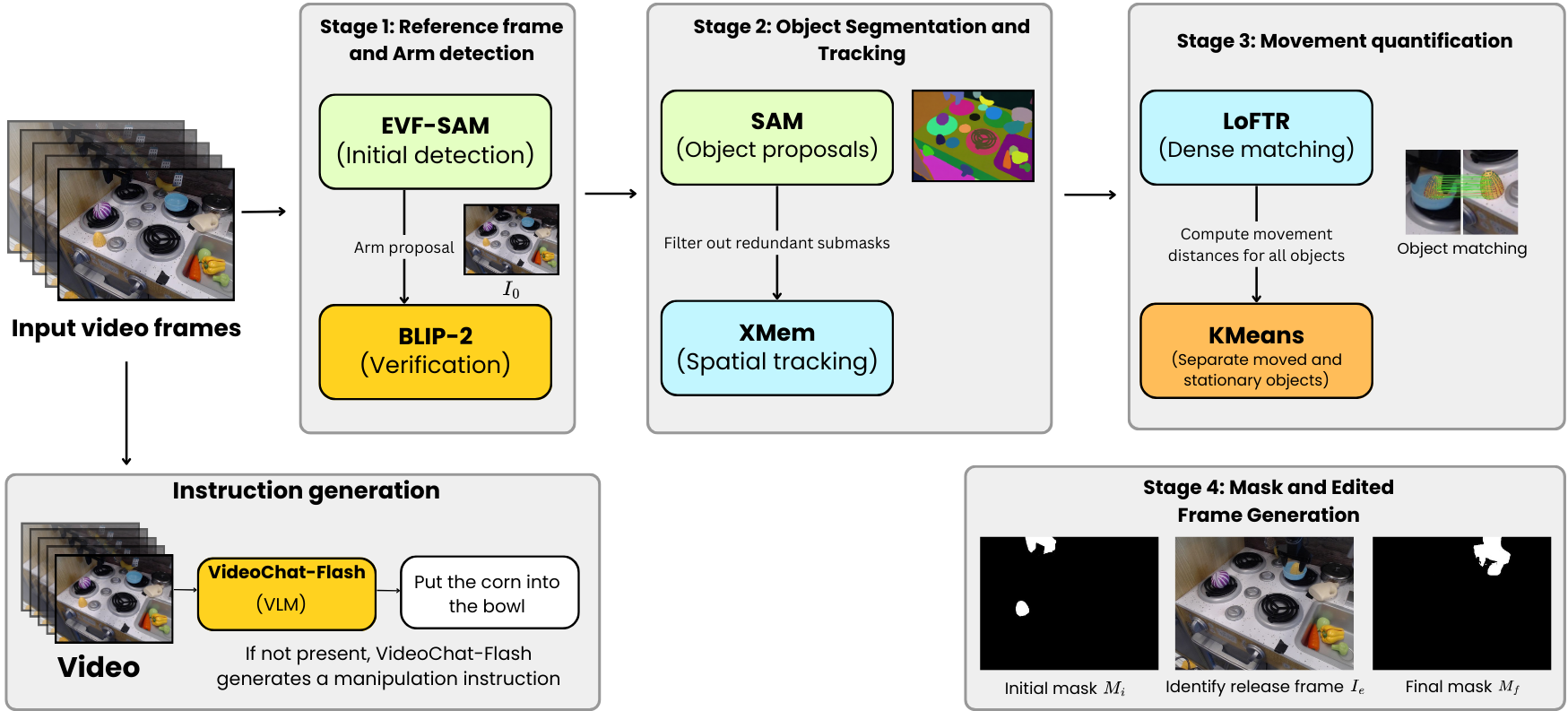}
  \caption{Automated data-extraction pipeline. Each robot video is processed by a chain of vision foundation models to produce spatially-grounded training tuples $(I_o, I_e, M_i, M_f, T)$ without manual annotation.}
  \label{fig:data}
\end{figure}

Each training sample is a tuple $(I_o, I_e, M_i, M_f, T)$: the \emph{original image} $I_o$ (before manipulation), the \emph{edited image} $I_e$ (after manipulation), the \emph{initial edit-region mask} $M_i$ localizing the dynamic (to-be-manipulated) objects in $I_o$, the \emph{final edit-region mask} $M_f$ marking their post-manipulation locations in $I_e$, and the \emph{instruction} $T$. The masks encode the before/after spatial state of the manipulated objects. We extract these tuples from raw robot videos with no manual annotation, chaining six vision foundation models (\cref{fig:data}):
\textbf{(1)~Robotic-arm masking:} EVF-SAM~\cite{EVFSAM2024} proposes per-frame arm masks, verified by BLIP-2~\cite{BLIP2_2023} VQA; a multi-pass strategy selects the first frame with consistent arm detection as $I_o$.
\textbf{(2)~Object segmentation \& tracking:} SAM~\cite{SAM2023} produces class-agnostic masks on $I_o$, filtered and merged, then propagated through the video with XMem~\cite{XMem2022}.
\textbf{(3)~Dynamic-object identification:} per-object motion is quantified by LoFTR~\cite{LoFTR2021} feature matching (centroid distance as fallback); $K{=}2$ KMeans on these distances separates \emph{dynamic} from \emph{stationary} objects.
\textbf{(4)~Mask and edited-frame extraction:} $M_i$ is the union of arm and dynamic-object masks on $I_o$; the edited frame is detected when arm and dynamic-object masks cease to overlap (release), giving $I_e$ and the corresponding $M_f$.
\textbf{(5)~Instruction generation:} for datasets lacking text, $T$ is generated by a video VLM~\cite{VideoChatFlash2024}.

\noindent\textbf{Composite vs.\ atomic manipulations.}
We distinguish \emph{composite} instructions, which decompose into multiple necessary steps (e.g., ``open the drawer and put the apple in it''), from \emph{atomic} ones, which cannot be decomposed without introducing unnecessary (hallucinated) steps; the distinction is environment-dependent. SGE-Goal operates at the atomic level: composite instructions are decomposed into their necessary-and-sufficient atomic constituents, each processed in sequence with the generated goal $\hat{I}_e$ becoming the next step's $I_o$. Extraction is performed at atomic granularity.

\subsection{Decomposed goal-image synthesis}
\label{sec:method:edit}

\begin{figure}[t]
  \centering
  \includegraphics[width=1.0\linewidth]{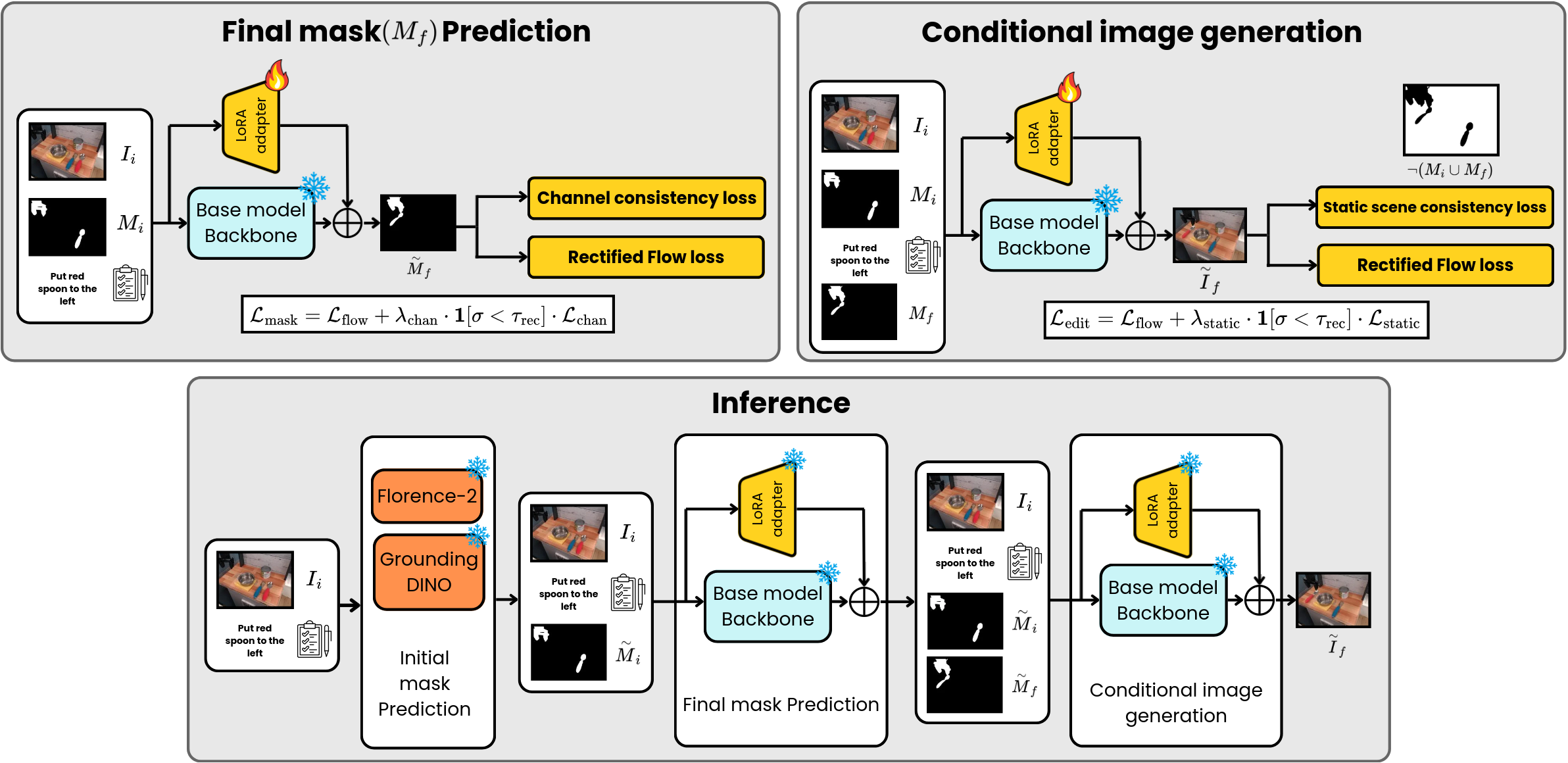}
  \caption{SGE-Goal synthesizes the goal image in three steps with explicit 2D spatial grounding via the initial mask $M_i$ and final mask $M_f$.}
  \label{fig:pipeline}
\end{figure}

\noindent\textbf{What we mean by ``spatial grounding.''}
SGE-Goal grounds editing in the \emph{2D image plane} via binary edit-region masks; it does not estimate 3D geometry, depth, or camera pose. The masks act as an explicit spatial prior that is absent in end-to-end editors, specifying \emph{where} objects are and \emph{where they should go}. We make this scope explicit to avoid overclaiming: ``grounding'' here is 2D localization of the edit, not metric 3D reasoning.

Given $(I_o, T)$, the pipeline sequentially predicts $M_i$, then $M_f$, then the goal image $\hat{I}_e$ (\cref{fig:pipeline}).

\noindent\textbf{Step 1: Initial mask $M_i$ (training-free).}
We localize the dynamic target objects in $I_o$ with an ensemble of pre-trained models and \emph{no} fine-tuning. The instruction is parsed for target-object names; candidate boxes come from Florence-2~\cite{Florence2_2024} and Grounding DINO~\cite{GroundingDINO2024} with progressively relaxed thresholds and per-object NMS; each candidate is verified by BLIP-2~\cite{BLIP2_2023} VQA, and survivors prompt SAM~\cite{SAM2023} to produce a pixel-level $M_i$.

\noindent\textbf{Step 2: Final mask $M_f$.}
We fine-tune a base image-to-image model with LoRA to predict where dynamic objects reside \emph{after} manipulation, with input $(I_o, T, M_i)$. Inputs are VAE-encoded; the original-image and $M_i$ latents are clean conditioning, while the target $M_f$ latent is noised by the rectified-flow schedule $M_f^t = (1-\sigma)M_f^0 + \sigma\epsilon$, $\epsilon\!\sim\!\mathcal{N}(0,I)$. The model predicts the velocity via the rectified-flow loss~\cite{RectifiedFlow2023}:
\begin{equation}
  \mathcal{L}_{\text{flow}} = \mathbb{E}_{\sigma,\epsilon}\!\left[\,\|\,v_\theta(M_f^t,\sigma) - (\epsilon - M_f^0)\,\|^2\right].
  \label{eq:flow}
\end{equation}
From this velocity we recover the one-step estimate of the clean mask latent, $\hat{M}_f^0 = \epsilon - v_\theta(M_f^t,\sigma)$. Since $\hat{M}_f$ is binary (identical RGB channels), an auxiliary \emph{channel-consistency} loss is applied when $\sigma\!<\!\tau_{\text{rec}}$ (so the one-step estimate decodes cleanly): decoding $\hat{M}_f^0$ to $\hat{M}_f^{\text{img}}$ and summing over channel pairs $c\!<\!c'$ from $\{R,G,B\}$,
\begin{equation}
  \mathcal{L}_{\text{chan}} = \frac{1}{HW}\sum_{h,w}\sum_{c<c'}\big(\hat{M}_f^{c}-\hat{M}_f^{c'}\big)^2 .
  \label{eq:chan}
\end{equation}
The objective is $\mathcal{L}_{\text{mask}} = \mathcal{L}_{\text{flow}} + \lambda_{\text{chan}}\mathbf{1}[\sigma\!<\!\tau_{\text{rec}}]\mathcal{L}_{\text{chan}}$ with $\lambda_{\text{chan}}{=}25$, $\tau_{\text{rec}}{=}0.35$ (swept on validation data).

\noindent\textbf{Step 3: Goal image $\hat{I}_e$.}
A second LoRA-fine-tuned model synthesizes the goal from $(I_o, T, M_i, M_f)$. The original, $M_i$, and $M_f$ latents are clean conditioning; the target $I_e$ latent is noised as above and trained with $\mathcal{L}_{\text{flow}}$ (\cref{eq:flow}, with $I_e$ for $M_f$). To preserve static regions, an auxiliary \emph{static-scene-consistency} loss penalizes deviations outside the union mask $M_\cup = \max(M_i, M_f)$ (gated by $\sigma\!<\!\tau_{\text{rec}}$):
\begin{equation}
  \mathcal{L}_{\text{static}} = \tfrac{1}{CHW}\,\big\|\,(\hat{I}_e^{\text{img}} - I_o^{\text{img}})\odot(1-\bar{M}_\cup)\,\big\|^2,
  \label{eq:static}
\end{equation}
where the superscript $^{\text{img}}$ denotes the VAE-decoded pixel-space image (the one-step estimate and the original are both decoded so the loss is computed in pixel space), and $\bar{M}_\cup$ is the channel-mean of the decoded union mask $M_\cup$ (gradient not propagated through the mask). The objective is $\mathcal{L}_{\text{edit}} = \mathcal{L}_{\text{flow}} + \lambda_{\text{static}}\mathbf{1}[\sigma\!<\!\tau_{\text{rec}}]\mathcal{L}_{\text{static}}$ with $\lambda_{\text{static}}{=}7$, $\tau_{\text{rec}}{=}0.35$ (swept on validation data).

\noindent\textbf{Why decomposition helps.}
The decomposition turns one entangled task (generate $I_e$ from $I_o,T$) into disentangled sub-tasks: foundation-model grounding (Step~1), pure spatial reasoning (Step~2), and appearance synthesis given explicit spatial guidance (Step~3). End-to-end editors must learn spatial reasoning and synthesis jointly and, lacking explicit spatial signals, take shortcuts---editing in place or drifting in the background. As a result, SGE-Goal concentrates editing quality on the manipulation-relevant axes---adherence and the absence of spurious changes---that downstream policies actually consume (\cref{sec:exp:dissociation}).

\subsection{A 15-dimension diagnostic evaluator}
\label{sec:method:eval}

To analyze \emph{why} goal generators differ, we use a fully automated evaluator scoring each triplet $(I_o, I_e, T)$ along 15 manipulation-specific dimensions, grouped into \textbf{adherence} (R14 source, R15 destination), \textbf{object preservation} (R1--R4: stationary, moved, removed, hallucinated objects), \textbf{photorealism} (R13), and \textbf{scene artifacts} (R5--R12: color/texture/shape, lighting, camera), using only open models (a video VLM~\cite{VideoChatFlash2024}, SAM, XMem, BLIP-2, LoFTR, Grounding DINO, Qwen2.5~\cite{Qwen2.5_2024}); full per-dimension definitions and the verbatim VLM prompts are detailed in the supplement (\cref{sec:appendix:methods,sec:appendix:vlm-prompts}). We aggregate the 15 dimensions into a single score by a relevance-tiered weighted average of per-method ranks, splitting each tier's budget equally among its dimensions: adherence $60\%$ ($30\%$ each to R14/R15), object preservation $24\%$, photorealism $8\%$, and scene artifacts $8\%$. On 278 human-annotated triplets it attains $91.7\%$ agreement with human judgment---well above the strongest general-purpose editing benchmark, I2I-Bench~\cite{I2IBench2025} ($71.2\%$; \cref{tab:human-align}). The gap reflects domain curation: where I2I-Bench scores generic quality with a single vision-language judge, our framework adds object-level tooling (SAM/XMem tracking, LoFTR motion, $2$-means clustering) to explicitly count the moved, removed, and hallucinated objects that destabilize policies. We use it strictly as a \emph{diagnostic} that decomposes goal quality into interpretable axes, not as a leaderboard we claim supersedes downstream evaluation (\cref{sec:exp:dissociation}).

\begin{table}[t]
  \caption{Agreement with human judgment (\%, $\uparrow$ higher is better). Our evaluator is far more human-aligned than existing automated frameworks; this makes it a trustworthy diagnostic, though downstream policy success remains the ground truth for utility.}
  \label{tab:human-align}
  \centering\footnotesize
  \setlength{\tabcolsep}{2.3pt}
  \begin{tabular}{@{}lcccccc@{}}
    \toprule
    & Adher. & Obj.\ Pres. & Sc.\ Art. & Realism & Avg. & I2I-Bench \\
    & (R14--15) & (R1--4) & (R5--12) & (R13) & (R1--15) & \\
    \midrule
    Align.\ (\%) & 85.5 & 90.3 & 92.8 & 100.0 & \textbf{91.7} & 71.2 \\
    \bottomrule
  \end{tabular}
\end{table}

%% file: sec/4_experiments.tex
\section{Experiments}
\label{sec:exp}

\subsection{Setup}
\label{sec:exp:setup}

\noindent\textbf{Datasets.}
We process BridgeDataV2 (teleoperated), CALVIN-ABCD~\cite{CALVIN2022}, and a combined LIBERO-10/90~\cite{LIBERO2024} with our pipeline (\cref{sec:method:data}), yielding ${\sim}41$k/$22$k/$6.5$k training and $2$k/$1.1$k/$376$ test tuples respectively. Long-horizon (composite) samples retain their atomic decomposition for chained evaluation.

\noindent\textbf{Base models and baselines.}
We fine-tune two open base editors with SGE-Goal: FLUX.1~Kontext~[Dev]~\cite{FluxKontext2024} and FLUX.2~[klein]~\cite{FLUX2klein2025}. We compare against their untuned bases; the open editor Qwen Image Edit~\cite{QwenImageEdit2024}; manipulation-specific generators SuSIE~\cite{SuSIE2023} and Clover~\cite{Clover2024}; and proprietary editors GPT Image~1.5 and Gemini~3 Pro Image (Nano Banana Pro).

\noindent\textbf{Downstream protocol (primary).}
Our central metric is downstream policy success. We condition two pre-trained, frozen policies: GR-MG~\cite{GRMG2025} and GHIL-Glue~\cite{GHILGlue2025}, on each method's goal images in CALVIN and report success rate (SR\%) vs.\ \emph{chain length} (CL, the number of chained atomic steps, $1$--$5$); holding the policy fixed and varying only the goal generator isolates goal-image utility. As an oracle we also condition on the ground-truth goal (CL${=}1$ only). Both policies were trained for multi-subgoal conditioning; driving them with a single goal per atomic step matches SGE-Goal's design and yields a fair \emph{relative} comparison.

\noindent\textbf{Real-world-aligned benchmark.}
To test whether our conclusions hold beyond CALVIN, we additionally evaluate on \textbf{SIMPLER-Bridge}, the WidowX/BridgeData-V2 setup of SIMPLER~\cite{SIMPLER2024}. SIMPLER is a simulator built as a proxy for real-robot evaluation. Across over $1{,}500$ paired simulated-and-real rollouts its success rates correlate strongly with physical-robot success, ranking BridgeData-V2 policies correctly on all but one task, which makes it one of the most reliable and widely used substitutes for costly real-robot rollouts when evaluating manipulation policies. We drive the frozen GHIL-Glue policy with each method's goal image on four atomic WidowX tasks ($24$ episodes each) and report early-stop reached-success SR. GHIL-Glue is the low-level policy of the SuSIE system, so SuSIE's goals are maximally in-distribution here, a home-field advantage we account for when reading \cref{tab:simpler}.

\noindent\textbf{Editing-quality protocol (diagnostic).}
For analysis we also score every goal image with our 15-dimension evaluator (\cref{sec:method:eval}) and with I2I-Bench~\cite{I2IBench2025}. For our evaluator we report the relevance-tiered weighted average of per-dimension ranks defined in \cref{sec:method:eval} (lower is better). For I2I-Bench we report its weighted dimension score (higher is better; its tier weighting is detailed in the supplement, \cref{sec:appendix:existing-eval}).

\noindent\textbf{Implementation \& reproducibility.}
Steps~2 and~3 of SGE-Goal are LoRA fine-tunes of the respective base editors under the rectified-flow objective with our auxiliary consistency losses (\cref{sec:method:edit}); Step~1 is training-free. Full training and inference hyperparameters and per-step validation/test curves are in the supplement. Code\refcode, model checkpoints\refckpt, the evaluation framework, and the automatically processed datasets\refdata{} are publicly released; no proprietary model is required to reproduce our pipeline or evaluator.

\begin{figure}[t]
  \centering
  \includegraphics[width=\linewidth]{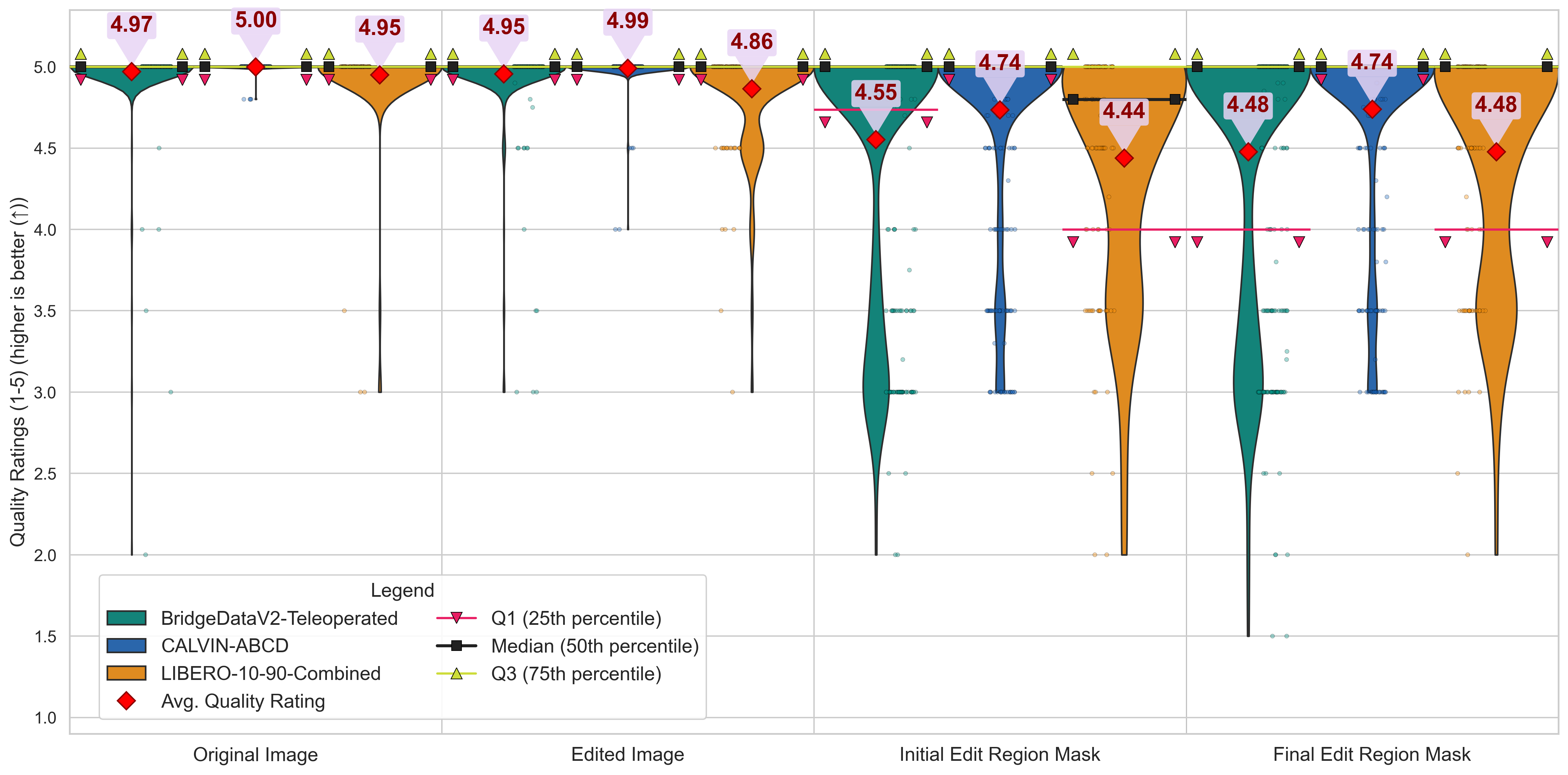}
  \caption{\textbf{Automatically extracted training data is high quality.} Human quality ratings ($1$--$5$) for the four components of each extracted tuple: original image $I_o$, edited image $I_e$, initial mask $M_i$, final mask $M_f$, across the three datasets ($50$ annotators, $500$ tuples/dataset). Diamonds mark per-component means; the lowest is $4.44/5$.}
  \label{fig:dataquality}
\end{figure}

\noindent\textbf{Training-data quality.}
Because SGE-Goal trains on automatically extracted tuples, we first verify the pipeline. Fifty annotators rated $500$ tuples per dataset, scoring each component $(I_o, I_e, M_i, M_f)$ on a $1$--$5$ scale (\cref{fig:dataquality}). Ratings concentrate at the top on all three datasets (lowest mean $4.44/5$); since identical thresholds span real and simulated, single- and multi-camera settings, the pipeline generalizes without per-environment tuning.

\subsection{Downstream policy success}
\label{sec:exp:downstream}

\input{sec/tab_downstream}

\Cref{tab:downstream} and \cref{fig:chaining} report our primary result, measured on CALVIN with the two frozen policies. \textbf{SGE-Goal goal images yield the best downstream policies under both GR-MG and GHIL-Glue, at every chain length.} Under GR-MG at CL${=}1$, SGE-Goal (FLUX.1~Kontext) reaches $73.1\%$, essentially matching the ground-truth oracle ($73.6\%$) and improving its untuned base by $+22.8$ points ($50.3\!\to\!73.1$); the synthesized goals are near-indistinguishable from real ones for the policy. Critically, it surpasses every proprietary and specialized baseline, including Gemini~3 Pro Image ($63.3\%$) by $+9.8$ points. The two policies rank the generators near-identically (the Spearman correlation between their per-generator success rates is $\rho{=}0.98$), so the ordering is a property of the goal images, not of one policy.

\begin{figure}[t]
  \centering
  \includegraphics[width=\linewidth]{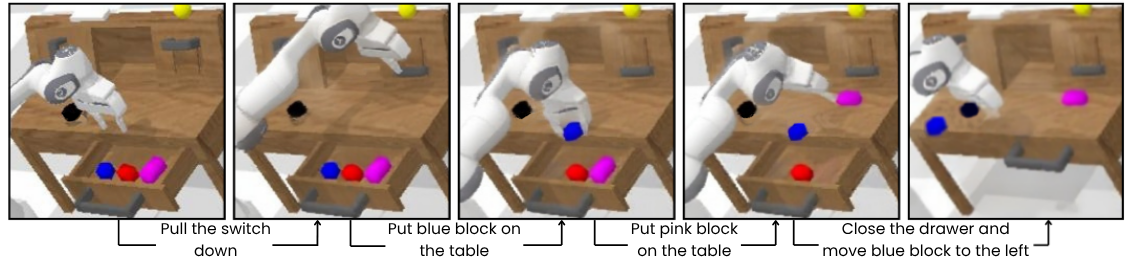}
  \caption{\textbf{Chained multi-step prediction by SGE-Goal.} The generated goal at each step becomes the input for the next. Scene structure is preserved across all five steps and only the instructed objects move, limiting the drift that destroys long-horizon performance for entangled editors.}
  \label{fig:chained}
\end{figure}

\noindent\textbf{Chaining robustness is decisive for long-horizon tasks.}
Real manipulations are multi-step, so the right operating point is CL${>}1$, where the picture is not close (\cref{fig:chaining}). From CL${=}1$ to CL${=}5$, SGE-Goal (FLUX.1~Kontext) falls $73.1\!\to\!30.9\%$ under GR-MG, whereas Gemini collapses $63.3\!\to\!4.4\%$ and SuSIE $55.8\!\to\!1.0\%$, so at CL${=}5$ SGE-Goal delivers a $7\times$ higher success rate than the best proprietary editor. Because SGE-Goal changes only the spatially-grounded edit region, errors compound slowly, whereas entangled editors accumulate spurious changes and background drift that compound catastrophically (\cref{fig:chained}). This graceful degradation, not a marginal CL${=}1$ gap, is the core practical advantage.

\input{sec/tab_simpler}

\noindent\textbf{The same picture holds on real-world-aligned SIMPLER-Bridge.}
\Cref{tab:simpler} reports atomic success under GHIL-Glue. SGE-Goal is the strongest open generator on both base models (F2K $10.4$, FK $8.3$) and second overall, behind only Gemini~3 Pro Image ($12.5$); both variants surpass SuSIE ($7.3$) despite SuSIE's in-distribution advantage. On this real-world-aligned benchmark SGE-Goal is therefore near state of the art rather than ahead of every proprietary system. The remaining gap to Gemini likely reflects the richer textures and lighting of real BridgeData, where its web-scale pretraining gives stronger priors for organic, real-world objects (e.g., carrots). Still, the operative conclusion holds: a small open editor fine-tuned with SGE-Goal produces goal images that drive real-robot-relevant manipulation as well as or better than far larger proprietary editors.

\input{sec/tab_vt}
\noindent\textbf{Visual goals beat text goals at all chain lengths.}
Conditioning GR-MG on an SGE-Goal image plus text dominates text-only conditioning at every chain length (\cref{tab:vt}): $73.1$ vs.\ $59.0\%$ at CL${=}1$, widening to $30.9$ vs.\ $0.2\%$ at CL${=}5$. This verifies, in our own closed-loop setup, the visual-goal benefit that MUTEX~\cite{MUTEX2023} reports for single-step LIBERO-100, and \emph{extends} it across the long-horizon regime, justifying our focus on producing the most useful \emph{visual} goals.

\input{sec/tab_runtime}
\noindent\textbf{Cost.}
End-to-end inference takes ${\sim}10.6$\,s per step on one A10G GPU, entirely on open models (\cref{tab:runtime}); data generation is a one-time, offline cost amenable to parallelization.

\subsection{Image editing evaluation results}
\label{sec:exp:editing}

\subsubsection{Editing-quality comparison}

\begin{figure}[t]
  \centering
  \includegraphics[width=\linewidth]{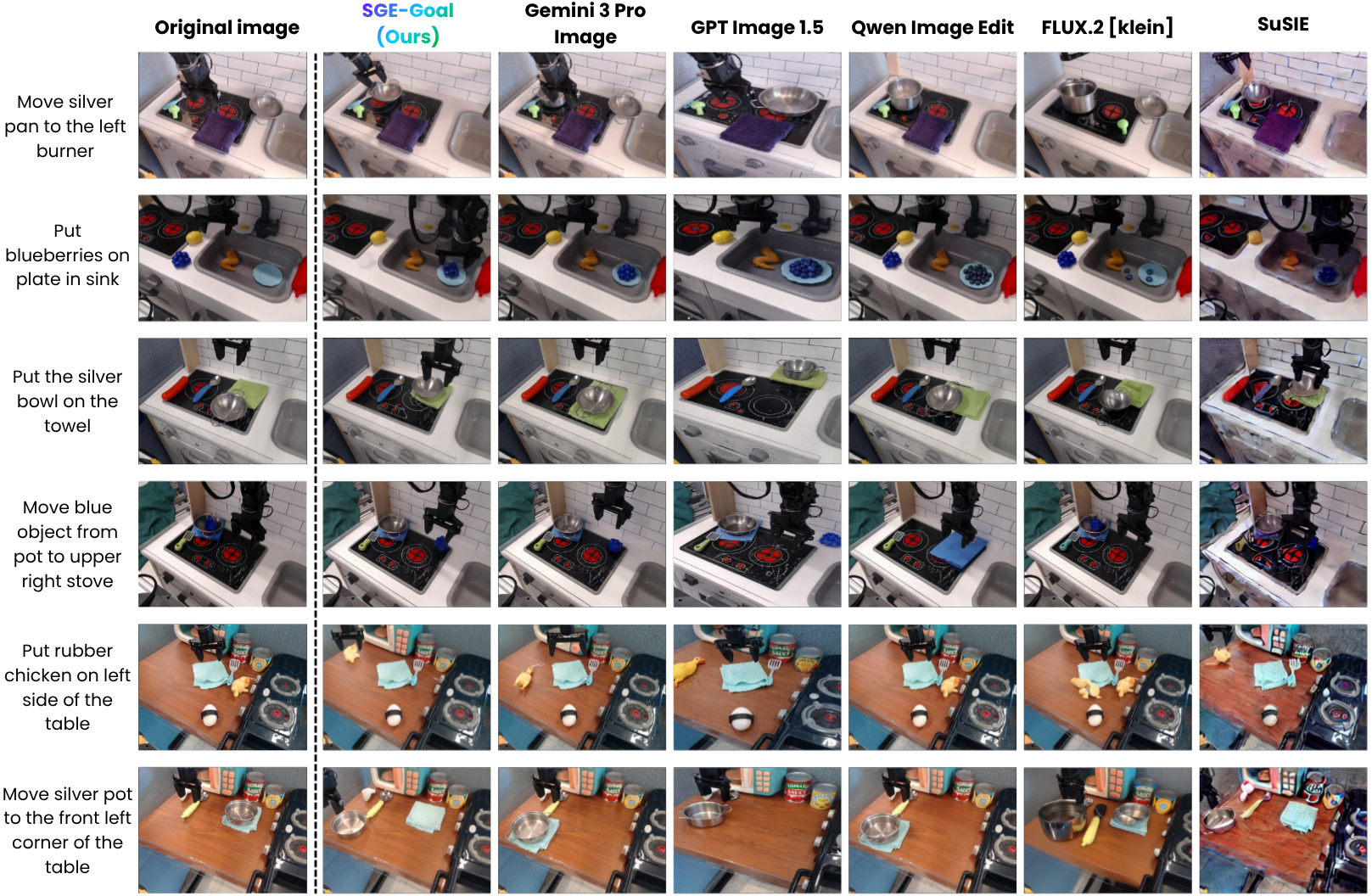}
  \caption{\textbf{Qualitative goal-image comparison.} For each instruction, the goal image produced by SGE-Goal (Ours, FLUX.1~Kontext) and five baselines. SGE-Goal moves the \emph{correct} object to the \emph{correct} final state while preserving the rest of the scene, whereas proprietary editors (Gemini~3 Pro Image, GPT Image~1.5) render more photorealistic frames but often miss the target object or alter unrelated content. As we show next, this photorealism does not translate into downstream success, while adherence does (\cref{tab:dissoc}).}
  \label{fig:qual}
\end{figure}

\input{sec/tab_editing}

If we instead rank generators by image-editing quality, we recover the field's usual ordering, and it is \emph{wrong} for manipulation. \Cref{tab:editing} shows that the proprietary editors lead the editing metrics. Gemini tops our human-aligned evaluator and GPT tops I2I-Bench, with SGE-Goal close behind on BridgeDataV2 and CALVIN, yet both proprietary editors are only mid-pack downstream, exactly inverting the result. By editing quality Gemini looks best, yet by policy success it is mid-pack. The aggregate editing-quality score is only weakly predictive of downstream success (Spearman $\rho{\approx}0.67$ for both evaluators) and, crucially, mis-orders the top methods. Full per-dataset editing scores, including LIBERO, are in the supplement (\cref{sec:appendix:existing-eval}).

\subsubsection{Why editing quality misleads}
\label{sec:exp:dissociation}

\input{sec/tab_dissociation}

The 15-dimension decomposition explains the inversion (\cref{tab:dissoc}; visible qualitatively in \cref{fig:qual}; the full per-dimension figure is in the supplement). We correlate each dimension with downstream success across the generators, separately for CALVIN (nine generators, both policies) and SIMPLER-Bridge (eight generators). The result is a sharp \textbf{dissociation}. Source and destination \emph{adherence} strongly predict policy success under both CALVIN policies ($|\rho|$ from $0.88$ to $0.95$; \cref{tab:dissoc}) and on real-world-aligned SIMPLER-Bridge ($|\rho|{=}0.85$ and $0.66$), whereas \emph{photorealism}, editing-artifact quality, and overall image quality are weakly correlated at best ($|\rho|\!\lesssim\!0.25$ on CALVIN and $0.02$ on Bridge). On CALVIN, an exact permutation test against GR-MG success at CL${=}1$ confirms this is no small-sample artifact; the reported $p$-value is the fraction of generator-label permutations that yield a correlation at least as strong as the observed one, so a small $p$ indicates the correlation is unlikely under chance. Source adherence gives $\rho{=}0.98$ ($p{<}10^{-3}$) and destination adherence $\rho{=}0.92$ ($p{=}0.001$), while photorealism is indistinguishable from chance ($\rho{=}0.03$, $p{=}0.95$). Concretely, SGE-Goal leads on adherence yet scores far lower on photorealism (raw rates in the supplement). A policy consumes adherence and the absence of spurious changes, not visual polish, so an editing-quality leaderboard cannot substitute for downstream evaluation.

\noindent\textbf{The misalignment is not specific to one benchmark.}
The editing-quality inversion appears on \emph{all three} datasets and under \emph{both} editing evaluators, with proprietary editors topping the editing ranking everywhere yet sitting mid-pack downstream (\cref{tab:editing}). The adherence-over-photorealism dissociation likewise holds on \emph{both} downstream benchmarks, CALVIN and real-world-aligned SIMPLER-Bridge. The mismatch between editing quality and control utility is thus a property of the metrics, not of one environment.

\subsection{Ablations}
\label{sec:exp:ablation}

\input{sec/tab_ablation_sr}

\Cref{tab:ablation-sr} ablates the spatial components by downstream success rate, the metric most directly measuring the utility of the generated goal images, across both CALVIN policies (all chain lengths) and SIMPLER-Bridge. Throughout, SGE-Goal (full) $\geq$ w/o final mask $\geq$ w/o masks holds universally, and SGE-Goal (full) $\geq$ w/o final mask $\geq$ base in all but one cell (FLUX.1~Kontext under GR-MG at CL${=}2$). Each mask therefore helps and the masked variants beat the untuned editor, so the gain is a property of the goal images, not of one policy. The \emph{final} edit-region mask, which encodes where objects must end up, is the dominant ingredient, lifting FLUX.1~Kontext from $60.1\%$ to $73.1\%$ (GR-MG) and from $13.8\%$ to $29.2\%$ (GHIL-Glue) at CL${=}1$. Notably, \emph{w/o masks} can drop below \emph{base} under GR-MG at longer chains, evidence that unguided fine-tuning degrades the base editor (plausibly catastrophic forgetting) where mask guidance does not. At CL${=}5$ the full model still retains $30.9\%$ and $17.3\%$ (FK, F2K) under GR-MG while every ablation falls below $4\%$, consistent with our thesis that adherence drives utility. The matching editing-metric ablations (our evaluator and I2I-Bench, on BridgeDataV2 and CALVIN) appear in the supplement (\cref{sec:appendix:edit-ablation}) and agree.

%% file: sec/tab_downstream.tex
\begin{table}[t]
  \caption{\textbf{Downstream policy success (SR\%, $\uparrow$) on CALVIN}, varying only the goal generator, across all chain lengths for two pre-trained, frozen goal-conditioned policies. SGE-Goal yields the best policy at \emph{every} chain length under \emph{both} policies, matches the GR-MG oracle at CL${=}1$ while approaching the GHIL-Glue oracle, and degrades most gracefully. Columns are chain lengths $1$--$5$ under each policy. FK${=}$FLUX.1~Kontext~[Dev], F2K${=}$FLUX.2~[klein], CL${=}$chain length, base${=}$untuned editor. \best{Best} (excl.\ oracle) per column in bold; SGE-Goal rows shaded; ``--'' not run.}
  \label{tab:downstream}
  \centering\scriptsize
  \setlength{\tabcolsep}{3pt}
  \resizebox{\columnwidth}{!}{%
  \begin{tabular}{l ccccc ccccc}
    \toprule
    \multirow{2}{*}{Method} & \multicolumn{5}{c}{GR-MG~\cite{GRMG2025}} & \multicolumn{5}{c}{GHIL-Glue~\cite{GHILGlue2025}} \\
    \cmidrule(lr){2-6}\cmidrule(lr){7-11}
     & 1 & 2 & 3 & 4 & 5 & 1 & 2 & 3 & 4 & 5 \\
    \midrule
    \textit{GT goal (oracle)} & \textit{73.6} & -- & -- & -- & -- & \textit{39.0} & -- & -- & -- & -- \\
    \oursrow SGE-Goal (FK)    & \best{73.1} & \best{59.7} & \best{47.3} & \best{36.8} & \best{30.9} & \best{29.2} & \best{12.0} & \best{3.7} & \best{0.8} & 0.0 \\
    \oursrow SGE-Goal (F2K)   & 71.8 & 53.8 & 34.8 & 21.4 & 17.3 & 22.8 & 8.1 & 1.5 & 0.1 & 0.0 \\
    Gemini 3 Pro Image        & 63.3 & 36.9 & 18.1 & 8.1 & 4.4 & 13.2 & 2.4 & 0.0 & 0.0 & 0.0 \\
    Clover                    & 63.2 & 34.2 & 12.8 & 3.6 & 1.9 & 17.3 & 3.2 & 0.2 & 0.0 & 0.0 \\
    GPT Image 1.5             & 60.2 & 36.7 & 17.1 & 6.2 & 3.4 & 8.0 & 0.7 & 0.0 & 0.0 & 0.0 \\
    Qwen Image Edit           & 59.4 & 33.7 & 13.4 & 4.6 & 2.2 & 7.2 & 1.1 & 0.0 & 0.0 & 0.0 \\
    SuSIE                     & 55.8 & 28.4 & 9.9 & 2.6 & 1.0 & 6.1 & 0.1 & 0.0 & 0.0 & 0.0 \\
    FLUX.2 klein (base)       & 52.0 & 29.6 & 13.4 & 5.2 & 2.9 & 2.9 & 0.0 & 0.0 & 0.0 & 0.0 \\
    FLUX.1 Kontext (base)     & 50.3 & 28.6 & 12.8 & 4.8 & 2.6 & 2.6 & 0.0 & 0.0 & 0.0 & 0.0 \\
    \bottomrule
  \end{tabular}%
  }
\end{table}

%% file: sec/tab_simpler.tex
\begin{table}[t]
  \caption{\textbf{Downstream policy success on SIMPLER-Bridge} (SR\%, $\uparrow$ higher is better). A real-world-aligned WidowX/BridgeData-V2 simulator (\cref{sec:exp:setup}) under the frozen GHIL-Glue~\cite{GHILGlue2025} goal-conditioned policy, across four atomic tasks (24 episodes each). SGE-Goal is the strongest open generator and second overall, behind only Gemini~3 Pro Image; both SGE-Goal variants surpass SuSIE despite SuSIE's home-field advantage, since GHIL-Glue is SuSIE's own low-level policy. \best{Best} per column in bold; SGE-Goal rows shaded; base${=}$untuned editor; F2K${=}$FLUX.2~klein, FK${=}$FLUX.1~Kontext.}
  \label{tab:simpler}
  \centering\footnotesize
  \setlength{\tabcolsep}{4.5pt}
  \begin{tabular}{lccccc}
    \toprule
    Method & Spoon & Carrot & Stack & Eggpl. & Mean \\
    \midrule
    Gemini 3 Pro Image        & \best{16.7} & \best{29.2} & 0.0 & 4.2 & \best{12.5} \\
    \oursrow SGE-Goal (F2K)   & \best{16.7} & 16.7 & 0.0 & \best{8.3} & 10.4 \\
    \oursrow SGE-Goal (FK)    & 8.3 & 20.8 & 0.0 & 4.2 & 8.3 \\
    SuSIE                     & \best{16.7} & 8.3 & \best{4.2} & 0.0 & 7.3 \\
    Qwen Image Edit           & 8.3 & 4.2 & 0.0 & 0.0 & 3.1 \\
    GPT Image 1.5             & 4.2 & 0.0 & 0.0 & 4.2 & 2.1 \\
    FLUX.2 klein (base)       & 4.2 & 4.2 & 0.0 & 0.0 & 2.1 \\
    FLUX.1 Kontext (base)     & 4.2 & 4.2 & 0.0 & 0.0 & 2.1 \\
    Clover                    & 4.2 & 0.0 & 0.0 & 0.0 & 1.0 \\
    \bottomrule
  \end{tabular}
\end{table}

%% file: sec/tab_vt.tex
\begin{table}[t]
  \caption{\textbf{Visual goals beat text goals at every chain length.} GR-MG success rate (SR\%, $\uparrow$) under image${+}$text (I${+}$T) vs.\ text-only (T) goal conditioning; the visual goal's benefit grows with chain length. I${+}$T uses an SGE-Goal (FLUX.1~Kontext) goal image. This corroborates and extends MUTEX~\cite{MUTEX2023}, which on LIBERO-100 reports (single step) image-only $61.6\%>$ I${+}$T $59.2\%>$ text-only $50.1\%$. CL${=}$chain length.}
  \label{tab:vt}
  \centering\footnotesize
  \setlength{\tabcolsep}{6pt}
  \begin{tabular}{lccccc}
    \toprule
    Conditioning & CL1 & CL2 & CL3 & CL4 & CL5 \\
    \midrule
    \oursrow Image${+}$text (I${+}$T) & \best{73.1} & \best{59.7} & \best{47.3} & \best{36.8} & \best{30.9} \\
    Text only (T) & 59.0 & 37.8 & 13.1 & 2.0 & 0.2 \\
    \bottomrule
  \end{tabular}
\end{table}

%% file: sec/tab_runtime.tex
\begin{table}[t]
  \caption{\textbf{Runtime} (seconds, $\downarrow$ lower is better). End-to-end goal-image inference is ${\sim}10.6$\,s on a single A10G GPU, entirely on open models, with no proprietary API. The one-time, offline data-generation pipeline (A6000) is separate from inference.}
  \label{tab:runtime}
  \centering\footnotesize
  \setlength{\tabcolsep}{4pt}
  \begin{tabular}{@{}lc@{\hspace{8pt}}lc@{}}
    \toprule
    \multicolumn{2}{c}{Inference (A10G)} & \multicolumn{2}{c}{Data generation (A6000)} \\
    \cmidrule(r){1-2}\cmidrule(l){3-4}
    Component & Time & Stage & Time \\
    \midrule
    Initial mask ($M_i$)     & 2.47 & EVF-SAM + BLIP-2 & 4.46 \\
    Final mask ($M_f$)       & 5.16 & SAM (first frame) & 1.07 \\
    Goal image ($\hat{I}_e$) & 2.92 & XMem tracking & 7.76 \\
    \textbf{Total}           & \textbf{10.55} & LoFTR + KMeans & 8.74 \\
    \bottomrule
  \end{tabular}
\end{table}

%% file: sec/tab_editing.tex
\begin{table}[t]
  \caption{\textbf{Image-editing quality} on BridgeDataV2 (Br), CALVIN (Ca), LIBERO (Li): our human-aligned evaluator (weighted rank, $\downarrow$) and I2I-Bench (weighted score, $\uparrow$). By editing quality, the proprietary editors lead (Gemini tops our evaluator, GPT tops I2I-Bench) with SGE-Goal close behind on Br/Ca, yet both are only mid-pack downstream---\emph{the metrics invert} (\cref{tab:downstream}). \best{Best} per column in bold; SGE-Goal rows shaded. ``Kont.''~abbreviates Kontext.}
  \label{tab:editing}
  \centering\footnotesize
  \setlength{\tabcolsep}{3pt}
  \begin{tabular}{l ccc ccc}
    \toprule
    & \multicolumn{3}{c}{Our eval, rank ($\downarrow$)} & \multicolumn{3}{c}{I2I-Bench, score ($\uparrow$)} \\
    \cmidrule(lr){2-4}\cmidrule(l){5-7}
    Method & Br & Ca & Li & Br & Ca & Li \\
    \midrule
    Gemini 3 Pro Image               & \best{2.07} & \best{2.65} & \best{1.81} & 0.785 & 0.707 & 0.675 \\
    GPT Image 1.5                    & 5.15 & 5.09 & 3.04 & \best{0.801} & \best{0.716} & \best{0.774} \\
    Qwen Image Edit                  & 4.14 & 4.17 & 2.06 & 0.646 & 0.642 & 0.680 \\
    \oursrow SGE-Goal (FLUX.2 klein) & 2.99 & 3.41 & 4.98 & 0.655 & 0.626 & 0.484 \\
    \oursrow SGE-Goal (FLUX.1 Kont.) & 3.81 & 3.18 & 5.54 & 0.649 & 0.616 & 0.479 \\
    Clover                           & --   & 5.48 & --   & --    & 0.606 & --    \\
    SuSIE                            & 6.16 & 7.86 & 5.39 & 0.486 & 0.407 & 0.377 \\
    FLUX.1 Kontext (base)            & 5.48 & 7.33 & 6.51 & 0.533 & 0.449 & 0.378 \\
    FLUX.2 klein (base)              & 6.20 & 5.83 & 6.66 & 0.568 & 0.470 & 0.324 \\
    \bottomrule
  \end{tabular}
\end{table}

%% file: sec/tab_dissociation.tex
\begin{table}[t]
  \caption{\textbf{Adherence/photorealism dissociation.} Strength of correlation (absolute Spearman $|\rho|$; higher means more predictive) between goal-image \emph{adherence} or \emph{photorealism} and downstream policy success. CALVIN values are averaged over chain lengths $1$--$5$ for the GR-MG and GHIL-Glue (GHIL) policies; SIMPLER-Bridge is atomic. Adherence strongly predicts success on every benchmark, whereas photorealism does not. The full 15-dimension version is in the supplement.}
  \label{tab:dissoc}
  \centering\footnotesize
  \setlength{\tabcolsep}{5pt}
  \begin{tabular}{@{}lccc@{}}
    \toprule
    & \multicolumn{2}{c}{CALVIN $|\rho|$} & Bridge \\
    \cmidrule(lr){2-3}
    Dimension & GR-MG & GHIL & $|\rho|$ \\
    \midrule
    \multicolumn{4}{@{}l}{\textit{Adherence}} \\
    \quad Source adherence (R14)       & \best{0.92} & \best{0.95} & \best{0.85} \\
    \quad Destination adherence (R15)  & \best{0.90} & \best{0.88} & \best{0.66} \\
    \midrule
    \multicolumn{4}{@{}l}{\textit{Photorealism}} \\
    \quad Photorealism (R13)           & 0.25 & 0.07 & 0.02 \\
    \bottomrule
  \end{tabular}
\end{table}

%% file: sec/tab_ablation_sr.tex
\begin{table}[t]
  \caption{\textbf{Ablation by downstream success rate} (SR\%, $\uparrow$ higher is better). Each spatial component improves the policy on CALVIN, under both the GR-MG and GHIL-Glue policies across all chain lengths ($1$--$5$), and on SIMPLER-Bridge; the masks matter most under long-horizon chaining. ``full'' uses both masks and the static-scene loss; base${=}$untuned editor. \best{Best} per block in bold; SGE-Goal rows shaded.}
  \label{tab:ablation-sr}
  \centering\scriptsize
  \setlength{\tabcolsep}{3pt}
  \resizebox{\columnwidth}{!}{%
  \begin{tabular}{l ccccc ccccc c}
    \toprule
    & \multicolumn{10}{c}{CALVIN (SR\% at chain length)} & \multirow{3}{*}{\shortstack{SIMPLER-\\Bridge}} \\
    \cmidrule(lr){2-11}
    & \multicolumn{5}{c}{GR-MG} & \multicolumn{5}{c}{GHIL-Glue} & \\
    \cmidrule(lr){2-6}\cmidrule(lr){7-11}
    & 1 & 2 & 3 & 4 & 5 & 1 & 2 & 3 & 4 & 5 & \\
    \midrule
    \multicolumn{12}{@{}l}{\textbf{FLUX.1 Kontext}} \\
    \oursrow SGE-Goal (full)  & \best{73.1} & \best{59.7} & \best{47.3} & \best{36.8} & \best{30.9} & \best{29.2} & \best{12.0} & \best{3.7} & \best{0.8} & 0.0 & \best{8.3} \\
    \quad w/o Final Mask      & 60.1 & 28.4 & 12.9 & 5.7 & 3.5 & 13.8 & 0.3 & 0.0 & 0.0 & 0.0 & 5.2 \\
    \quad w/o Masks           & 50.7 & 21.3 & 7.5 & 2.8 & 1.1 & 7.3 & 0.0 & 0.0 & 0.0 & 0.0 & 5.2 \\
    \quad base                & 50.3 & 28.6 & 12.8 & 4.8 & 2.6 & 2.6 & 0.0 & 0.0 & 0.0 & 0.0 & 2.1 \\
    \midrule
    \multicolumn{12}{@{}l}{\textbf{FLUX.2 klein}} \\
    \oursrow SGE-Goal (full)  & \best{71.8} & \best{53.8} & \best{34.8} & \best{21.4} & \best{17.3} & \best{22.8} & \best{8.1} & \best{1.5} & \best{0.1} & 0.0 & \best{10.4} \\
    \quad w/o Final Mask      & 63.5 & 35.0 & 15.2 & 5.7 & 2.9 & 13.0 & 1.5 & 0.0 & 0.0 & 0.0 & 7.3 \\
    \quad w/o Masks           & 60.0 & 32.1 & 12.5 & 4.1 & 1.9 & 12.6 & 1.2 & 0.0 & 0.0 & 0.0 & 2.1 \\
    \quad base                & 52.0 & 29.6 & 13.4 & 5.2 & 2.9 & 2.9 & 0.0 & 0.0 & 0.0 & 0.0 & 2.1 \\
    \bottomrule
  \end{tabular}%
  }
\end{table}

%% file: sec/5_discussion.tex
\section{Discussion and conclusion}
\label{sec:discussion}

\noindent\textbf{Practical recommendation.}
For deploying goal-conditioned policies, our results invert the instinct to reach for the strongest available editor. A small open base model fine-tuned with SGE-Goal (on-premise, ${\sim}10.6$\,s/edit; \cref{tab:runtime}) produces better control targets than a proprietary API, with a decisive advantage on the multi-step tasks that dominate real deployments. Goal generators should be selected, and benchmarked, by downstream success, not editing-quality leaderboards.

\noindent\textbf{Beyond manipulation.}
The principle of spatially-grounded decomposed editing should extend to other domains where objects undergo safety-critical \emph{displacement} under scarce task-specific data (e.g.\ visualizing post-operative anatomical configurations); we leave such transfers to future work.

\noindent\textbf{Limitations.}
We report these candidly. \emph{(i)~2D grounding:} SGE-Goal works in the image plane and does not reason about 3D geometry, so large scale or viewpoint changes and heavy self-occlusion remain challenging. \emph{(ii)~Consistency is not uniform:} by performing genuine edits, SGE-Goal can perturb nearby objects more than a model that edits little; its advantage is adherence and the suppression of policy-relevant spurious changes, not minimal change. \emph{(iii)~Downstream scope:} our closed-loop evaluation spans two CALVIN policies and SIMPLER-Bridge, driving sub-goal-trained policies with a single goal per atomic step; physical-robot rollouts are a next step. \emph{(iv)~Pipeline dependence:} data generation chains large, human-supervised foundation models, so it is compute-intensive (\cref{tab:runtime}) and can be brittle out of distribution (novel objects, viewpoints, textures); extraction is reliable in aggregate (\cref{fig:dataquality}). The supplement (\cref{sec:supp:failure}) analyzes failure cases of both pipelines.

\noindent\textbf{Conclusion.}
Our SGE-Goal framework turns small open models into the best goal generators on CALVIN and second only to Gemini~3 Pro Image on SIMPLER-Bridge. Its strength comes from treating a goal image as a control target rather than an artwork, since image-editing quality is a misaligned proxy for goal-image utility on both CALVIN and real-world-aligned SIMPLER-Bridge. We have released our code\refcode, datasets\refdata, and checkpoints\refckpt{} for research on visual goals.

%% file: sec/supp_ported.tex
\section{I2I-Bench}
\label{sec:appendix:existing-eval}

I2I-Bench~\cite{I2IBench2025} offers 30 evaluation dimensions, but only the 15 Single-Reference dimensions are valid for evaluating manipulation goal image synthesis pipelines. We report the weighted average of these 15 dimension scores as the single metric for comparing methods. The evaluation results presented in this section are computed on the respective test sets of BridgeDataV2-teleoperated, CALVIN-ABCD, and LIBERO-10-90-Combined.

\subsection{Weight Assignment}
\label{sec:appendix:i2ibench}
I2I-Bench provides 15 Single-Reference editing-quality dimensions but no task-specific way to combine them. We use them to build a \emph{balanced image-editing metric with a focus on robotic manipulation}. It is \emph{balanced} in that every dimension contributes, so the aggregate spans the full spectrum of editing quality---instruction adherence, scene preservation, placement, visual coherence, and general image quality---rather than collapsing onto any single axis. It is \emph{focused} in that, per dimension, the editing-quality aspects most pertinent to a manipulation goal (whether the instructed edit is performed and the surrounding scene preserved) are weighted above the general visual-quality aspects, while aspects irrelevant to our domain (aesthetics, text rendering) are weighted least. We group dimensions into tiers and split each tier's budget equally among its members.

\noindent\textbf{Instruction adherence (30\%; 10\% each).} Object-manipulation accuracy, action/state-change accuracy, and instruction-following accuracy measure whether the edit performs the change the instruction specifies---the editing-quality aspect most pertinent to a manipulation goal, so it receives the largest budget.

\noindent\textbf{Scene preservation (24\%; 12\% each).} Non-edited-region fidelity and subject-identity fidelity measure whether content outside the edit is left unchanged; spurious background or identity changes are undesirable in a goal image, so these are weighted comparably to adherence.

\noindent\textbf{Placement \& attribute accuracy (16\%; 8\% each).} Spatial accuracy and local-attribute accuracy measure whether edited content is placed correctly with the right local properties---also pertinent to a faithful edit, and, like the two tiers above, weighted more per dimension than the general visual-quality dimensions below.

\noindent\textbf{Visual coherence (20\%; 5\% each).} Physical plausibility, blending naturalness, editing artifacts, and composition \& interaction assess general image coherence and freedom from generation artifacts. As standard editing-quality criteria, all four are retained so the metric stays balanced, but each carries a smaller per-dimension weight since it describes the image's general appearance rather than the requested edit.

\noindent\textbf{Ancillary quality (10\%; 2.5\% each).} Overall image quality, aesthetic quality, world-knowledge reasoning, and text-content accuracy capture generic properties largely orthogonal to robotic manipulation, so they receive the lowest per-dimension weight while still contributing to a balanced metric.

\subsection{I2I-Bench Results}
Table~\ref{tab:i2ibench-combined} presents the weighted average of I2I-Bench Single-Reference dimension scores for all compared methods on Bridge\-DataV2-tele\-operated, CALVIN-ABCD, and LIBERO-10-90-Com\-bined.

\begin{table}[t]
  \caption{Weighted average of I2I-Bench Single-Reference dimension scores ($\uparrow$) on BridgeDataV2-teleoperated (Bridge), CALVIN-ABCD (CALVIN), and LIBERO-10-90-Combined (LIBERO). Best per column in bold (proprietary editors top this metric); SGE-Goal rows shaded. ``Kont.''~abbreviates Kontext.}
  \label{tab:i2ibench-combined}
  \centering\footnotesize
  \setlength{\tabcolsep}{4pt}
  \begin{tabular}{lccc}
    \toprule
    & \multicolumn{3}{c}{I2I-Bench score ($\uparrow$)} \\
    \cmidrule(l){2-4}
    Method & Bridge & CALVIN & LIBERO \\
    \midrule
    GPT Image 1.5                & \best{0.801} & \best{0.716} & \best{0.774} \\
    Gemini 3 Pro Image           & 0.785 & 0.707 & 0.675 \\
    \oursrow SGE-Goal (FLUX.2 klein)  & 0.655 & 0.626 & 0.484 \\
    \oursrow SGE-Goal (FLUX.1 Kont.)  & 0.649 & 0.616 & 0.479 \\
    Qwen Image Edit              & 0.646 & 0.642 & 0.680 \\
    Clover                       & ---   & 0.606 & ---   \\
    FLUX.2 klein (base)          & 0.568 & 0.470 & 0.324 \\
    FLUX.1 Kontext (base)        & 0.533 & 0.449 & 0.378 \\
    SuSIE                        & 0.486 & 0.407 & 0.377 \\
    \bottomrule
  \end{tabular}
\end{table}

\begin{figure*}[t]
    \centering
    \includegraphics[width=0.95\linewidth]{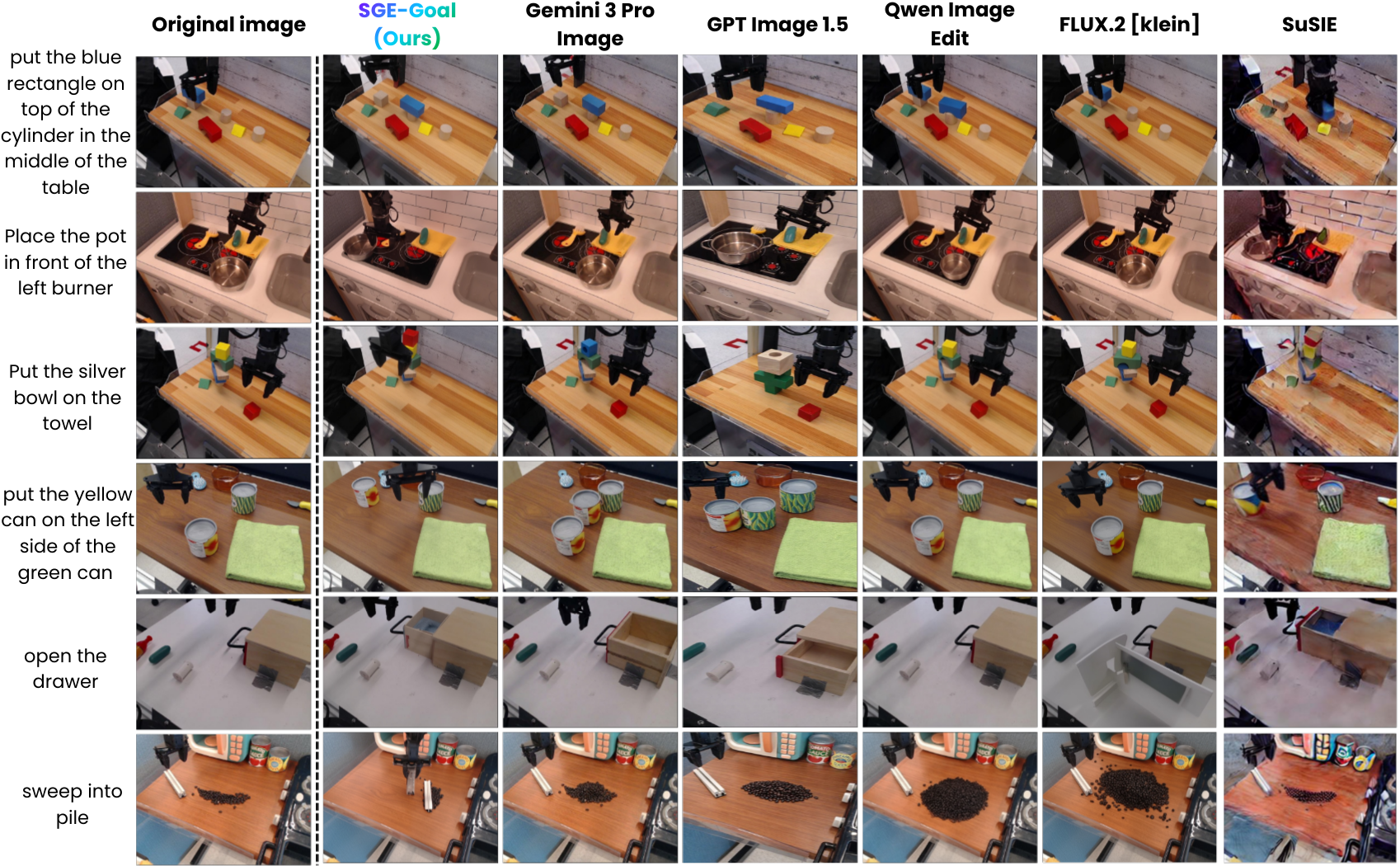}
    \caption{Additional subjective comparisons on BridgeDataV2-teleoperated. SGE-Goal (FLUX.2 [klein]) applies consistent, instruction-adherent edits.}
    \label{fig:comparison-bridge-supp}
\end{figure*}

\begin{figure*}[t]
    \centering
    \includegraphics[width=0.95\linewidth]{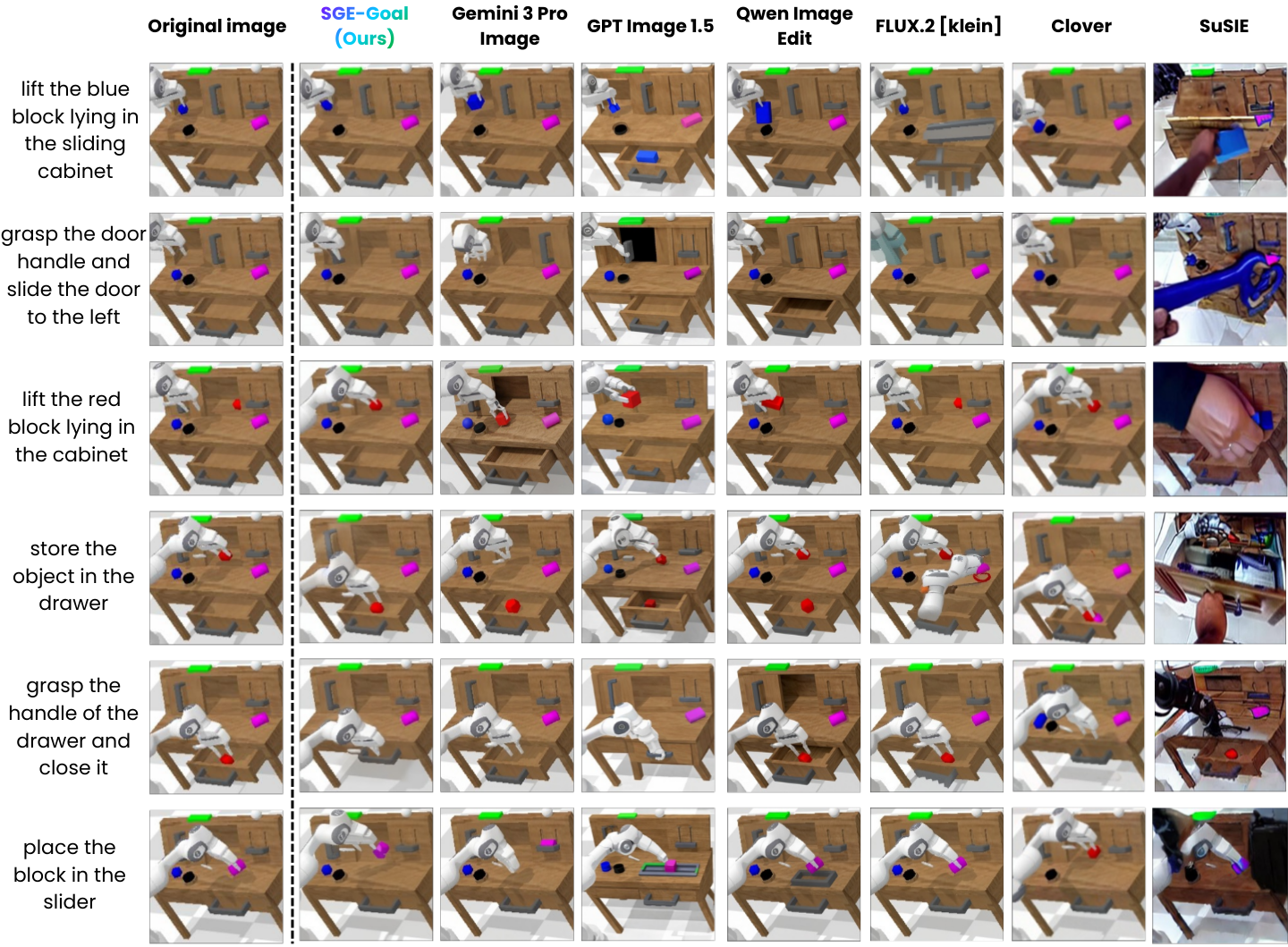}
    \caption{Subjective comparisons on CALVIN-ABCD. SGE-Goal (FLUX.2 [klein]) applies consistent, instruction-adherent edits.}
    \label{fig:comparison-calvin-supp}
\end{figure*}

\begin{figure*}[t]
    \centering
    \includegraphics[width=0.95\linewidth]{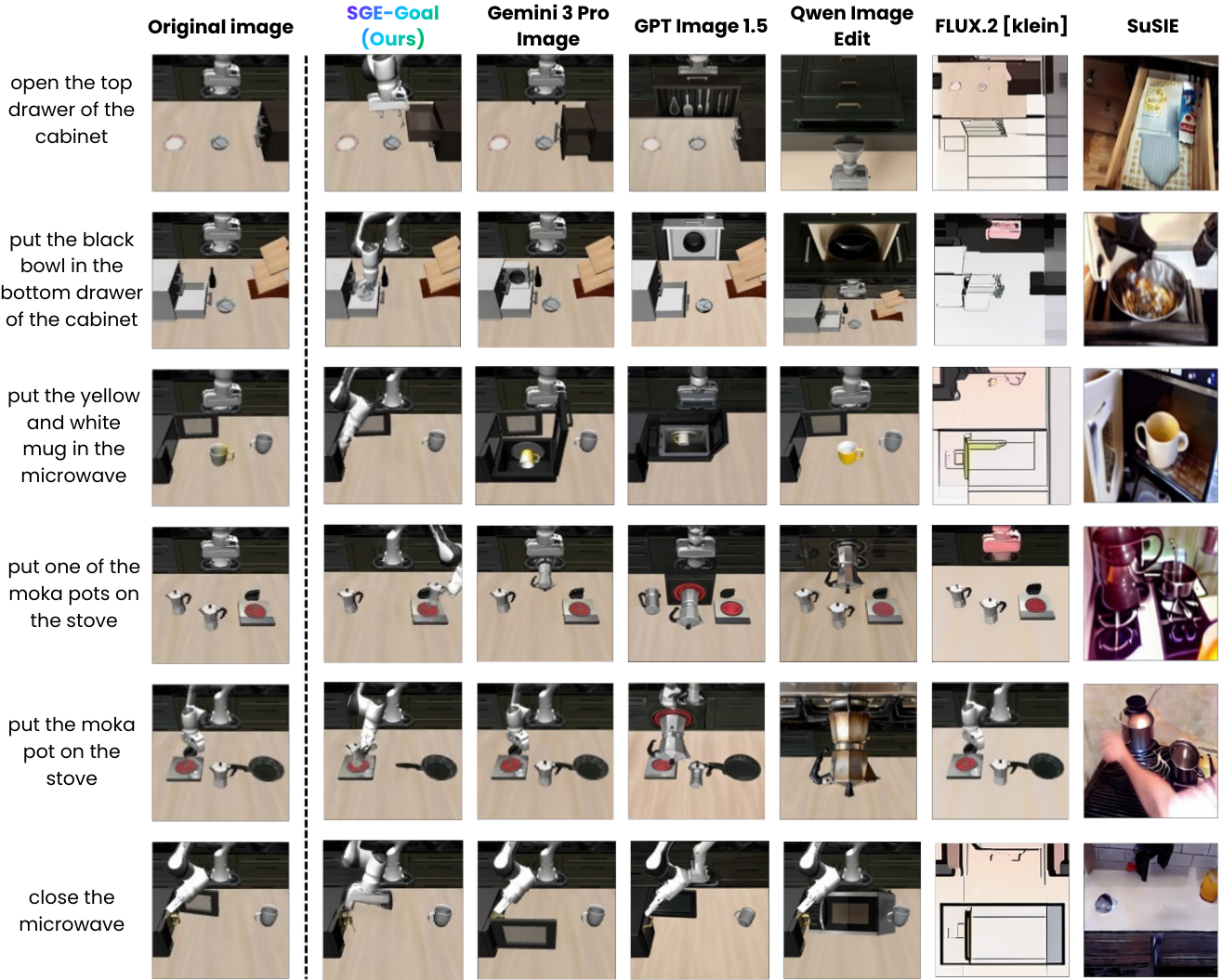}
    \caption{Subjective comparisons on LIBERO-10-90-Combined. SGE-Goal (FLUX.2 [klein]) applies consistent, instruction-adherent edits.}
    \label{fig:comparison-libero-supp}
\end{figure*}

\section{Editing-metric ablations}
\label{sec:appendix:edit-ablation}

\Cref{tab:supp-ablation-ours,tab:i2ibench-ablation} ablate the spatial components under our 15-dimension evaluator and under I2I-Bench, on BridgeDataV2 and CALVIN. Adding the masks improves both editing metrics on both datasets, corroborating the success-rate ablation in the main paper and its repetition under Octo-ft (\cref{tab:supp-octo-ablation}). The one exception is FLUX.2 [klein] under our evaluator on CALVIN, where the no-mask variant slightly edges the full model. A small reordering like this under an editing metric, while the full model keeps the highest success rate under all three CALVIN policies, is exactly the editing-versus-utility gap that motivates our work.

\begin{table}[t]
  \caption{Editing-metric ablation under our 15-dimension evaluator (weighted average rank, $\downarrow$ lower is better), on BridgeDataV2 and CALVIN.}
  \label{tab:supp-ablation-ours}
  \centering\footnotesize
  \setlength{\tabcolsep}{6pt}
  \begin{tabular}{lcc}
    \toprule
    Configuration & Bridge ($\downarrow$) & CALVIN ($\downarrow$) \\
    \midrule
    \multicolumn{3}{@{}l}{\textit{FLUX.1 Kontext [Dev]}} \\
    \oursrow \quad SGE-Goal (full)   & \best{1.89} & \best{1.67} \\
    \quad w/o final mask             & 2.01 & 2.19 \\
    \quad w/o masks                  & 2.97 & 2.83 \\
    \quad base                       & 3.13 & 3.32 \\
    \midrule
    \multicolumn{3}{@{}l}{\textit{FLUX.2 [klein]}} \\
    \oursrow \quad SGE-Goal (full)   & \best{1.39} & 2.33 \\
    \quad w/o final mask             & 2.59 & 2.56 \\
    \quad w/o masks                  & 2.72 & \best{2.08} \\
    \quad base                       & 3.30 & 3.04 \\
    \bottomrule
  \end{tabular}
\end{table}

\begin{table}[t]
  \caption{I2I-Bench ablation (weighted score, $\uparrow$ higher is better) on BridgeDataV2 and CALVIN. Mask guidance helps under the external metric on both datasets.}
  \label{tab:i2ibench-ablation}
  \centering\footnotesize
  \setlength{\tabcolsep}{6pt}
  \begin{tabular}{lcc}
    \toprule
    Configuration & Bridge ($\uparrow$) & CALVIN ($\uparrow$) \\
    \midrule
    \multicolumn{3}{@{}l}{\textit{FLUX.1 Kontext [Dev]}} \\
    \oursrow \quad SGE-Goal (full)   & \best{0.649} & \best{0.616} \\
    \quad w/o final mask             & 0.585 & 0.557 \\
    \quad w/o masks                  & 0.526 & 0.513 \\
    \quad base                       & 0.533 & 0.449 \\
    \midrule
    \multicolumn{3}{@{}l}{\textit{FLUX.2 [klein]}} \\
    \oursrow \quad SGE-Goal (full)   & \best{0.655} & \best{0.626} \\
    \quad w/o final mask             & 0.633 & 0.618 \\
    \quad w/o masks                  & 0.620 & 0.614 \\
    \quad base                       & 0.568 & 0.470 \\
    \bottomrule
  \end{tabular}
\end{table}

\section{Data Pipeline and Evaluation Frameworks}
\label{sec:appendix:methods}

\subsection{Data-Extraction Pipeline Details}
\label{sec:appendix:data-pipeline}

The main paper provides a concise overview of our data generation pipeline. Here we describe each stage in full detail.

\subsubsection{Stage 1: Robotic Arm Masking and First-Frame Selection.}
The pipeline samples first 20\% of frames uniformly from the input manipulation video and runs EVF-SAM~\cite{EVFSAM2024} on each sampled frame with the text prompt ``robotic arm'' to produce candidate arm masks. Each candidate mask is verified with BLIP-2~\cite{BLIP2_2023} visual question answering by posing the question ``Is it a robotic gripper?'' on the masked region. A frame is accepted as containing a valid arm detection only if the arm is consistently detected across multiple consecutive sampled frames (requiring consecutive detection confirmations). The first such frame is selected as the original image $I_o$.

\subsubsection{Stage 2a: SAM Object Proposal Generation and Filtering.}
SAM ViT-H~\cite{SAM2023} is applied to $I_o$ to generate class-agnostic segmentation proposals. The resulting masks undergo two deduplication steps: (i)~\textit{sub-mask removal}---any mask whose intersection-over-union (IoU) with a larger mask exceeds 0.82 is discarded as redundant; (ii)~\textit{complete-linkage merging}---masks with pairwise IoU $\geq$ 0.82 are merged into a single mask. The arm mask from Stage~1 is excluded from the object proposals. Each surviving mask represents a distinct object in the scene.

\subsubsection{Stage 2b: XMem Temporal Object Tracking.}
The filtered masks are loaded as initial annotations and tracked through the video using XMem~\cite{XMem2022} (top-$k$ filtering = 30, memory update every 5 frames, 128 memory prototypes, frames resized to 480px on the long edge). This provides per-object segmentation masks on every frame of the video, enabling downstream movement analysis.

\subsubsection{Stage 3: Dynamic Object Identification via Motion Clustering.}
For each tracked object, movement distance between its region in $I_o$ and its region in a later frame is computed using LoFTR~\cite{LoFTR2021} (outdoor weights). LoFTR performs dense feature matching between the two object crops, and the distance is taken as the minimum of (a)~the mean Euclidean distance between matched feature point positions and (b)~the Euclidean distance between the two mask centroids. If LoFTR finds no matches, the centroid distance serves as the sole metric. The pipeline searches the last 25\% of video frames for appropriate comparison frames.

KMeans clustering ($K{=}2$) is applied to the computed distances across all objects to separate them into two groups: the cluster with the higher mean distance is labeled \textit{dynamic} (manipulated objects), and the other is labeled \textit{stationary} (background objects).

\subsubsection{Stage 4: Edit Mask Generation and Edited Frame Identification.}
The initial edit-region mask $M_i$ is constructed as the union of the arm mask and all dynamic-object masks on $I_o$. To identify the edited frame $I_e$---the frame where manipulation completes---the pipeline monitors the overlap between the arm mask and the dynamic-object masks across frames. As long as the arm overlaps with the dynamic objects (overlap threshold = 0.25), the arm is still interacting with them. The edited frame is identified as the last frame where this overlap ceases---the arm has released the objects for the final time. The final edit-region mask $M_f$ is the union of the arm mask and dynamic-object masks on the identified edited frame.

\subsubsection{Stage 5: Instruction Generation and Dataset Assembly.}
For datasets that lack textual manipulation instructions, VideoChat-Flash VLM~\cite{VideoChatFlash2024} generates instructions $T$ from the video content. The resulting $(I_o, I_e, M_i, M_f, T)$ tuples are split into training, validation, and test sets using MD5 hash-based assignment (seed 42) for reproducibility.

\subsubsection{Discussion.}
We highlight two key design decisions underlying our data extraction pipeline and discuss their rationale.

\noindent\textbf{Movement-Based vs.\ Open-Vocabulary Object Detection.}
An alternative approach to identifying the target (dynamic) object in the scene would be to use a large language model to extract a textual description of the target object from the edit instruction, and then use an open-vocabulary object segmentation model to predict its mask directly. However, this approach is fundamentally limited by the imperfect open-vocabulary capability of current segmentation models---they may fail to detect the target object due to vocabulary limitations, scene occlusions, or ambiguous phrasing in the edit instruction. Our movement-based approach circumvents these issues entirely: rather than needing to identify a specific object by name, it computes movement distances for \textit{all} objects in the scene and selects those exhibiting significant displacement. This makes the pipeline robust to unusual object names, occluded objects, and imprecise instructions alike.

\noindent\textbf{Composite vs.\ Atomic Manipulations.}
As described in the main paper, we distinguish \textit{composite} manipulations from \textit{atomic} ones. A composite manipulation can be decomposed into multiple constituent steps (e.g., ``open the drawer and put the apple in it'' decomposes into ``open the drawer'' followed by ``put the apple in the drawer''), whereas an atomic manipulation cannot be decomposed further without introducing unnecessary steps. Whether a manipulation is composite or atomic can depend on the environment---for example, ``put the broccoli in the pot'' is atomic if the pot is open, but composite if the pot has a lid that must first be removed.

It is critical to decompose composite manipulations into \textit{only} their necessary and sufficient atomic constituents, without hallucinating extra steps. Introducing redundant decomposition steps wastes computational resources during both training data generation and inference, and can confuse the downstream manipulation policy by presenting intermediate goal states that do not correspond to meaningful sub-tasks. By operating strictly at the atomic level, our pipeline ensures that each generated training sample corresponds to a single, well-defined manipulation action.

\noindent\textbf{Dataset-Specific Handling of Atomic Granularity.}
Since SGE-Goal operates at the atomic level, all training data must consist of atomic manipulation video segments. The three datasets we process differ in how they expose this granularity. BridgeDataV2-teleoperated already consists of atomic samples and can be processed directly. CALVIN-ABCD provides composite sequences but includes explicit start--end frame annotations for each constituent atomic step, so we simply split at the annotated boundaries. LIBERO provides composite samples without official atomic splits; we therefore use an LLM to decompose each composite instruction into atomic constituents and detect the corresponding frame spans from gripper open/close transitions. The resulting atomic segments from all three datasets are then processed by our pipeline (Stages~1--5) to produce training tuples $(I_o, I_e, M_i, M_f, T)$.

\begin{figure}[t]
    \centering
    \includegraphics[width=1.0\linewidth]{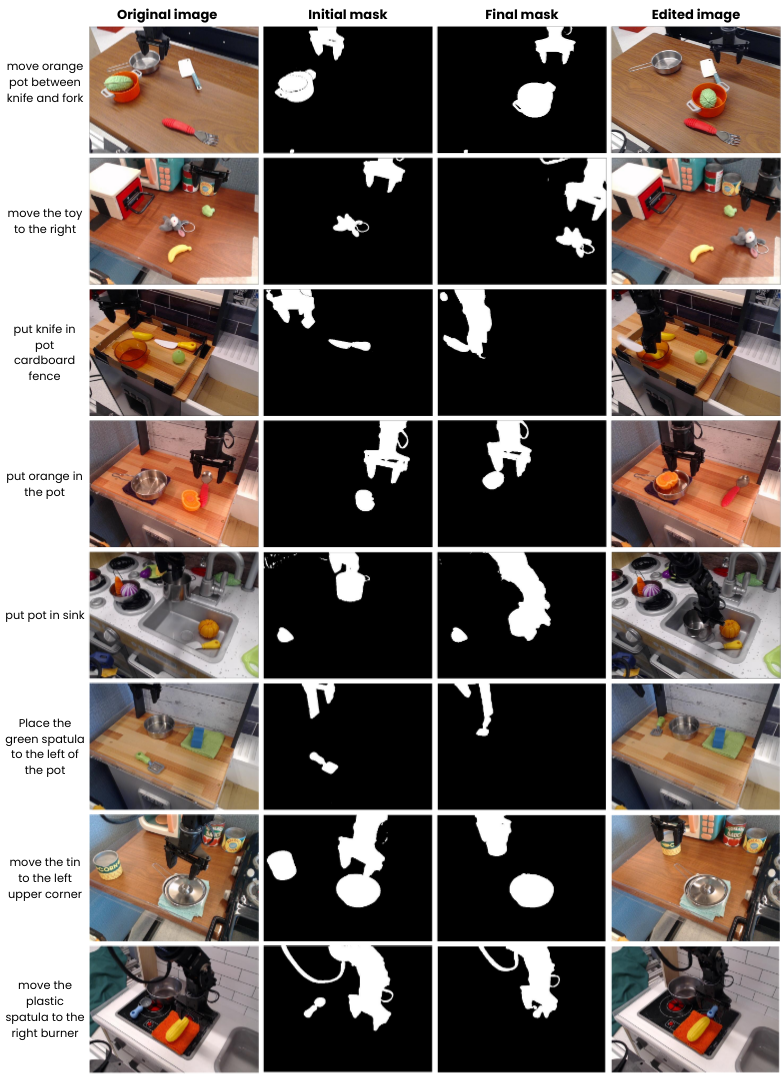}
    \caption{Training data samples from BridgeDataV2-teleoperated generated by our automated data extraction pipeline. Each sample consists of the original image $I_o$, initial edit-region mask $M_i$, final edit-region mask $M_f$, edited image $I_e$, and the manipulation instruction $T$.}
    \label{fig:data-samples-bridge}
\end{figure}

\begin{figure}[t]
    \centering
    \includegraphics[width=1.0\linewidth]{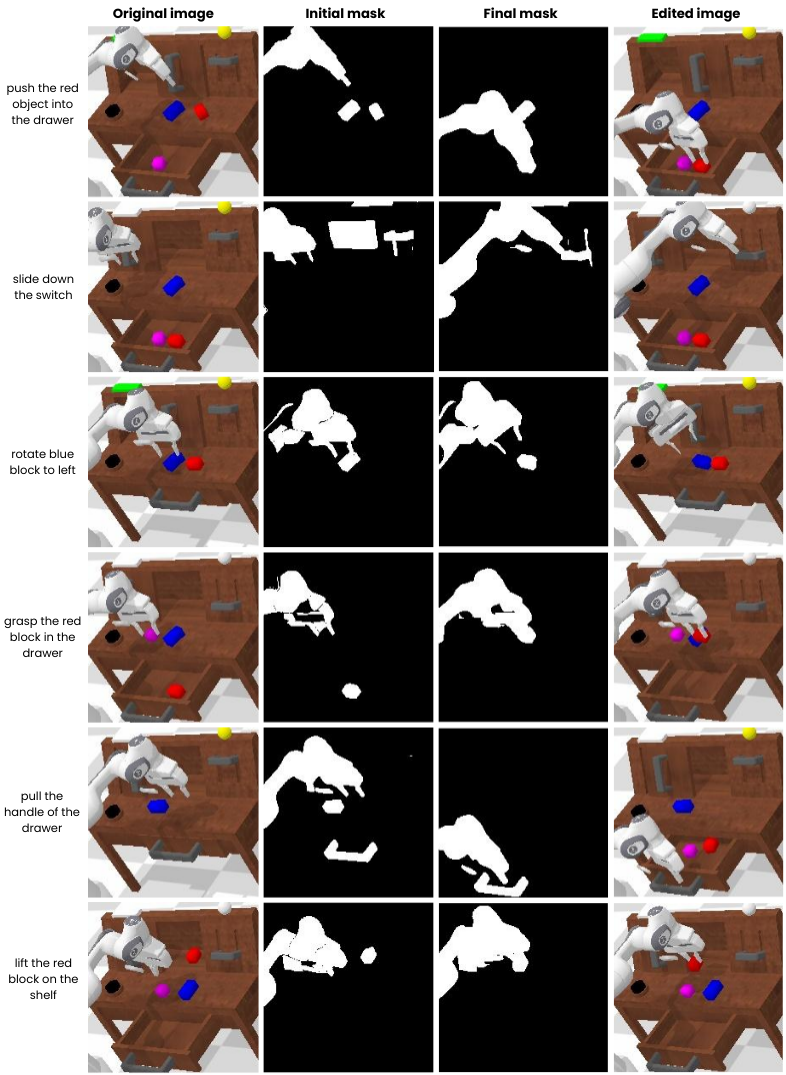}
    \caption{Training data samples from CALVIN-ABCD generated by our automated data extraction pipeline. Each sample consists of the original image $I_o$, initial edit-region mask $M_i$, final edit-region mask $M_f$, edited image $I_e$, and the manipulation instruction $T$.}
    \label{fig:data-samples-calvin}
\end{figure}

\begin{figure}[t]
    \centering
    \includegraphics[width=1.0\linewidth]{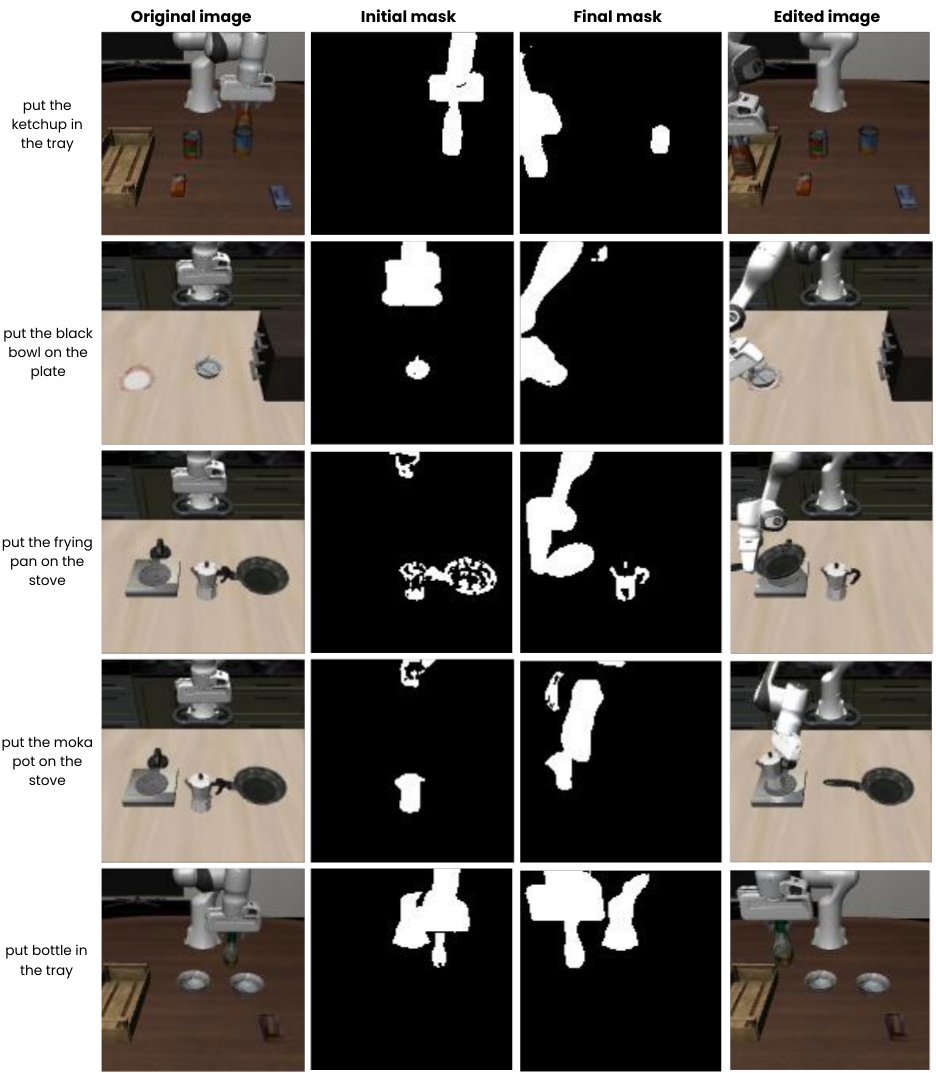}
    \caption{Training data samples from LIBERO-10-90-Combined generated by our automated data extraction pipeline. Each sample consists of the original image $I_o$, initial edit-region mask $M_i$, final edit-region mask $M_f$, edited image $I_e$, and the manipulation instruction $T$.}
    \label{fig:data-samples-libero}
\end{figure}

\subsection{Evaluation Dimension Details}
\label{sec:appendix:eval-framework}

The main paper provides a concise overview of our 15-dimension evaluation framework. Here we describe the implementation of each evaluation dimension in detail. All VLM-based dimensions use VideoChat-Flash (VCF)~\cite{VideoChatFlash2024} with deterministic decoding (\texttt{do\_sample=False, temperature=0.0}). VCF-based dimensions construct a two-frame video (original $\rightarrow$ edited) and issue structured chain-of-thought prompts. The verbatim prompts are provided in Section~\ref{sec:appendix:vlm-prompts}.

\subsubsection{R1---Unmentioned Stationary Objects.}
This dimension counts instruction-mentioned objects that should have moved but remained stationary. The pipeline first extracts the set of target objects from the manipulation instruction. For each target object, a multi-step VCF verification is performed: (i)~a \textit{movement check} prompt (Section~\ref{sec:appendix:vlm-prompts}, R1) asks whether the object moved or changed between the original and edited frames; (ii)~a \textit{position confirmation} prompt verifies whether the object is still at its original position. Action-specific handling is implemented for instructions involving removal (checking visibility in the edited frame), opening/closing (checking state change with dedicated prompts), and flipping/turning (checking orientation change with container-aware and non-container prompts, plus an orientation fallback). An object is counted as ``unmentioned stationary'' if VCF confirms it did not move (or its state did not change) despite the instruction requiring it.

\subsubsection{R2---Unmentioned Moved Objects.}
This dimension counts objects that should have remained stationary but were displaced within the scene. The pipeline chains four models: (i)~SAM~\cite{SAM2023} generates object proposals on $I_o$; (ii)~XMem~\cite{XMem2022} tracks each proposal to obtain its mask in $I_e$; (iii)~LoFTR~\cite{LoFTR2021} computes per-object movement distances between the original and tracked masks (minimum of feature-matching distance and centroid distance); (iv)~KMeans ($K{=}2$) clustering separates objects into moved and stationary groups based on these distances. Objects classified as ``moved'' are then filtered: a local LLM (Qwen2.5-7B-Instruct~\cite{Qwen2.5_2024}) extracts the target object names from the instruction (see the instruction target extraction prompt in Section~\ref{sec:appendix:vlm-prompts}, R2/R3), and BLIP-2~\cite{BLIP2_2023} identifies each moved object. Objects that match the instruction targets are excluded, as their movement is expected. The remaining count gives the number of unmentioned moved objects.

\subsubsection{R3---Unmentioned Removed Objects.}
This dimension counts objects that should have remained in the scene but vanished entirely. It shares the SAM $\rightarrow$ XMem pipeline with R2, but instead of computing movement distances, it identifies objects whose tracked masks have near-zero area in the edited frame---indicating that the object has disappeared from the scene. As in R2, BLIP-2 identifies each vanished object, and objects whose removal is warranted by the instruction are excluded from the count.

\subsubsection{R4---Hallucinated/Newly-Appeared Objects.}
This dimension counts objects that appear in $I_e$ without the instruction warranting their presence. The pipeline employs an extensive multi-step VCF + Grounding DINO~\cite{GroundingDINO2024} approach: (i)~VCF extracts instruction objects and inventories all objects visible in Frame~1; (ii)~for each instruction target, a \textit{duplicate object detection} prompt (Section~\ref{sec:appendix:vlm-prompts}, R4) checks whether the original object remains at its initial position while a copy appears at the new location---a common ``duplication hallucination'' pattern; (iii)~\textit{structural/appearance change detection} and \textit{object substitution detection} prompts verify that manipulated objects retain their identity; (iv)~a \textit{new object scan} prompt identifies objects present in Frame~2 but absent from Frame~1; (v)~Grounding DINO provides bounding-box-level verification for candidate hallucinations; (vi)~a \textit{human hand / extra arm detection} prompt checks for anomalous human hands or additional robotic arms; (vii)~each candidate hallucination undergoes a final \textit{random hallucination verification} prompt to reduce false positives.

\subsubsection{R5, R6, R7---Unrelated Color, Texture, Shape Changes.}
These three dimensions detect per-object property changes not warranted by the instruction. The pipeline first isolates individual objects using SAM $\rightarrow$ XMem tracking: SAM generates object proposals on $I_o$, XMem tracks them to $I_e$, and only objects with significant mask area in both frames are retained. For each retained object, isolated two-frame crops (object region on black background) are extracted from both $I_o$ and $I_e$ and assembled into a short two-frame video. VCF then evaluates each crop pair with per-property prompts (Section~\ref{sec:appendix:vlm-prompts}, R5--R7): R5 checks for color changes, R6 for texture changes, and R7 for shape deformation. The count of objects exhibiting each type of unwarranted change is reported.

\subsubsection{R8, R9---Unrelated Brightness Changes.}
These dimensions detect unwarranted scene brightening (R8) or darkening (R9) using a hybrid CV + VLM ensemble. The CV component converts both images to the LAB color space and computes histogram-based brightness metrics: mean brightness difference, percentage change ratio, and percentile-based differences. Detection thresholds are: mean difference $>$ 4.5, ratio $>$ 4\%, and percentile difference $>$ 6.0. The ensemble operates in ``cv\_primary'' mode: the CV analysis serves as the primary detector, with VCF prompts (Section~\ref{sec:appendix:vlm-prompts}, R8--R9) providing secondary confirmation when the CV analysis is ambiguous.

\subsubsection{R10, R11, R12---Unrelated Camera Changes.}
These dimensions detect unwarranted zoom (R10), crop (R11), and perspective/angle shifts (R12). The primary detector uses AKAZE feature detection + RANSAC homography estimation between $I_o$ and $I_e$. From the estimated homography matrix: zoom is detected via the log-scale of the transformation (threshold: 0.015); crop is detected by analyzing black/masked regions in $I_e$; perspective change is detected via projection coefficients (threshold: $9 \times 10^{-4}$) and shear/rotation components (threshold: 0.15). A \textit{perspective precedence rule} prevents perspective changes from being mislabeled as zoom: if perspective is detected and the scale change is not large (log-scale $<$ 0.15), zoom is suppressed. VCF serves as a secondary detector (Section~\ref{sec:appendix:vlm-prompts}, R10--R12), with the CV result taking priority in the ensemble.

\subsubsection{R13---Photorealism.}
This dimension detects loss of photorealistic quality in the edited image. The pipeline combines VCF assessment with CV texture metrics. VCF performs a two-stage evaluation: (i)~an \textit{artistic style detection} prompt (Section~\ref{sec:appendix:vlm-prompts}, R13) checks whether the edited image exhibits non-photographic characteristics (cartoon, sketch, painting, etc.); (ii)~an \textit{AI artifact detection} prompt checks for diffusion model artifacts (plasticky surfaces, waxy appearance, over-smoothing, airbrushed look). The CV component computes texture metrics including Laplacian variance (blur detection), edge density, and gradient magnitude statistics, providing quantitative measures of texture quality degradation.

\subsubsection{R14---Source Adherence.}
This dimension verifies whether the correct object was manipulated. VCF first identifies which object changed between the two frames using an \textit{object identification} prompt (Section~\ref{sec:appendix:vlm-prompts}, R14). A \textit{source verification} prompt then confirms whether the identified object matches the target object specified in the instruction. If VCF identifies a different object as having changed, or if the target object appears unchanged, the sample is flagged as a source adherence failure.

\subsubsection{R15---Destination Adherence.}
This dimension verifies whether the manipulated object reached the correct final state or location. VCF issues multiple specialized prompts (Section~\ref{sec:appendix:vlm-prompts}, R15) depending on the instruction type: \textit{orientation check} for flip/turn instructions, \textit{state change check} for open/close instructions, \textit{arrangement check} for placement into containers, \textit{pickup check} for removal from a location, \textit{take-out check} for extraction, and \textit{destination check} for placement at a specific location. A \textit{generic destination fallback} prompt handles cases not covered by the specialized prompts.

\subsection{Detailed Analysis of Existing Evaluation Frameworks}
\label{sec:appendix:eval-pitfalls}

\noindent\textbf{PICABench Limitations:}
\begin{itemize}
    \item Highly dependent on various types of human annotations, limiting scalability.
    \item Uses its own proprietary data---difficult to obtain equivalent evaluation setup for robotic manipulation.
    \item Cannot evaluate arbitrary (original image, edited image, instruction) triplets.
\end{itemize}

\vspace{1mm}
\noindent\textbf{RISEBench Limitations:}
\begin{itemize}
    \item Not specifically curated for robotic manipulation domain; dimensions are not comprehensive enough.
    \item Dimensions like ``appearance consistency'' are high-level without granular breakdowns (e.g., no separation of moved/stationary/hallucinated objects).
    \item Uses only GPT-4.1 as sole judge, inferior to our ensemble approach.
\end{itemize}

\vspace{1mm}
\noindent\textbf{I2I-Bench Limitations:}
\begin{itemize}
    \item Not specifically curated for robotic manipulation; only a few dimensions relate to object manipulation.
    \item Model diversity limited to LLMs, segmentation models, and embedding models for feature similarity.
    \item Lacks specialized models for tracking (XMem), movement distance computation (LoFTR), and moved/stationary object clustering.
\end{itemize}

\vspace{1mm}
\noindent\textbf{DICE Limitations:}
\begin{itemize}
    \item Provides only binary coherence (coherent vs. incoherent) without granular categorization into specific issue types.
    \item Training the coherence estimator required 116 manually annotated samples with GPT-4o rationale generation and human review.
    \item Creating a ``DICE for robotic manipulation'' variant would require substantial human intervention.
\end{itemize}

\vspace{1mm}
\noindent\textbf{GIE-Bench Limitations:}
\begin{itemize}
    \item Too rudimentary and not comprehensive.
    \item Not a granular evaluation approach---only Masked Aligned MSE and Functional Correctness.
    \item Not at all curated for the robotic manipulation domain.
\end{itemize}

\section{VLM Evaluation Prompts}
\label{sec:appendix:vlm-prompts}

All VLM-based evaluation dimensions use \textbf{VideoChat-Flash} (VCF; Open\-GVLab/\hspace{0pt}Video\-Chat-Flash-Qwen2\_5-7B\_\hspace{0pt}Intern\-Video2-1B) with deterministic decoding (\texttt{do\_sample=False, temperature=0.0}). Each dimension constructs a two-frame video (original $\rightarrow$ edited) and issues one or more structured prompts. Below we list every prompt verbatim, organized by evaluation dimension. Placeholders in curly braces (e.g., \texttt{\{instruction\}}, \texttt{\{object\_name\}}) are filled at runtime.

\subsection{R1---Unmentioned Stationary Objects}

R1 checks whether objects \emph{not} mentioned in the editing instruction remained unchanged. Different instruction types require different verification strategies; the prompts below are organized by instruction category.

\paragraph{Case A: Movement instructions (\texttt{move}/\texttt{put}/\texttt{pick\,up}).}
Two-step confirmation: first detect whether the object moved, then verify it is still at its original position.

\textbf{Step~1---Movement Check:}
\begin{lstlisting}[basicstyle=\tiny\ttfamily,breaklines=true]
This is a 2-frame video: Frame 1 is BEFORE, Frame 2 is AFTER.

Look at the "{object_name}".

Did the "{object_name}" MOVE or CHANGE between Frame 1 and Frame 2?
(Moving means: position changed, rotated, flipped, appeared, disappeared, or any change)

Answer YES or NO:
\end{lstlisting}

\textbf{Step~2---Position Confirmation:}
\begin{lstlisting}[basicstyle=\tiny\ttfamily,breaklines=true]
This is a 2-frame video: Frame 1 = BEFORE, Frame 2 = AFTER.

Look at the "{object_name}" in Frame 1. Note its EXACT position.

In Frame 2, is there STILL a "{object_name}" at that SAME EXACT position?
(A copy may have appeared elsewhere - focus ONLY on the original spot)

Answer YES if original spot still has it, NO if original spot is empty:
\end{lstlisting}

\paragraph{Case B: Removal instructions (\texttt{remove}/\texttt{take\,away}).}
A single visibility check determines whether each non-target object remains in the scene.

\textbf{Removal Visibility Check:}
\begin{lstlisting}[basicstyle=\tiny\ttfamily,breaklines=true]
This is a 2-frame video: Frame 1 (BEFORE), Frame 2 (AFTER).

The task was to REMOVE certain objects from the scene.

Looking at Frame 2 (the AFTER/second frame):
Can you see a "{object_desc}" still present in the scene?

Answer YES if you can see it anywhere in Frame 2.
Answer NO only if you are CERTAIN it is completely gone.

YES or NO:
\end{lstlisting}

\paragraph{Case C: State-change instructions (\texttt{open}/\texttt{close}).}
Two prompts verify whether a non-target object's state changed and whether the original state persists.

\textbf{Step~1---State Change Check:}
\begin{lstlisting}[basicstyle=\tiny\ttfamily,breaklines=true]
This is a 2-frame video: Frame 1 (BEFORE), Frame 2 (AFTER).

The instruction was to {action} the "{object_name}".

In Frame 1, the "{object_name}" appears to be {original_state}.

Look carefully at Frame 2:
Is the "{object_name}" now {target_state} in Frame 2?

Focus on the actual physical state (door position, lid position, flap position).
Answer YES if it is now {target_state}.
Answer NO if it still looks {original_state}.

YES or NO:
\end{lstlisting}

\textbf{Step~2---State Persistence Check:}
\begin{lstlisting}[basicstyle=\tiny\ttfamily,breaklines=true]
This is a 2-frame video: Frame 1 (BEFORE), Frame 2 (AFTER).

In Frame 1, the "{object_name}" is {original_state}.

Look very carefully at Frame 2:
Is the original {original_state} "{object_name}" STILL visible in Frame 2?
(Ignore any new elements. Focus ONLY on whether the original {original_state} version persists.)

Answer YES if the {original_state} version is still visible in Frame 2.
Answer NO if the {original_state} version is completely gone.

YES or NO:
\end{lstlisting}

\paragraph{Case D: Orientation instructions (\texttt{flip}/\texttt{turn}).}
Container objects (bowls, cups) use opening-visibility checks; other objects use orientation matching. A fallback prompt handles cases where the two-step check is inconclusive.

\textbf{Container Objects---Frame~1:}
\begin{lstlisting}[basicstyle=\tiny\ttfamily,breaklines=true]
This is a 2-frame video: Frame 1 (BEFORE), Frame 2 (AFTER).

Look at the "{object_name}" in Frame 1 ONLY.
Can you look down INTO the opening/inside of the "{object_name}" in Frame 1?
Or is the "{object_name}" flipped over, tilted, or lying on its side so the opening faces away?

Answer OPENING_VISIBLE if you can see into the {object_name}.
Answer OPENING_HIDDEN if the {object_name} is flipped/tilted so you cannot see inside:
\end{lstlisting}

\textbf{Container Objects---Frame~2:}
\begin{lstlisting}[basicstyle=\tiny\ttfamily,breaklines=true]
This is a 2-frame video: Frame 1 (BEFORE), Frame 2 (AFTER).

Look at the "{object_name}" in Frame 2 ONLY.
Can you look down INTO the opening/inside of the "{object_name}" in Frame 2?
Or is the "{object_name}" flipped over, tilted, or lying on its side so the opening faces away?

Answer OPENING_VISIBLE if you can see into the {object_name}.
Answer OPENING_HIDDEN if the {object_name} is flipped/tilted so you cannot see inside:
\end{lstlisting}

\textbf{Non-Container Objects---Frame~1:}
\begin{lstlisting}[basicstyle=\tiny\ttfamily,breaklines=true]
This is a 2-frame video: Frame 1 (BEFORE), Frame 2 (AFTER).

Look at the "{object_name}" in Frame 1 ONLY.
Is the "{object_name}" currently {original_orientation} in Frame 1?

Answer YES or NO:
\end{lstlisting}

\textbf{Non-Container Objects---Frame~2:}
\begin{lstlisting}[basicstyle=\tiny\ttfamily,breaklines=true]
This is a 2-frame video: Frame 1 (BEFORE), Frame 2 (AFTER).

Look at the "{object_name}" in Frame 2 ONLY.
Is the "{object_name}" now {target_orientation} in Frame 2?

Answer YES or NO:
\end{lstlisting}

\textbf{Fallback---Direct Orientation Comparison:}
\begin{lstlisting}[basicstyle=\tiny\ttfamily,breaklines=true]
This is a 2-frame video: Frame 1 (BEFORE), Frame 2 (AFTER).

The instruction was: "{instruction}"
This means the "{object_name}" should change from {original_orientation} to {target_orientation}.

Compare the "{object_name}" in Frame 1 vs Frame 2:
Did the orientation of the "{object_name}" ACTUALLY CHANGE between the two frames?

Answer CHANGED or SAME:
\end{lstlisting}

\subsection{R2---Unmentioned Moved Objects / R3---Unmentioned Removed Objects}

R2 and R3 primarily use SAM $\rightarrow$ XMem $\rightarrow$ LoFTR (R2) or vanishing-mask detection (R3) $\rightarrow$ BLIP-2 filtering. Instruction-target extraction is performed by a local LLM (Qwen2.5-7B-Instruct) rather than VCF:

\textbf{Instruction Target Extraction (Local LLM---Qwen2.5-7B-Instruct):}
\begin{lstlisting}[basicstyle=\tiny\ttfamily,breaklines=true]
You are an AI assistant that extracts object names from manipulation instructions.

Manipulation Instruction:
"{instruction}"

Your task is to extract the names of objects that are INTENDED TO BE MOVED according to the
manipulation instruction.

RULES:
1. Only extract objects that the instruction explicitly says should be moved/picked up/placed/put.
2. Do NOT include destination objects (where something is placed onto/into) unless they are also
   being moved.
3. Do NOT include the robot, robotic arm, gripper, or any agent performing the action.
4. Return object names as a simple comma-separated list.
5. If no objects are intended to be moved, return "none".

EXAMPLES:
- "The robotic arm moves the red cube to the right" -> "red cube"
- "The robotic arm picks up the apple and puts it on the plate" -> "apple"
  (plate is the destination, not moved)
- "The robotic arm removes the blue cup from the table" -> "blue cup"
- "The robotic arm moves the book to the top of the jar" -> "book"
  (jar is the destination, not moved)
- "Put the bottle in the drawer" -> "bottle" (drawer is the destination, not moved)
- "Move both the apple and the orange to the basket" -> "apple, orange"
  (basket is the destination)
- "The robotic arm places the cup on top of the saucer" -> "cup"
  (saucer is the destination surface)
- "The robotic arm puts the egg on top of the bread" -> "egg"
  (bread is the destination surface, not moved)
- "Turn the knob clockwise" -> "knob"

Now extract the objects from the given instruction. Output ONLY the comma-separated object names
(or "none"):
\end{lstlisting}

\subsection{R4---Hallucinated/Newly-Appeared Objects}

R4 uses a multi-step pipeline combining VCF with Grounding DINO. It first extracts instruction objects and inventories Frame~1. Then, for each instruction-target object, it checks for duplication, structural changes, and substitution. Finally, it scans for entirely new objects, human hands / extra arms, destination anomalies, and verifies each candidate.

\textbf{Same-Object Verification:}
\begin{lstlisting}[basicstyle=\tiny\ttfamily,breaklines=true]
TASK: Determine if two detected objects refer to the SAME physical object in the scene.

Detected object 1: "{obj1}"
Detected object 2: "{obj2}"

Look at Frame 2 (the after image) carefully.

Question: In Frame 2, are "{obj1}" and "{obj2}" referring to the SAME physical object, 
or are they TWO DIFFERENT physical objects in the scene?

Consider:
- If there's only ONE object that could be called either "{obj1}" or "{obj2}", they are the SAME.
- If there are TWO SEPARATE objects (one that is "{obj1}" and another that is "{obj2}"), they are DIFFERENT.

Answer with exactly one of:
SAME_OBJECT - They refer to the same physical object in the scene
DIFFERENT_OBJECTS - They are two distinct physical objects in the scene

ANSWER: [SAME_OBJECT or DIFFERENT_OBJECTS]
\end{lstlisting}

\textbf{Instruction Object Extraction:}
\begin{lstlisting}[basicstyle=\tiny\ttfamily,breaklines=true]
TASK: Extract ALL objects mentioned in this robotic manipulation instruction.

Instruction: "{instruction}"

List ALL physical objects mentioned in the instruction, including:
- Objects being moved/picked up
- Objects being placed onto or into (destination objects)
- Any other objects referenced

Do NOT include: the robot, arm, gripper, or abstract locations like "left" or "right".

For example:
- "put the apple on the plate" -> apple, plate
- "move the cup next to the bowl" -> cup, bowl
- "place sushi on top of omelette" -> sushi, omelette

Output format: comma-separated list of object names only.
If no specific objects mentioned, output: none
\end{lstlisting}

\textbf{Frame~1 Object Inventory:}
\begin{lstlisting}[basicstyle=\tiny\ttfamily,breaklines=true]
TASK: Complete Object Inventory of Frame 1

Look at Frame 1 (the BEFORE image) and list ALL distinct objects you can see.

Include EVERYTHING:
- Food items (fruits, vegetables, prepared food, etc.)
- Utensils (spoons, forks, knives, etc.)
- Containers (bowls, plates, cups, boxes, etc.)
- Toys (balls, blocks, cubes, etc.)
- Tools and other items
- Background objects

Do NOT include:
- The robotic arm
- The table/surface itself
- Walls/floor

Output format: comma-separated list of all objects visible in Frame 1.
\end{lstlisting}

\textbf{Duplicate Object Detection (per instruction-target object):}
\begin{lstlisting}[basicstyle=\tiny\ttfamily,breaklines=true]
TASK: Duplicate Object Detection

You are comparing Frame 1 (BEFORE) and Frame 2 (AFTER) of a robotic manipulation.
The instruction was: "{instruction}"
Focus on this object: "{target}"

CRITICAL QUESTION: In Frame 2 (the AFTER image), do you see TWO or MORE instances of "{target}"?

Look VERY carefully at Frame 2 and answer:
1. How many "{target}" objects do you see in Frame 1?
2. How many "{target}" objects do you see in Frame 2?
3. Is there still a "{target}" at its ORIGINAL position from Frame 1 in Frame 2?

A DUPLICATION HALLUCINATION occurs when:
- The original "{target}" is STILL at its original position in Frame 2
- AND there is ALSO a "{target}" at a new position (where it was moved to)
- Result: You see 2 copies of the same object in Frame 2

A CORRECT EDIT means:
- The "{target}" moved from its original position to a new position
- The original position is now EMPTY
- Result: You see only 1 "{target}" in Frame 2, just at a different location

ANSWER FORMAT:
Count_in_Frame1: [number]
Count_in_Frame2: [number]
Original_position_empty_in_Frame2: [YES or NO]
Verdict: [DUPLICATED or NOT_DUPLICATED]
\end{lstlisting}

\textbf{Structural/Appearance Change Detection (per instruction-target object):}
\begin{lstlisting}[basicstyle=\tiny\ttfamily,breaklines=true]
TASK: Structural/Appearance Change Detection

Compare the "{target}" in Frame 1 (BEFORE) vs Frame 2 (AFTER).

The instruction was: "{instruction}"

Focus on the PHYSICAL STRUCTURE and SHAPE of "{target}":

1. SHAPE: Does "{target}" have the SAME shape in both frames?
   - Same length/width/height?
   - Same curvature/angles?
   - Same overall silhouette?

2. STRUCTURE: Does "{target}" have the SAME structure?
   - Same number of parts?
   - Parts in same arrangement?
   - Nothing bent, broken, or deformed?

3. APPEARANCE: Does "{target}" look like the SAME object?
   - Could this be a different "{target}" swapped in?
   - Does it look significantly altered?

IMPORTANT: Position changes are OK. We're looking for SHAPE/STRUCTURE changes.

A STRUCTURAL HALLUCINATION means:
- The object's shape changed significantly (bent, stretched, shrunk, deformed)
- Parts are missing or added
- It looks like a DIFFERENT version of the object

Structural_Change: [NONE, MINOR, or MAJOR]
Description: [what structural changes do you see, if any]
Verdict: [STRUCTURAL_HALLUCINATION or NO_STRUCTURAL_CHANGE]
\end{lstlisting}

\textbf{Object Substitution Detection (placement instructions only):}
\begin{lstlisting}[basicstyle=\tiny\ttfamily,breaklines=true]
TASK: Object Substitution Detection

The instruction says: "{instruction}"
The instruction wants to place/move: "{target}"

Look at Frame 2 (AFTER image) carefully at the destination location:

1. What object is NOW at the destination (the {destination})?
2. Is this the same type of object as "{target}", or is it a DIFFERENT object type?

Examples of SUBSTITUTION:
- Instruction: "put sweet potato in pan" but a CORN is in the pan = SUBSTITUTION
- Instruction: "put red apple in bowl" but a GREEN BALL is in the bowl = SUBSTITUTION

Examples of NO SUBSTITUTION:
- The "{target}" is at the destination = NO SUBSTITUTION
- The destination is empty = NO SUBSTITUTION
- The "{target}" moved but looks the same = NO SUBSTITUTION

Answer format:
Object_at_destination: [what object is at the destination in Frame 2]
Is_it_the_same_as_target: [YES if same type as "{target}", NO if different]
Was_substituted: [YES if a different object was placed instead, NO otherwise]
\end{lstlisting}

\textbf{New Object Scan:}
\begin{lstlisting}[basicstyle=\tiny\ttfamily,breaklines=true]
TASK: Find Completely NEW Objects

Compare Frame 1 (BEFORE) and Frame 2 (AFTER).

Objects that existed in Frame 1: {frame1_objects}

Look for objects in Frame 2 that:
1. Were NOT present ANYWHERE in Frame 1
2. Are real physical objects (not shadows, reflections, or blur)
3. Are NOT the robotic arm

Check CAREFULLY for:
- Any NEW balls (red ball, blue ball, etc.)
- Any NEW fruits or vegetables
- Any NEW utensils or tools
- Any objects that appear "out of nowhere"

List any NEW objects that appear in Frame 2 but did NOT exist in Frame 1.
Be thorough - check all areas: foreground, background, corners, edges.

Output: comma-separated list of new objects, or "none" if no new objects found.
\end{lstlisting}

\textbf{Human Hand / Extra Arm Detection:}
\begin{lstlisting}[basicstyle=\tiny\ttfamily,breaklines=true]
TASK: Detect HUMAN HANDS or EXTRA ROBOTIC ARMS

Look at Frame 2 (AFTER image) VERY carefully.

1. HUMAN HAND/ARM CHECK - Look for ANY of these in Frame 2:
   - Human fingers (5 fingers, skin color)
   - Human palm or hand
   - Human arm or forearm (skin visible, not metal)
   - A hand that looks like it belongs to a person, not a robot
   - IMPORTANT: If you see an object being held or placed by what looks like human fingers, answer YES

2. EXTRA ROBOT ARM CHECK:
   - Count robotic arms/grippers in Frame 1
   - Count robotic arms/grippers in Frame 2
   - If Frame 2 has MORE robot arms than Frame 1, answer YES

Answer format (answer YES or NO for each):
Human_hand_visible: [YES if any human hand/fingers/arm visible, NO otherwise]
Number_of_robot_arms_frame1: [count]
Number_of_robot_arms_frame2: [count]
Extra_arm_appeared: [YES if more robot arms in Frame 2, NO otherwise]
List_of_anomalies: [list any human body parts or extra arms found, or None]
\end{lstlisting}

\textbf{Unexpected Objects at Destination (placement instructions only):}
\begin{lstlisting}[basicstyle=\tiny\ttfamily,breaklines=true]
TASK: Check for unexpected objects at destination

Compare Frame 1 (BEFORE) and Frame 2 (AFTER).
Focus on the {destination}.

1. List ALL objects inside/on the {destination} in Frame 1 (BEFORE):
2. List ALL objects inside/on the {destination} in Frame 2 (AFTER):

Are there any EXTRA objects in/on the {destination} in Frame 2 that were NOT inside/on it in Frame 1?
(The object from the instruction "{instruction}" is expected - only list UNEXPECTED extras)

New_unexpected_objects: [list any extra unexpected objects, or NONE]
\end{lstlisting}

\textbf{Amorphous Hallucination Detection (placement instructions only):}
\begin{lstlisting}[basicstyle=\tiny\ttfamily,breaklines=true]
TASK: Check for AMORPHOUS hallucinations (spills, liquid, residue, etc.)

Compare Frame 1 (BEFORE) and Frame 2 (AFTER).
Focus on: {destination}

Look for any NEW substances that appeared in/on the {destination} in Frame 2:
- Liquid, fluid, or spilled substance
- Powder, residue, or coating
- Any amorphous material that wasn't there before

IMPORTANT: Only flag substances that are NOT part of the intended instruction "{instruction}".

Is there any NEW amorphous substance (liquid, fluid, residue, etc.) in/on the {destination} in Frame 2 that was NOT there in Frame 1?

Answer: [YES - describe what appeared] or [NO - no new substances]
\end{lstlisting}

\textbf{Random Hallucination Verification (per candidate object):}
\begin{lstlisting}[basicstyle=\tiny\ttfamily,breaklines=true]
TASK: Verify if "{obj}" is a Random Hallucination

Object found in Frame 2: "{obj}"
Manipulation instruction: "{instruction}"
Instruction objects: {instruction_objects}

Please verify by answering:

Q1: Is "{obj}" visible ANYWHERE in Frame 1 (before image)?
- Look carefully at the entire Frame 1
- Answer: [YES_IN_FRAME1 or NO_NOT_IN_FRAME1]

Q2: Is "{obj}" one of the instruction objects ({instruction_objects}) or directly related to them?
- Answer: [YES_RELATED or NO_UNRELATED]

Q3: Is "{obj}" a shadow, reflection, or part of the robot arm?
- Answer: [YES_ARTIFACT or NO_REAL_OBJECT]

FINAL DETERMINATION:
- If Q1=NO_NOT_IN_FRAME1 AND Q2=NO_UNRELATED AND Q3=NO_REAL_OBJECT -> HALLUCINATION
- Otherwise -> NOT_HALLUCINATION

Q1_ANSWER: [YES_IN_FRAME1 or NO_NOT_IN_FRAME1]
Q2_ANSWER: [YES_RELATED or NO_UNRELATED]
Q3_ANSWER: [YES_ARTIFACT or NO_REAL_OBJECT]
FINAL: [HALLUCINATION or NOT_HALLUCINATION]
\end{lstlisting}

\subsection{R5, R6, R7---Unrelated Color, Texture, and Shape Changes}

For each tracked object, isolated two-frame crops are fed to VCF with the following per-attribute prompts:

\textbf{R5---Color Change:}
\begin{lstlisting}[basicstyle=\tiny\ttfamily,breaklines=true]
Compare this object in Frame 1 (before) and Frame 2 (after).
Has this object's COLOR clearly changed? (e.g., changed from one distinct color to another,
became a different color, went from multi-colored to single-colored)

Ignore: brightness differences, lighting changes, shadows, viewing angle differences.
Only answer "Yes" for CLEAR, OBVIOUS color changes.

Answer ONLY "Yes" or "No".
\end{lstlisting}

\textbf{R6---Texture Change:}
\begin{lstlisting}[basicstyle=\tiny\ttfamily,breaklines=true]
Compare this object in Frame 1 (before) and Frame 2 (after).
Has this object's SURFACE TEXTURE been clearly altered?

Answer "Yes" ONLY if you see clear evidence of:
- Surface became noticeably blurry, smeared, or lost fine detail
- Surface became noticeably noisy, grainy, or pixelated
- Surface pattern fundamentally changed (textured became smooth, striped became plain)
- Surface material or finish clearly changed
- Surface quality clearly degraded (looks processed or AI-generated)

Answer "No" if:
- The surface texture looks essentially the same in both frames
- Differences are only in lighting, brightness, or viewing angle
- The object is partially hidden but visible parts have same texture
- You are unsure or the change is very subtle

Answer ONLY "Yes" or "No".
\end{lstlisting}

\textbf{R7---Shape Change:}
\begin{lstlisting}[basicstyle=\tiny\ttfamily,breaklines=true]
Compare this object in Frame 1 (before) and Frame 2 (after).

CRITICAL: Has this object been PHYSICALLY DEFORMED or WARPED?

This means the actual physical object has been permanently altered - NOT just viewed
from a different angle, partially hidden, or moved to a new position.

Answer "Yes" ONLY if you see UNDENIABLE evidence of:
- The object is bent, crushed, warped, stretched, or physically deformed
- The object has a completely different physical shape (square became round, half became whole)
- Parts have clearly broken off or been physically added
- The object looks visibly enlarged, shrunk, or distorted in shape

Answer "No" if ANY of these apply:
- Object just moved to a new position (same shape, different location)
- Object is partially hidden behind something (that is occlusion, NOT deformation)
- Object looks slightly different due to camera angle or perspective change
- You are not absolutely certain the shape physically changed
- The difference could be explained by viewing angle, partial visibility, or lighting

IMPORTANT: When in doubt, answer "No". Only say "Yes" for OBVIOUS physical deformation.

Answer ONLY "Yes" or "No".
\end{lstlisting}

\subsection{R8, R9---Unrelated Brightness Changes}

VCF is used in a hybrid ensemble with CV-based histogram analysis. The VLM prompts:

\textbf{R8---Brighter:}
\begin{lstlisting}[basicstyle=\tiny\ttfamily,breaklines=true]
TASK: Detect if the scene became BRIGHTER between Frame 1 and Frame 2.
Instruction: "{instruction}"

Compare the OVERALL ILLUMINATION and AMBIENT LIGHTING between:
- Frame 1 (Original image, before edit)
- Frame 2 (Edited image, after edit)

FOCUS ON:
1. Background brightness - Is the background more lit up?
2. Surface reflections - Do surfaces appear more reflective/washed out?
3. Ambient light level - Does the overall scene look more illuminated?
4. Shadow intensity - Are shadows lighter/less visible?
5. Highlight areas - Are bright spots more prominent or overexposed?

IGNORE (do NOT count as lighting changes):
- Robot arm appearance/brightness
- Object movement (a white object moving is NOT a lighting change)
- Color changes (focus only on luminosity/brightness)

Did the scene become noticeably BRIGHTER in Frame 2 compared to Frame 1?
(This means the ambient light level increased, not just an object moved)

Reasoning: [Analyze the specific brightness indicators listed above]
Verdict: Yes or No
\end{lstlisting}

\textbf{R9---Darker:}
\begin{lstlisting}[basicstyle=\tiny\ttfamily,breaklines=true]
TASK: Detect if the scene became DARKER between Frame 1 and Frame 2.
Instruction: "{instruction}"

Compare the OVERALL ILLUMINATION and AMBIENT LIGHTING between:
- Frame 1 (Original image, before edit)
- Frame 2 (Edited image, after edit)

FOCUS ON:
1. Background darkness - Is the background less lit?
2. Shadow intensity - Are shadows darker/more prominent?
3. Ambient light level - Does the overall scene look less illuminated?
4. Detail visibility - Are details harder to see in dark areas?
5. Overall dimness - Does the scene look like the lights were turned down?

IGNORE (do NOT count as lighting changes):
- Robot arm appearance
- Object removal (if an object leaves, it's not a lighting change)
- Cast shadows from object movement

Did the scene become noticeably DARKER in Frame 2 compared to Frame 1?
(This means the ambient light level decreased, not just an object was removed)

Reasoning: [Analyze the specific darkness indicators listed above]
Verdict: Yes or No
\end{lstlisting}

\subsection{R10, R11, R12---Unrelated Camera Changes}

CV geometry (AKAZE + RANSAC homography) serves as the primary detector. VCF is used as a secondary detector:

\textbf{Camera Transform Detection:}
\begin{lstlisting}[basicstyle=\tiny\ttfamily,breaklines=true]
Frame 1 = Original image. Frame 2 = Edited image.
Instruction: "{instruction}"

TASK: Detect UNRELATED camera changes (NOT requested in instruction).
Focus on BACKGROUND elements only. IGNORE changes to objects being edited.

ZOOM detection:
- Compare how much of the room/scene is visible
- Zoomed OUT = can see MORE of the room/table/surroundings in Frame 2
- Zoomed IN = can see LESS of the room (closer view) in Frame 2
- Answer YES if zoom changed

PERSPECTIVE detection - LOOK CAREFULLY:
- Is the camera viewing from a DIFFERENT HEIGHT or ANGLE?
- Low-angle = camera looking UP at the scene
- High-angle = camera looking DOWN (top-down view)
- Side-angle = camera shifted left/right, one side of objects appears larger
- Tilted = scene appears rotated or slanted
- EVEN SUBTLE changes in viewpoint angle should be marked YES
- If background elements appear from different angle, answer YES

CROP detection:
- Are there solid BLACK bars or regions in Frame 2?
- Answer YES only for obvious black masked areas

JSON only: {"Zoom": "Yes"/"No", "Perspective": "Yes"/"No", "Crop": "Yes"/"No", "Reason": "brief"}
\end{lstlisting}

\textbf{Zoom-Focused Check:}
\begin{lstlisting}[basicstyle=\tiny\ttfamily,breaklines=true]
Look at Frame 1 and Frame 2 carefully. Ignore any changes to objects -- focus ONLY on the camera's field of view.

ZOOM CHECK:
1. Count how many distinct objects/surfaces are visible at the EDGES of each frame.
2. Can you see objects at the edges of one frame that are NOT visible in the other?
3. Does the scene in one frame extend further (showing more walls, table surface, floor)?

SIGNS OF ZOOM:
- Frame 2 shows MORE of the scene than Frame 1 = ZOOMED OUT
- Frame 2 shows LESS of the scene than Frame 1 = ZOOMED IN
- Objects near the edges appear/disappear between frames = ZOOM CHANGE
- The "field of view" (how much of the room/table is visible) is different = ZOOM

If BOTH frames show the SAME extent of the scene with the SAME edges, answer No.
If the visible area is DIFFERENT in ANY way, answer Yes.

Answer JSON only: {"Zoom": "Yes"/"No"}
\end{lstlisting}

\subsection{R13---Photorealism}

\hspace*{\parindent}\textbf{Stage~1---Artistic Style Detection:}
\begin{lstlisting}[basicstyle=\tiny\ttfamily,breaklines=true]
Look at Frame 2 (the edited image) and answer:

Does Frame 2 show ANY of these NON-PHOTOGRAPHIC characteristics?
- Cartoon or comic book style (flat colors, bold outlines)
- Sketch or drawing (pencil lines, hatching)
- Painting (visible brush strokes, oil/watercolor texture)
- Illustration (stylized non-realistic rendering)
- Chalk, pastel, crayon, or charcoal look
- Poster art (posterized/banded colors)
- Sparkle, glitter, or twinkling effects added

If the image looks like ARTWORK/ILLUSTRATION instead of a PHOTOGRAPH, answer Yes.
If it looks like a normal photograph, answer No.

Answer: Yes or No
\end{lstlisting}

\textbf{Stage~2---AI Artifact Detection:}
\begin{lstlisting}[basicstyle=\tiny\ttfamily,breaklines=true]
Compare Frame 1 (original photo) with Frame 2 (edited image).

Look at the TEXTURE QUALITY of objects in both frames.

Does Frame 2 show any of these AI/diffusion model artifacts?
1. PLASTICKY surfaces - objects look like plastic toys
2. WAXY appearance - surfaces look like wax figures
3. OVER-SMOOTHED textures - natural texture detail is lost
4. AIRBRUSHED look - overly perfect, no natural imperfections
5. CGI/SYNTHETIC look - objects appear computer-generated

Has the edited image (Frame 2) lost photographic authenticity due to ANY of these artifacts?

Answer: Yes or No
\end{lstlisting}

\subsection{R14---Source Adherence}

R14 verifies that the correct object was manipulated (not a different one).

\textbf{Object Identification:}
\begin{lstlisting}[basicstyle=\tiny\ttfamily,breaklines=true]
This is a 2-frame video: Frame 1 (BEFORE), Frame 2 (AFTER).

What SINGLE object changed position or state between Frame 1 and Frame 2?
Name the object that moved, was added, was removed, or changed state:
\end{lstlisting}

\textbf{Source Verification:}
\begin{lstlisting}[basicstyle=\tiny\ttfamily,breaklines=true]
This is a 2-frame video: Frame 1 (BEFORE), Frame 2 (AFTER).

Target object: "{source_obj}"

QUESTION: Did the "{source_obj}" get {change_desc} between Frame 1 and Frame 2?
(Not a different object, specifically the "{source_obj}")

Answer YES if the "{source_obj}" changed position, state, or was manipulated.
Answer NO if a different object changed instead, or "{source_obj}" is unchanged.

Verdict: [Yes/No]
\end{lstlisting}

\subsection{R15---Destination Adherence}

R15 verifies that the object reached its intended destination or underwent the intended state change. Different instruction types trigger different prompts.

\textbf{Orientation Change (flip/turn instructions):}
\begin{lstlisting}[basicstyle=\tiny\ttfamily,breaklines=true]
This is a 2-frame video: Frame 1 (BEFORE) and Frame 2 (AFTER).

Compare Frame 1 and Frame 2 carefully.

QUESTION: Has the "{source_obj}" been turned or rotated in Frame 2 compared to Frame 1?
Answer Yes ONLY if the "{source_obj}" has a CLEARLY different orientation.
Answer No if the "{source_obj}" looks the same in both frames.

Verdict: [Yes/No]
\end{lstlisting}

\textbf{State Change (open/close instructions):}
\begin{lstlisting}[basicstyle=\tiny\ttfamily,breaklines=true]
This is a 2-frame video: Frame 1 (BEFORE) and Frame 2 (AFTER).

Compare Frame 1 and Frame 2 carefully.

QUESTION: Has the "{source_obj}" been {state_verb} in Frame 2 compared to Frame 1?
Answer Yes if the "{source_obj}" was {opposite_verb} in Frame 1 and is now {state_verb} in Frame 2.
Answer No if it looks the same in both frames.

Verdict: [Yes/No]
\end{lstlisting}

\textbf{Arrangement Check (triggered after state change is confirmed):}
\begin{lstlisting}[basicstyle=\tiny\ttfamily,breaklines=true]
This is a 2-frame video: Frame 1 (BEFORE) and Frame 2 (AFTER).

Look at the "{source_obj}" in Frame 2.
Is there any object ON TOP of or INSIDE the "{source_obj}" in Frame 2 that was NOT there in Frame 1?

Answer YES if a new object appeared on/in the "{source_obj}" after the state change.
Answer NO if nothing new is on/in the "{source_obj}".

Verdict: [Yes/No]
\end{lstlisting}

\textbf{Pickup Check (pick\,up instructions):}
\begin{lstlisting}[basicstyle=\tiny\ttfamily,breaklines=true]
This is a 2-frame video: Frame 1 (BEFORE) and Frame 2 (AFTER).

Compare how the "{source_obj}" relates to the "{original_location}" in Frame 1 vs Frame 2.

QUESTION: Has the "{source_obj}" been picked up or moved from the "{original_location}"?
Answer Yes if the "{source_obj}" has been lifted, moved away, or repositioned from the "{original_location}".
Answer No if the "{source_obj}" is still in the same position on the "{original_location}" as in Frame 1.

Verdict: [Yes/No]
\end{lstlisting}

\textbf{Take-Out Check (take\,out instructions):}
\begin{lstlisting}[basicstyle=\tiny\ttfamily,breaklines=true]
This is a 2-frame video: Frame 1 (BEFORE) and Frame 2 (AFTER).

In Frame 2, can you see the "{source_obj}" somewhere OUTSIDE of the "{original_location}"?

Answer Yes if "{source_obj}" is visible in Frame 2 and NOT in/on the "{original_location}".
Answer No if "{source_obj}" has vanished, is missing, or is still in the "{original_location}".

Verdict: [Yes/No]
\end{lstlisting}

\textbf{Destination Check (standard placement instructions):}
\begin{lstlisting}[basicstyle=\tiny\ttfamily,breaklines=true]
This is a 2-frame video: Frame 1 (BEFORE) and Frame 2 (AFTER).

In Frame 2, is the "{source_obj}" located at/on/in the "{dest_option}"?

Answer Yes ONLY if you can clearly see "{source_obj}" at the "{dest_option}".
Answer No if "{source_obj}" is missing, vanished, or NOT at the "{dest_option}".

Verdict: [Yes/No]
\end{lstlisting}

\textbf{Generic Destination Fallback:}
\begin{lstlisting}[basicstyle=\tiny\ttfamily,breaklines=true]
This is a 2-frame video: Frame 1 (BEFORE) and Frame 2 (AFTER).

In Frame 2 compared to Frame 1, has any object been newly placed at/on/in the "{primary_dest}"?

Answer Yes if you can see a new object at the "{primary_dest}" in Frame 2 that was not there in Frame 1.
Answer No if nothing new is at the "{primary_dest}".

Verdict: [Yes/No]
\end{lstlisting}